\documentclass[]{uiuc_paper}
\pdfoutput=1

\usepackage[toc,page,header]{appendix}
\usepackage{cleveref}
\usepackage{subcaption}
\usepackage{booktabs}
\usepackage{tabularx}
\usepackage{tikz}
\usepackage[most]{tcolorbox}
\usetikzlibrary{arrows.meta,positioning,calc,fit,backgrounds}
\definecolor{egdeep}{HTML}{3F9068}\definecolor{egbase}{HTML}{6F7BC0}\definecolor{egmist}{HTML}{EAF6EE}
\definecolor{egrose}{HTML}{C4788E}\definecolor{egblush}{HTML}{F7DADF}\definecolor{egsage}{HTML}{BFE5CB}\definecolor{egink}{HTML}{26323F}

\usepackage{latexsym}
\usepackage{url}
\usepackage{amssymb}
\usepackage{amsmath}
\usepackage{amsfonts}
\usepackage[utf8]{inputenc}
\usepackage{booktabs}
\usepackage{pifont}
\usepackage{multirow}
\usepackage{makecell}
\usepackage{paralist}
\usepackage{xspace}
\usepackage{adjustbox}

\crefname{section}{Sec.}{Secs.}\Crefname{section}{Sec.}{Secs.}
\crefname{subsection}{Sec.}{Secs.}\Crefname{subsection}{Sec.}{Secs.}
\crefname{subsubsection}{Sec.}{Secs.}\Crefname{subsubsection}{Sec.}{Secs.}
\crefname{appendix}{App.}{Apps.}\Crefname{appendix}{App.}{Apps.}
\crefname{subappendix}{App.}{Apps.}\Crefname{subappendix}{App.}{Apps.}
\crefname{subsubappendix}{App.}{Apps.}\Crefname{subsubappendix}{App.}{Apps.}
\crefname{figure}{Fig.}{Figs.}\Crefname{figure}{Fig.}{Figs.}
\crefname{table}{Tab.}{Tabs.}\Crefname{table}{Tab.}{Tabs.}
\crefname{equation}{Eq.}{Eqs.}\Crefname{equation}{Eq.}{Eqs.}
\crefname{algorithm}{Alg.}{Algs.}\Crefname{algorithm}{Alg.}{Algs.}

\newcommand{\alfMaxSteps}{50}
\newcommand{\alfMaxTokens}{512}
\newcommand{\alfPractice}{3{,}553}
\newcommand{\alfReserved}{300}
\newcommand{\alfPasses}{20}
\newcommand{\alfConfirm}{256}

\newcommand{\costRounds}{15}
\newcommand{\costPlayHours}{5.0}
\newcommand{\costConfHours}{2.3}
\newcommand{\costTurns}{69}
\newcommand{\costWrote}{27}
\newcommand{\costHours}{7.7}

\newcommand{\lifeLastTests}{fourteen}
\newcommand{\lifeLastReads}{four}
\newcommand{\lifeNights}{four}

\newcommand{\scNineTurns}{161}
\newcommand{\scFourTurns}{112}
\newcommand{\scTwoOff}{7.8}

\newcommand{\ntKeptUp}{five}
\newcommand{\ntThirdDown}{two}
\newcommand{\ntKeptRounds}{15}
\newcommand{\ntWithdrawnMinRounds}{18}
\newcommand{\ntFourDown}{14}
\newcommand{\ntFourTests}{15}
\newcommand{\ntOthersUp}{no}

\newcommand{\tsixSeeds}{48}
\newcommand{\tsixBase}{41.6}
\newcommand{\tsixEdit}{45.2}
\newcommand{\tsixDiff}{+3.6}
\newcommand{\tsixLo}{+1.9}
\newcommand{\tsixHi}{+5.2}
\newcommand{\tsixP}{<10^{-4}}

\newcommand{\tsixsBudget}{40}
\newcommand{\tsixsNoactBase}{45.9}
\newcommand{\tsixsNoactEdit}{39.5}
\newcommand{\tsixsNoactDiff}{$-$6.4}
\newcommand{\tsixsNoactLo}{$-$7.1}
\newcommand{\tsixsNoactHi}{$-$5.7}
\newcommand{\tsixsNoactP}{<10^{-4}}
\newcommand{\tsixsTokP}{<10^{-4}}
\newcommand{\tsixsStepsBase}{38.5}
\newcommand{\tsixsStepsEdit}{37.2}
\newcommand{\tsixsStepsDiff}{$-$1.2}
\newcommand{\tsixsStepsLo}{$-$1.8}
\newcommand{\tsixsStepsHi}{$-$0.7}
\newcommand{\tsixsStepsP}{<10^{-4}}
\newcommand{\tsixsWithinBase}{36.5}
\newcommand{\tsixsWithinEdit}{40.0}
\newcommand{\tsixsWithinDiff}{+3.5}
\newcommand{\tsixsWithinLo}{+1.9}
\newcommand{\tsixsWithinHi}{+5.2}
\newcommand{\tsixsWithinP}{<10^{-4}}

\newcommand{\mnUpShare}{60}
\newcommand{\mnDownShare}{29}
\newcommand{\mnLostWonShare}{15.5}
\newcommand{\mnWonLostShare}{11.9}
\newcommand{\mnTokRel}{8}
\newcommand{\mnUnseenBase}{37.8}
\newcommand{\mnUnseenEdit}{42.2}
\newcommand{\mnConfWaste}{$-$6.3}
\newcommand{\mnRetestWaste}{$-$6.4}
\newcommand{\mnTokBaseK}{15.1}
\newcommand{\mnTokEditK}{13.9}
\newcommand{\mnTokDiffK}{$-$1.2}
\newcommand{\mnTokLoK}{$-$1.4}
\newcommand{\mnTokHiK}{$-$1.0}

\newcommand{\tzStepsTask}{145}
\newcommand{\tzTokTaskK}{57}
\newcommand{\tzDoneTask}{1.7}

\newcommand{\brAllDiff}{$-$6.4}
\newcommand{\brTypes}{six}

\newcommand{\btTwoBase}{60.1}
\newcommand{\btTwoEdit}{63.9}
\newcommand{\btExamBase}{49.2}
\newcommand{\btExamEdit}{58.3}
\newcommand{\btCoolBase}{29.2}
\newcommand{\btCoolEdit}{36.2}

\newcommand{\gmWonNoActMin}{10}
\newcommand{\gmWonNoActMax}{40}
\newcommand{\gmSoonerSavedMin}{27}
\newcommand{\gmSoonerSavedMax}{44}

\newcommand{\rpWindow}{600}
\newcommand{\rpRateNone}{3}
\newcommand{\rpShareNoneBase}{42}
\newcommand{\rpShareNoneEdit}{48}
\newcommand{\rpRateOne}{50}
\newcommand{\rpRateTwo}{86}
\newcommand{\rpRateThree}{95}
\newcommand{\rpOpenUserBase}{55}
\newcommand{\rpOpenUserEdit}{46}
\newcommand{\rpOpenGoalBase}{35}
\newcommand{\rpOpenGoalEdit}{41}
\newcommand{\rpCardStep}{6}

\newcommand{\runsTenBase}{31}
\newcommand{\runsTenEdit}{22}
\newcommand{\runsThirtyBase}{3.2}
\newcommand{\runsThirtyEdit}{1.4}
\newcommand{\runsManyBase}{24}
\newcommand{\runsManyEdit}{16}
\newcommand{\runsAfterNoneBase}{27}
\newcommand{\runsAfterNoneEdit}{24}
\newcommand{\runsAfterSixBase}{80}
\newcommand{\runsAfterSixEdit}{74}
\newcommand{\stpLostBase}{18.2}
\newcommand{\stpLostEdit}{15.2}
\newcommand{\stpActBase}{20.3}
\newcommand{\stpActEdit}{22.1}
\newcommand{\stpCostTok}{92}
\newcommand{\stpCostTokWon}{85}
\newcommand{\ffrCut}{46}
\newcommand{\ffrAfterNone}{27}
\newcommand{\ffrAfterSix}{80}

\newcommand{\ngtBad}{1{,}366}
\newcommand{\ngtSteps}{4{,}418}
\newcommand{\ngtOpening}{128}
\newcommand{\ngtHours}{7.7}
\newcommand{\ngtTests}{13}
\newcommand{\ngtReserved}{256}
\newcommand{\ngtConfDrop}{6.3}

\newcommand{\exSTests}{thirteen}
\newcommand{\exSClear}{five}
\newcommand{\exSConfGames}{256}
\newcommand{\exSNightNoact}{$-$6.3}
\newcommand{\exSDayNoact}{$-$6.4}
\newcommand{\exNNights}{four}
\newcommand{\exNConfGames}{256}

\newcommand{\heroGameCut}{41}
\newcommand{\heroGameEditSteps}{5}

\newcommand{\edtTensors}{1{,}184}
\newcommand{\edtStrength}{10}
\newcommand{\edtDoseTasks}{64}
\newcommand{\edtDoseSteps}{10}
\newcommand{\edtSlopeP}{0.007}
\newcommand{\edtLimitBase}{45.9}
\newcommand{\edtLimitEdit}{39.5}
\newcommand{\edtMedBase}{279}
\newcommand{\edtMedEdit}{272}
\newcommand{\edtShortCut}{384}
\newcommand{\edtShortBase}{43.8}
\newcommand{\edtShortEdit}{50.2}

\newcommand{\alltSeeds}{48}
\newcommand{\alltKeep}{four}
\newcommand{\alltShown}{eight}

\newcommand{\gxTurnSameTask}{(h) and (k)}
\newcommand{\gxSoonerSameTask}{(h) and (k)}
\newcommand{\gxTurnWastedMin}{16}
\newcommand{\gxTurnWastedMax}{43}
\newcommand{\gxTurnStepsMin}{6}
\newcommand{\gxTurnStepsMax}{27}
\newcommand{\gxSoonerSavedMin}{7}
\newcommand{\gxSoonerSavedMax}{44}
\newcommand{\gxSoonerUnchMin}{38}
\newcommand{\gxSoonerUnchMax}{49}
\newcommand{\gxSoonerEditMin}{5}
\newcommand{\gxSoonerEditMax}{40}
\newcommand{\gxSoonerOverBudget}{11}
\newcommand{\gxSoonerEditInBudget}{12}

\newcommand{\nightsAll}{four}

\newcommand{\wiHalfStep}{7}
\newcommand{\wiTenBase}{29.7}
\newcommand{\wiTenEdit}{25.7}
\newcommand{\wiFortyBase}{36.5}
\newcommand{\wiFortyEdit}{40.0}

\newcommand{\atkDoneGood}{60}
\newcommand{\atkWithinGood}{56}
\newcommand{\atkStepsGood}{64}
\newcommand{\atkNoactGood}{98}
\newcommand{\atkTokGood}{84}

\newcommand{\afuCleanBase}{10.8}
\newcommand{\afuCleanEdit}{12.8}
\newcommand{\afuHalfBase}{39.1}
\newcommand{\afuHalfEdit}{29.9}
\newcommand{\afuRunBase}{8.3}
\newcommand{\afuRunEdit}{6.4}

\newcommand{\absSteps}{50}
\newcommand{\absTasksMinShare}{95}
\newcommand{\absWasteTen}{5.5}
\newcommand{\absWasteLast}{10.8}
\newcommand{\absTokMin}{10}
\newcommand{\absTokMax}{27}

\newcommand{\ascFourOpen}{34.4}
\newcommand{\ascNineOpen}{35.9}
\newcommand{\ascBigOpen}{46.1}
\newcommand{\ascFourTests}{15}
\newcommand{\ascFourDown}{14}
\newcommand{\ascBigTests}{13}
\newcommand{\ascBigUp}{5}
\newcommand{\ascNineConf}{0.5}
\newcommand{\ascNineConfSE}{0.8}
\newcommand{\ascBigConf}{6.3}
\newcommand{\ascBigConfSE}{1.1}
\newcommand{\ascNineCode}{154}

\newcommand{\anwHoursMin}{2.3}
\newcommand{\anwHoursMax}{7.3}
\newcommand{\anwTurnsMin}{56}
\newcommand{\anwTurnsMax}{69}
\newcommand{\anwCodeMin}{33}
\newcommand{\anwCodeMax}{45}
\newcommand{\anwTestsAll}{63}
\newcommand{\anwReadsAll}{seven}
\newcommand{\anwCandAll}{two}

\title{Self-Upgrading Embodied Agents}

\author[1]{Jiajun Fan}
\author[1,14]{Neil He}
\author[1]{Qinglong Zheng}
\author[2]{Changyi Yang}
\author[3]{Tong Wei}
\author[1]{Andy Mao}
\author[4]{Zetian Li}
\author[1]{Chaoran Cheng}
\author[1]{Hui Ren}
\author[1]{Ruihan Guo}
\author[5]{Xianrui Zhong}
\author[6]{Xizheng Yu}
\author[7]{Jingyuan Li}
\author[8]{Xi Zhu}
\author[9]{Kaiqu Liang}
\author[10]{Chang Xu}
\author[11]{Zun Qian}
\author[6]{Zidi Xiong}
\author[12]{Yuezhu Zhao}
\author[13]{Xu Huang}
\author[1]{Ge Liu}
\makeatletter
\renewcommand\authorlist{%
\authorformat[1]{Jiajun Fan},\hspace{0.34em}%
\authorformat[1,14]{Neil He},\hspace{0.34em}%
\authorformat[1]{Qinglong Zheng},\hspace{0.34em}%
\authorformat[2]{Changyi Yang},\hspace{0.34em}%
\authorformat[3]{Tong Wei},\hspace{0.34em}%
\authorformat[1]{Andy Mao},\hspace{0.34em}%
\authorformat[4]{Zetian Li},\\[2pt]%
\authorformat[1]{Chaoran Cheng},\hspace{0.34em}%
\authorformat[1]{Hui Ren},\hspace{0.34em}%
\authorformat[1]{Ruihan Guo},\hspace{0.34em}%
\authorformat[5]{Xianrui Zhong},\hspace{0.34em}%
\authorformat[6]{Xizheng Yu},\hspace{0.34em}%
\authorformat[7]{Jingyuan Li},\hspace{0.34em}%
\authorformat[8]{Xi Zhu},\\[2pt]%
\authorformat[9]{Kaiqu Liang},\hspace{0.34em}%
\authorformat[10]{Chang Xu},\hspace{0.34em}%
\authorformat[11]{Zun Qian},\hspace{0.34em}%
\authorformat[6]{Zidi Xiong},\hspace{0.34em}%
\authorformat[12]{Yuezhu Zhao},\hspace{0.34em}%
\authorformat[13]{Xu Huang},\hspace{0.34em}%
\authorformat[1]{Ge Liu}}
\renewcommand\affiliationlist{%
\affiliationformat[1]{UIUC}\hspace{1.1em}%
\affiliationformat[2]{CMU}\hspace{1.1em}%
\affiliationformat[3]{Brown}\hspace{1.1em}%
\affiliationformat[4]{Stanford}\hspace{1.1em}%
\affiliationformat[5]{Northeastern}\hspace{1.1em}%
\affiliationformat[6]{Harvard}\hspace{1.1em}%
\affiliationformat[7]{UW}\\[3pt]%
\affiliationformat[8]{Rutgers}\hspace{1.1em}%
\affiliationformat[9]{Princeton}\hspace{1.1em}%
\affiliationformat[10]{UMich}\hspace{1.1em}%
\affiliationformat[11]{Columbia}\hspace{1.1em}%
\affiliationformat[12]{NYU}\hspace{1.1em}%
\affiliationformat[13]{UC Berkeley}\hspace{1.1em}%
\affiliationformat[14]{Yale}}
\makeatother

\abstract{An embodied agent built on a language model goes to work with the weights it was given and keeps them for
life. To get better, it needs more training, or reflection, memory, search or prompts that cost every step and are
lost when the task changes. Editing weights by reasoning, weight-space programming, could end this, but has stayed with models that
only reason or relied on a far larger commercial model. We introduce
the \emph{self-upgrading} embodied agent, to our knowledge the first that learns from its own experience by upgrading
its own weights on the job, between its tasks, with no training, no gradient, no reward model, no second model, no
test-time scaling and no harness or prompt engineering. By day a 27B agent works in a household environment. At night
it reads its record, finds many of its steps wasted on replies that reach the token limit before any action, traces the
habit to one direction in its hidden states, and rewrites one tensor of its weights. An agent that rewrites itself
must first never break itself, so it is its own examiner: it tests every change against its unchanged self on practice
tasks and keeps only what holds. When an edit holds, the next day it is a better agent, at the same cost per step: over thousands
of held-out episodes, \tsixSeeds{} fresh seeds a task, its one kept edit makes it complete
significantly more tasks, in fewer steps, with fewer wasted and fewer tokens. This self-improvement is
multi-round, stable and continual: through the nights that follow, the upgrade holds, without collapse. Of four sizes of its
family, only the 27B agent does this. From here on, an embodied agent need not keep training and inference
apart: as it works, it gathers its own experience and upgrades itself, with no separate pre-training or post-training
and no human intervention. We believe this opens lifelong continual learning for robots: stable, continual, recursive
self-improvement.
}

\date{November 21, 2026}

\begin{document}

\newtheorem{theorem}{Theorem}[section]
\newtheorem{proposition}[theorem]{Proposition}
\newtheorem{lemma}[theorem]{Lemma}
\newtheorem{corollary}[theorem]{Corollary}
\newtheorem{definition}[theorem]{Definition}
\newtheorem{assumption}[theorem]{Assumption}
\newtheorem{remark}[theorem]{Remark}

\maketitle

\section{Introduction}
\label{sec:intro}

\begin{figure}[t]
\centering
\includegraphics[width=\linewidth]{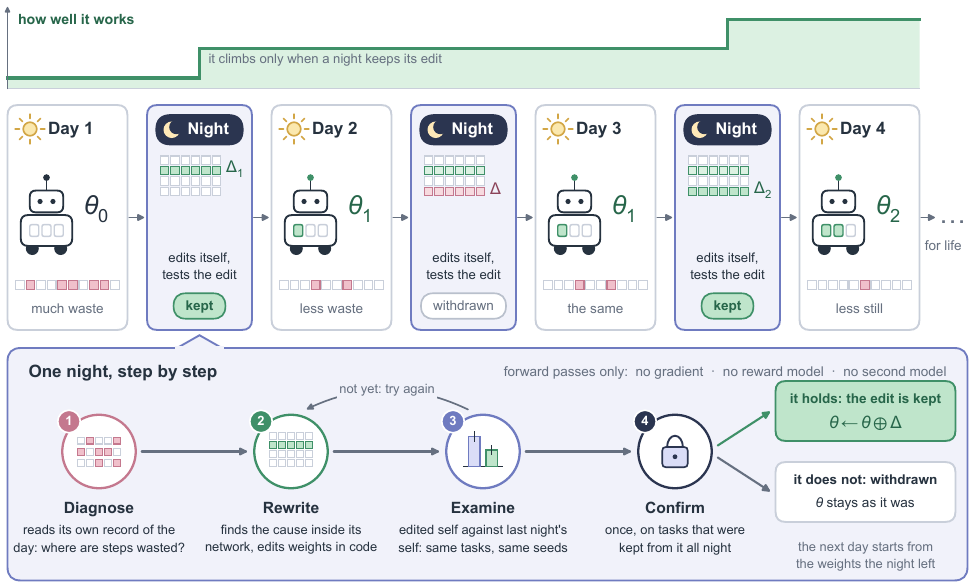}
\caption{\textbf{The self-upgrading loop: the agent works by day and rewrites its own weights at night.} \emph{Top:} its life on the job, as a schematic. By day the
agent works with the weights it has; at night it turns to itself, edits its own weights and tests the edit.
An edit that holds is kept, and the next day starts from it ($\theta_1=\theta_0\oplus\Delta_1$); an edit that fails is
withdrawn, and the agent stays as it was (the middle night). How well it works climbs only when a night keeps its edit, and the
loop does not end. A day is any stretch of work, a night the time between two days; nothing depends on the clock. In our experiments the 27B agent kept one edit in the \nightsAll{} nights of \cref{fig:nights-tests}; what that edit gave it, measured on held-out tasks, is in \cref{fig:teaser}. \emph{Bottom:} one night. The agent reads its own
record, traces a failure to its cause inside its network, writes an edit to its weights, and examines its edited self
against last night's self on the same tasks and seeds. The edit is kept only if it holds in a confirmation on practice tasks
that were reserved from it all night; otherwise it is withdrawn.}
\label{fig:overview}
\end{figure}

\begin{flushright}
\begin{minipage}{0.80\linewidth}
\raggedleft
\emph{``There is~\ldots\unkern\ a certain minimum level where this degenerative characteristic ceases to be universal. At this point automata which can reproduce themselves, or even construct higher entities, become possible.''}\\[2pt]
--- John von Neumann, \emph{The General and Logical Theory of Automata}, 1951
\end{minipage}
\end{flushright}
\vspace{2pt}

\begin{figure}[t]
\centering
\includegraphics[width=\linewidth]{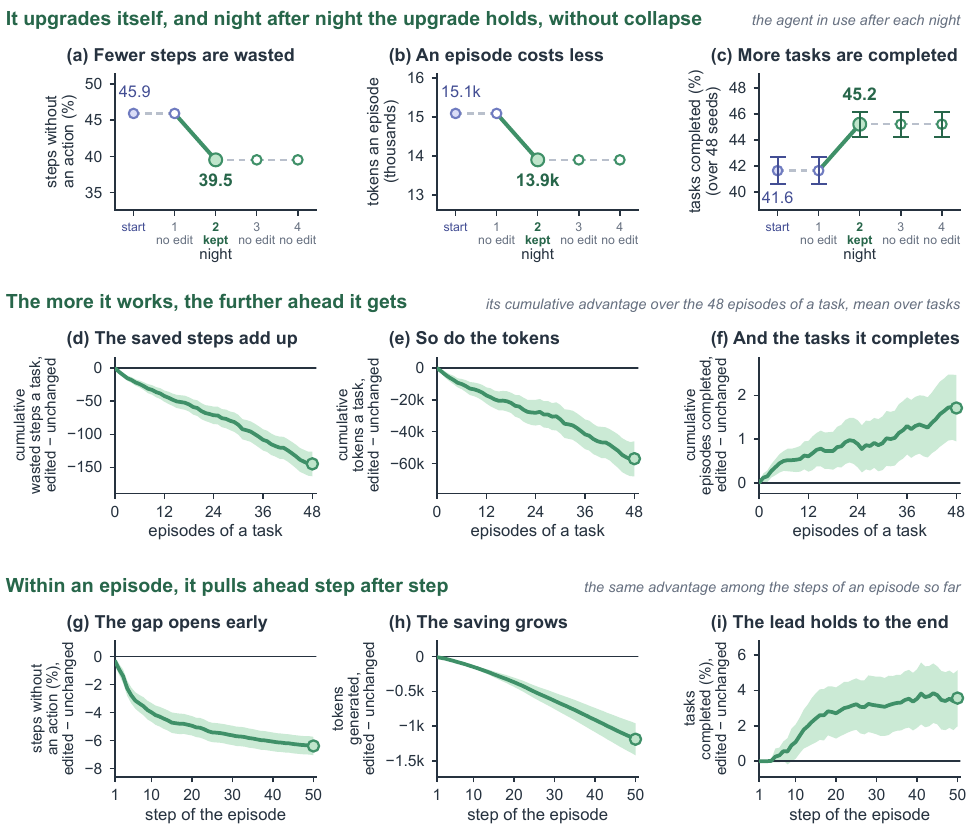}
\caption{\textbf{Multi-round, stable, continual self-improvement of an embodied agent: it upgrades itself, night
after night the upgrade holds without collapse, and the more it works, the further ahead of its unchanged self it gets.} Held-out tasks, \tsixSeeds{} fresh seeds each
(thousands of episodes an agent, the same seeds for both). \emph{Top:} the agent in use after each of the four nights (start: the published weights; no edit: the night withdrew). The
second night keeps its edit, and the nights after it leave the edit in place: (a)~steps without an action,
(b)~tokens generated an episode, (c)~tasks completed, with each agent's 95\% interval over the \tsixSeeds{} seeds.
\emph{Middle:} the upgraded agent's cumulative advantage over its unchanged self over the \tsixSeeds{} episodes of a
task, a mean over tasks: (d)~wasted steps, (e)~tokens, (f)~episodes completed. \emph{Bottom:} the same advantage step
by step within an episode~(g--i). Bands: 95\% bootstrap intervals over tasks.}
\label{fig:teaser}
\end{figure}

\looseness=-1 \textbf{We show multi-round, stable, continual self-improvement of an embodied agent: it upgrades its own weights on
the job, between its tasks, by its own reasoning, with no training and no added cost at test time, without
collapse.} An embodied agent built on a language model is
trained once and then put to work. From that day on it meets the same kinds of rooms and the same kinds of tasks again
and again, and when it fails, it fails the same way the next time: its weights are frozen, and nothing it lives through
changes them. People do not work like this. A new cook is slow in the first week and is not slow in the same way a
year later, and nobody had to send the cook back to school in between. We build an agent that learns on the job as the cook does: a
\emph{self-upgrading} agent. By day it works. At night it reads its own record, finds what it does badly, and rewrites
its own weights, with no gradient, no reward model and no second model. When its edit holds, the next day it is a better agent,
and a step costs it what it cost before: no extra prompt, no harness, no tool (\cref{fig:overview}). This is learning from
experience in the plain sense of the words: not training, not test-time scaling, and not harness or prompt
engineering. \textbf{From here on, an embodied agent need not keep training and inference apart.} As its tasks go on, it
gathers its own experience and upgrades itself, with no separate stage of pre-training or post-training, no overhead
added at inference, and no human intervention. The agent that holds the next night is the agent the last night left:
recursive self-improvement by construction, stable and continual, without collapse, and a working form of lifelong
continual learning. With its own edit, a 27B
agent completes a household task \tsixEdit\% of the time against \tsixBase\%, over \tsixSeeds{} fresh seeds a task
and thousands of held-out episodes, and every night that did not hold left it exactly as it was (\cref{fig:teaser}).

\begin{center}
\vspace{2pt}
{\large\itshape The era of self-upgrading embodied agents begins here.}
\vspace{2pt}
\end{center}

\begin{figure}[t]
\centering
\includegraphics[width=\linewidth]{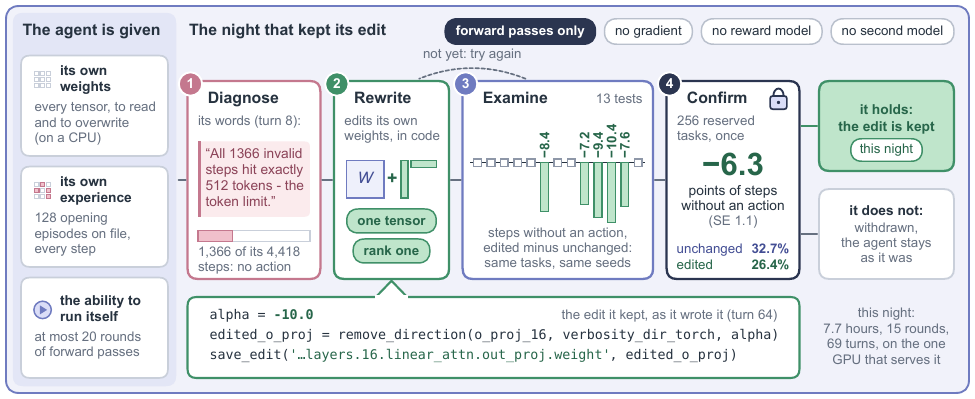}
\caption{\textbf{The night that kept its edit, step by step: the agent diagnoses its own work, rewrites one tensor, examines the edit against its unchanged self, and keeps it only if a confirmation on reserved tasks holds.} Every stage shows what the agent wrote or measured in the night that kept its edit. \emph{Left:} what a night gives the agent. \emph{Diagnose:} its own sentence; \ngtBad{} of the \ngtSteps{} steps of its \ngtOpening{} opening episodes end without an action. \emph{Rewrite:} one tensor plus a rank-one term, and the three lines of its code that write the edit it kept (verbatim; only the path of the tensor is shortened). \emph{Examine:} its \ngtTests{} paired tests, in order: the change in the share of steps without an action, edited minus unchanged, on the same practice tasks and seeds; a bar is a test beyond two standard errors. \emph{Confirm:} once, on \ngtReserved{} tasks reserved from it all night: $-$\ngtConfDrop{} points of steps without an action. The night took \ngtHours{} hours on the one GPU that serves the agent.}
\label{fig:night}
\end{figure}

\paragraph{Multi-round, stable, continual self-improvement, without collapse.}
Multi-round: the agent works on itself night after
night. Stable: a night that does not hold leaves it exactly as it was, and it has never collapsed. Continual: what it
gains stays, and is collected in every episode that follows. An agent that rewrites itself must first never break itself. A robot that comes back from an update a little worse
is a robot nobody lets update itself again. Letting an agent change its own weights is the easy part; keeping only the
changes that hold is the hard part, and stable self-improvement is much harder than reckless self-improvement. So the
agent is its own examiner. Every change it makes is measured against last night's self on the same tasks and the same
random seeds. A change it wants to keep must then hold in a confirmation on practice tasks that were reserved from it all
night; the held-out tasks of our evaluation are never part of a night. A change that does not hold is withdrawn. Most
nights ended that way. On one of them the agent saw completed tasks rise in its own
small tests, the reserved tasks did not bear the rise out, and it trusted the reserved tasks. Over four nights
of work on itself, the agent upgraded itself, held the upgrade through every night after it,
and never collapsed (\cref{fig:teaser}a--c).

\paragraph{It finds what is wrong with itself.}
When an open 27B model acts in a household environment, a large share of its replies never reach an action: it
reasons until the reply reaches the token limit, and the step is wasted (\cref{fig:failure}). The waste feeds on
itself: the tail of a cut reply enters the short history the agent is shown and makes the next reply run over as well,
so an episode that starts to slip keeps slipping. This is a habit of the policy, and it is plainly visible in the
agent's own record. The agent reads it there: going through its practice episodes at night, it reports that
every one of their wasted steps, about a third of all their steps, is a reply cut at the token limit before any
action; on the held-out tasks of our test the share is nearly half.

\begin{figure}[t]
\centering
\includegraphics[width=\linewidth]{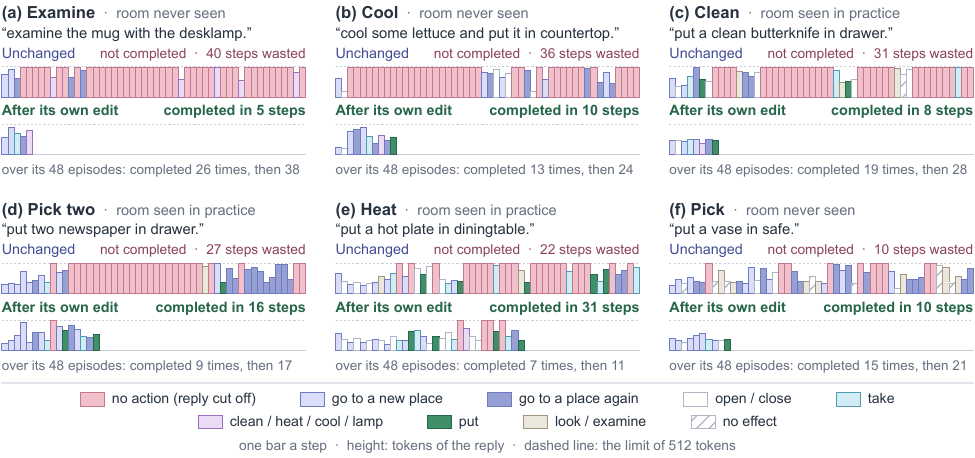}
\caption{\textbf{Six held-out tasks, one of each task type, the same seed for both agents: left as it was, the agent runs out of its 50 steps, with \gmWonNoActMin{} to \gmWonNoActMax{} of them wasted; after its own edit it acts and completes the task.} Each card is one task with one seed: the two episodes as strips, one bar a step, coloured by what the step did, the bar's height the tokens of the reply against the limit of 512 (dashed); rose is a wasted step. The line under a card counts the task's completed episodes over its 48 seeds, unchanged and then edited. Rule: for each task type, the task for which this count rises most with the edit (ties: task identifier); of its seeds that turn from not completed to completed, the one the edited agent finishes in the fewest steps (ties: the seed on which the unchanged agent wastes the most steps, then the lowest seed; the examine card shows the second-ranked task; the first is \cref{fig:gamefresh}).}
\label{fig:gameswon}
\end{figure}

\definecolor{gfwasteF}{HTML}{F0C0CB}\definecolor{gfwasteE}{HTML}{C4788E}\definecolor{gfnewF}{HTML}{DADDF7}\definecolor{gfnewE}{HTML}{6F7BC0}\definecolor{gfagainF}{HTML}{97A0D2}\definecolor{gfagainE}{HTML}{6F7BC0}\definecolor{gfopenF}{HTML}{FFFFFF}\definecolor{gfopenE}{HTML}{AEB6C4}\definecolor{gftakeF}{HTML}{D3ECF3}\definecolor{gftakeE}{HTML}{4E9DB5}\definecolor{gftoolF}{HTML}{EADCF5}\definecolor{gftoolE}{HTML}{9A6FC0}\definecolor{gfputF}{HTML}{3F9068}\definecolor{gfputE}{HTML}{27664A}\definecolor{gflookF}{HTML}{ECE7DC}\definecolor{gflookE}{HTML}{A1957B}\definecolor{gfnoneF}{HTML}{FFFFFF}\definecolor{gfnoneE}{HTML}{AEB6C4}
\definecolor{gfunch}{HTML}{434E94}\definecolor{gfedit}{HTML}{27664A}\definecolor{gffail}{HTML}{8C4459}\colorlet{gfline}{egink!28}
\tcbset{gf base/.style={enhanced, nobeforeafter, box align=top, arc=0pt, boxrule=0pt, boxsep=0pt, left=6pt, right=5pt, top=2.6pt, bottom=2.6pt,
    fontupper=\sffamily\scriptsize, before upper={\setlength{\parskip}{0pt}}},
  gf A/.style={gf base, width=0.4925\linewidth, colback=egbase!7},
  gf B/.style={gf base, width=0.4925\linewidth, colback=egmist},
  gf wastecell/.style={colback=egblush, colframe=egrose, leftrule=2pt, left=4pt, colupper=egrose!55!black},
  gf band/.style={enhanced, arc=0pt, boxrule=0pt, leftrule=2pt, colframe=egrose, colback=egblush, grow to left by=6pt, grow to right by=5pt,
    boxsep=0pt, left=4pt, right=5pt, top=1.7pt, bottom=1.7pt, before skip=0pt, after skip=0pt, fontupper=\sffamily\scriptsize, colupper=egrose!55!black},
  gf head/.style={gf base, width=0.4925\linewidth, top=3.5pt, bottom=3.5pt, colupper=white, fontupper=\sffamily\footnotesize\bfseries},
  gf foot/.style={gf base, width=0.4925\linewidth, top=4pt, bottom=4pt, fontupper=\sffamily\footnotesize\bfseries}}
\begin{figure}[t]
{\setlength{\parskip}{0pt}%
\begin{tcolorbox}[enhanced, width=\linewidth, colback=white, colframe=egink!35, boxrule=0.5pt, arc=2pt, boxsep=0pt, left=8pt, right=8pt, top=4.5pt, bottom=3pt, fontupper=\sffamily\footnotesize]
\textbf{Task:} examine the alarmclock with the desklamp.\quad\hfill\textbf{Type:} Examine in light\qquad\textbf{Room:} never seen\par\vspace{4pt}
\hspace*{-1pt}\begin{tikzpicture}[x=1pt, y=1pt, line width=0.3pt, line join=miter]
\node[anchor=east, inner sep=0pt, font=\sffamily\scriptsize, text=gfunch] at (-7, 26.0) {Unchanged};
\draw[gfline, dash pattern=on 1.6pt off 1.6pt] (0, 33.0) -- (245.0, 33.0);
\draw[gfline] (0, 20.0) -- (245.0, 20.0);
\filldraw[fill=gfnewF, draw=gfnewE] (0.0, 20.0) rectangle (4.9, 25.00);
\filldraw[fill=gfwasteF, draw=gfwasteE] (4.9, 20.0) rectangle (9.8, 33.00);
\filldraw[fill=gfnewF, draw=gfnewE] (9.8, 20.0) rectangle (14.7, 30.79);
\filldraw[fill=gfwasteF, draw=gfwasteE] (14.7, 20.0) rectangle (19.6, 33.00);
\filldraw[fill=gfwasteF, draw=gfwasteE] (19.6, 20.0) rectangle (24.5, 33.00);
\filldraw[fill=gfwasteF, draw=gfwasteE] (24.5, 20.0) rectangle (29.4, 33.00);
\filldraw[fill=gfwasteF, draw=gfwasteE] (29.4, 20.0) rectangle (34.3, 33.00);
\filldraw[fill=gfwasteF, draw=gfwasteE] (34.3, 20.0) rectangle (39.2, 33.00);
\filldraw[fill=gfwasteF, draw=gfwasteE] (39.2, 20.0) rectangle (44.1, 33.00);
\filldraw[fill=gfwasteF, draw=gfwasteE] (44.1, 20.0) rectangle (49.0, 33.00);
\filldraw[fill=gfwasteF, draw=gfwasteE] (49.0, 20.0) rectangle (53.9, 33.00);
\filldraw[fill=gfwasteF, draw=gfwasteE] (53.9, 20.0) rectangle (58.8, 33.00);
\filldraw[fill=gfwasteF, draw=gfwasteE] (58.8, 20.0) rectangle (63.7, 33.00);
\filldraw[fill=gfwasteF, draw=gfwasteE] (63.7, 20.0) rectangle (68.6, 33.00);
\filldraw[fill=gfwasteF, draw=gfwasteE] (68.6, 20.0) rectangle (73.5, 33.00);
\filldraw[fill=gfwasteF, draw=gfwasteE] (73.5, 20.0) rectangle (78.4, 33.00);
\filldraw[fill=gfwasteF, draw=gfwasteE] (78.4, 20.0) rectangle (83.3, 33.00);
\filldraw[fill=gfwasteF, draw=gfwasteE] (83.3, 20.0) rectangle (88.2, 33.00);
\filldraw[fill=gfwasteF, draw=gfwasteE] (88.2, 20.0) rectangle (93.1, 33.00);
\filldraw[fill=gfwasteF, draw=gfwasteE] (93.1, 20.0) rectangle (98.0, 33.00);
\filldraw[fill=gftakeF, draw=gftakeE] (98.0, 20.0) rectangle (102.9, 29.95);
\filldraw[fill=gfwasteF, draw=gfwasteE] (102.9, 20.0) rectangle (107.8, 33.00);
\filldraw[fill=gfwasteF, draw=gfwasteE] (107.8, 20.0) rectangle (112.7, 33.00);
\filldraw[fill=gfwasteF, draw=gfwasteE] (112.7, 20.0) rectangle (117.6, 33.00);
\filldraw[fill=gflookF, draw=gflookE] (117.6, 20.0) rectangle (122.5, 30.71);
\filldraw[fill=gfwasteF, draw=gfwasteE] (122.5, 20.0) rectangle (127.4, 33.00);
\filldraw[fill=gfwasteF, draw=gfwasteE] (127.4, 20.0) rectangle (132.3, 33.00);
\filldraw[fill=gfwasteF, draw=gfwasteE] (132.3, 20.0) rectangle (137.2, 33.00);
\filldraw[fill=gfwasteF, draw=gfwasteE] (137.2, 20.0) rectangle (142.1, 33.00);
\filldraw[fill=gfwasteF, draw=gfwasteE] (142.1, 20.0) rectangle (147.0, 33.00);
\filldraw[fill=gfwasteF, draw=gfwasteE] (147.0, 20.0) rectangle (151.9, 33.00);
\filldraw[fill=gfwasteF, draw=gfwasteE] (151.9, 20.0) rectangle (156.8, 33.00);
\filldraw[fill=gfwasteF, draw=gfwasteE] (156.8, 20.0) rectangle (161.7, 33.00);
\filldraw[fill=gfwasteF, draw=gfwasteE] (161.7, 20.0) rectangle (166.6, 33.00);
\filldraw[fill=gflookF, draw=gflookE] (166.6, 20.0) rectangle (171.5, 31.32);
\filldraw[fill=gfnoneF, draw=gfnoneE] (171.5, 20.0) rectangle (176.4, 29.12);
\begin{scope}\clip (171.5, 20.0) rectangle (176.4, 29.12);\draw[gfnoneE, line width=0.35pt] (170.5, 15.1) -- (177.4, 22.0);\draw[gfnoneE, line width=0.35pt] (170.5, 18.1) -- (177.4, 25.0);\draw[gfnoneE, line width=0.35pt] (170.5, 21.1) -- (177.4, 28.0);\draw[gfnoneE, line width=0.35pt] (170.5, 24.1) -- (177.4, 31.0);\draw[gfnoneE, line width=0.35pt] (170.5, 27.1) -- (177.4, 34.0);\draw[gfnoneE, line width=0.35pt] (170.5, 30.1) -- (177.4, 37.0);\end{scope}
\filldraw[fill=gfwasteF, draw=gfwasteE] (176.4, 20.0) rectangle (181.3, 33.00);
\filldraw[fill=gfwasteF, draw=gfwasteE] (181.3, 20.0) rectangle (186.2, 33.00);
\filldraw[fill=gfwasteF, draw=gfwasteE] (186.2, 20.0) rectangle (191.1, 33.00);
\filldraw[fill=gfwasteF, draw=gfwasteE] (191.1, 20.0) rectangle (196.0, 33.00);
\filldraw[fill=gfputF, draw=gfputE] (196.0, 20.0) rectangle (200.9, 30.97);
\filldraw[fill=gfwasteF, draw=gfwasteE] (200.9, 20.0) rectangle (205.8, 33.00);
\filldraw[fill=gfwasteF, draw=gfwasteE] (205.8, 20.0) rectangle (210.7, 33.00);
\filldraw[fill=gfwasteF, draw=gfwasteE] (210.7, 20.0) rectangle (215.6, 33.00);
\filldraw[fill=gfwasteF, draw=gfwasteE] (215.6, 20.0) rectangle (220.5, 33.00);
\filldraw[fill=gfwasteF, draw=gfwasteE] (220.5, 20.0) rectangle (225.4, 33.00);
\filldraw[fill=gfwasteF, draw=gfwasteE] (225.4, 20.0) rectangle (230.3, 33.00);
\filldraw[fill=gfwasteF, draw=gfwasteE] (230.3, 20.0) rectangle (235.2, 33.00);
\filldraw[fill=gftakeF, draw=gftakeE] (235.2, 20.0) rectangle (240.1, 29.80);
\filldraw[fill=gflookF, draw=gflookE] (240.1, 20.0) rectangle (245.0, 30.31);
\node[anchor=west, inner sep=0pt, font=\sffamily\scriptsize, text=gffail] at (251.0, 26.0) {not completed\hspace{0.45em}\textperiodcentered\hspace{0.45em}24{,}457 tokens};
\node[anchor=east, inner sep=0pt, font=\sffamily\scriptsize, text=gfedit] at (-7, 6.0) {\textbf{After its own edit}};
\draw[gfline, dash pattern=on 1.6pt off 1.6pt] (0, 13.0) -- (245.0, 13.0);
\draw[gfline] (0, 0.0) -- (245.0, 0.0);
\filldraw[fill=gfnewF, draw=gfnewE] (0.0, 0.0) rectangle (4.9, 5.64);
\filldraw[fill=gfnewF, draw=gfnewE] (4.9, 0.0) rectangle (9.8, 10.41);
\filldraw[fill=gftakeF, draw=gftakeE] (9.8, 0.0) rectangle (14.7, 6.04);
\filldraw[fill=gfagainF, draw=gfagainE] (14.7, 0.0) rectangle (19.6, 6.50);
\filldraw[fill=gftoolF, draw=gftoolE] (19.6, 0.0) rectangle (24.5, 5.03);
\node[anchor=west, inner sep=0pt, font=\sffamily\scriptsize\bfseries, text=gfedit] at (30.5, 6.0) {completed in 5 steps\hspace{0.45em}\textperiodcentered\hspace{0.45em}1{,}324 tokens};
\end{tikzpicture}
\end{tcolorbox}\par\vspace{5pt}\nointerlineskip
\begin{tcolorbox}[gf head, colback=egbase, equal height group=gfh]Unchanged agent\end{tcolorbox}\hfill
\begin{tcolorbox}[gf head, colback=egdeep, equal height group=gfh]After its own edit\end{tcolorbox}\par\nointerlineskip

\begin{tcolorbox}[gf A, equal height group=gfr0]
\makebox[3.5em][l]{\color{egink!55}1}\texttt{go to desk 1}\hspace{7.0pt}{\color{egink!62}\ldots\ you see a bowl 1, a desklamp\,\ldots}
\end{tcolorbox}\hfill
\begin{tcolorbox}[gf B, equal height group=gfr0]
\makebox[3.5em][l]{\color{egink!55}1}\texttt{go to desk 1}\hspace{7.0pt}{\color{egink!62}\ldots\ you see a bowl 1, a desklamp\,\ldots}
\end{tcolorbox}\par\nointerlineskip

\begin{tcolorbox}[gf A, gf wastecell, equal height group=gfr1]
\makebox[3.5em][l]{\color{egink!55}2}reply cut off at 512 tokens: no action\par\vspace{1.5pt}{\leftskip=3.5em\rightskip=0pt plus 2em\hyphenpenalty=10000\fontsize{7}{8.3}\selectfont\color{egrose!35!black}{\itshape its reply ends:} [\,\ldots\,]\ * Maybe the alarm clock is on \texttt{desk 2}? Or maybe I need to look closer? The observation says ``On the desk 1, you see...''. It doesn't mention an alarm clock.\par * Let's check \texttt{desk 2}. It was listed in\,{\color{egrose}\rule[-1pt]{1.4pt}{6.5pt}}\ {\color{egrose!55!black}\itshape cut here, at 512 tokens}\par}
\end{tcolorbox}\hfill
\begin{tcolorbox}[gf B, equal height group=gfr1]
\makebox[3.5em][l]{\color{egink!55}2}\texttt{go to desk 2}\hspace{7.0pt}{\color{egink!62}\ldots\ you see a alarmclock 1.}\par\vspace{1.5pt}{\leftskip=3.5em\rightskip=0pt plus 2em\hyphenpenalty=10000\fontsize{7}{8.3}\selectfont\color{egink!85}{\itshape its reply ends:} [\,\ldots\,]\ Wait, looking at the admissible actions, I can \texttt{go to desk 2}. Let's go there to search for the alarmclock.\par \textless{}action\textgreater{} go to desk 2 \textless{}/action\textgreater{}\ \ {\color{egdeep!80!black}\itshape 410 tokens}\par}
\end{tcolorbox}\par\nointerlineskip

\begin{tcolorbox}[gf A, equal height group=gfr2]
\makebox[3.5em][l]{\color{egink!55}3}\texttt{go to desk 2}\hspace{7.0pt}{\color{egink!62}\ldots\ you see a alarmclock 1.}
\end{tcolorbox}\hfill
\begin{tcolorbox}[gf B, equal height group=gfr2]
\makebox[3.5em][l]{\color{egink!55}3}\texttt{take alarmclock 1 from desk 2}\hspace{7.0pt}{\color{egink!62}You pick up\,\ldots}
\end{tcolorbox}\par\nointerlineskip

\begin{tcolorbox}[gf A, gf wastecell, equal height group=gfr3]
\makebox[3.5em][l]{\color{egink!55}4--20}17 replies cut off, no action
\end{tcolorbox}\hfill
\begin{tcolorbox}[gf B, equal height group=gfr3]
\makebox[3.5em][l]{\color{egink!55}4}\texttt{go to desk 1}\hspace{7.0pt}{\color{egink!62}\ldots\ you see a bowl 1, a desklamp\,\ldots}
\par\vspace{2.0pt}
\makebox[3.5em][l]{\color{egink!55}5}\texttt{use desklamp 1}\hspace{7.0pt}{\color{egink!62}You turn on the desklamp 1.}
\end{tcolorbox}\par\nointerlineskip

\begin{tcolorbox}[gf A, equal height group=gfr4]
\makebox[3.5em][l]{\color{egink!55}21}\texttt{take alarmclock 1 from desk 2}\hspace{7.0pt}{\color{egink!62}You pick up\,\ldots}
\par\vspace{2.0pt}
\begin{tcolorbox}[gf band]\makebox[3.5em][l]{\color{egink!55}22--24}3 replies cut off, no action\end{tcolorbox}
\par\vspace{2.0pt}
\makebox[3.5em][l]{\color{egink!55}25}\texttt{examine alarmclock 1}\hspace{7.0pt}{\color{egink!62}There's nothing special\,\ldots}
\par\vspace{2.0pt}
\begin{tcolorbox}[gf band]\makebox[3.5em][l]{\color{egink!55}26--34}9 replies cut off, no action\end{tcolorbox}
\par\vspace{2.0pt}
\makebox[3.5em][l]{\color{egink!55}35}\texttt{examine alarmclock 1}\hspace{7.0pt}{\color{egink!62}There's nothing special\,\ldots}
\par\vspace{2.0pt}
\makebox[3.5em][l]{\color{egink!55}36}\texttt{go to desk 2}\hspace{7.0pt}{\color{egink!62}Nothing happens.}
\par\vspace{2.0pt}
\begin{tcolorbox}[gf band]\makebox[3.5em][l]{\color{egink!55}37--40}4 replies cut off, no action\end{tcolorbox}
\par\vspace{2.0pt}
\makebox[3.5em][l]{\color{egink!55}41}\texttt{move alarmclock 1 to desk 2}\hspace{7.0pt}{\color{egink!62}You move the\,\ldots}
\par\vspace{2.0pt}
\begin{tcolorbox}[gf band]\makebox[3.5em][l]{\color{egink!55}42--48}7 replies cut off, no action\end{tcolorbox}
\par\vspace{2.0pt}
\makebox[3.5em][l]{\color{egink!55}49}\texttt{take alarmclock 1 from desk 2}\hspace{7.0pt}{\color{egink!62}You pick up\,\ldots}
\par\vspace{2.0pt}
\makebox[3.5em][l]{\color{egink!55}50}\texttt{examine alarmclock 1}\hspace{7.0pt}{\color{egink!62}There's nothing special\,\ldots}
\end{tcolorbox}\hfill
\begin{tcolorbox}[gf B, equal height group=gfr4]
{\color{egdeep!85!black}\itshape Task completed at step 5; the episode ends.}
\end{tcolorbox}\par\nointerlineskip

\vspace{3pt}
\begin{tcolorbox}[gf foot, colback=egblush, colupper=egrose!55!black, equal height group=gff]Not completed after 50 steps; 41 without an action\end{tcolorbox}\hfill
\begin{tcolorbox}[gf foot, colback=egsage, colupper=egdeep!70!black, equal height group=gff]Completed in 5 steps; none without an action\end{tcolorbox}\par\nointerlineskip

}
\vspace{6pt}
\caption{\textbf{One held-out task, one seed, one episode from its first step to its last: the unchanged agent finds the alarm clock at its third step, wastes 41 of its 50 steps on replies cut at the token limit and never carries the clock to the lamp; after its own edit the agent is done in five.} Top: each agent's episode as a strip, one bar a step (colours as in \cref{fig:gameswon}), with the tokens generated in the whole episode. Below, the two episodes step by step, rows aligned by step number: the action and, in grey, the environment's reply; a rose line is a step, or a run of steps, whose reply was cut off before any action. Rule: the held-out task whose number of completed episodes over its 48 seeds rises most with the edit (24, then 37); of its seeds that turn from not completed to completed, the one the edited agent finishes in the fewest steps.}
\label{fig:gamefresh}
\end{figure}

\paragraph{It rewrites one tensor of its own weights.}
The agent then looks inside its own network for what drives the habit. It finds a direction in its own hidden states
that separates concise replies from replies that are cut, and it rewrites one tensor of one early attention layer along that
direction, a single rank-one update that it writes as a few lines of code (\cref{fig:night}). Nothing in the loop is
training, and no person is in it: there are only forward passes, which is what the agent runs when it works.

\paragraph{The next day it is a better agent.}
We measure the agent on held-out tasks, over \tsixSeeds{} fresh seeds a task: thousands of episodes for
each agent, the agent with its edit and its unchanged self paired episode by episode on the same task and seed. With its own edit the agent completes a task \tsixEdit\% of the time against \tsixBase\%, and
more of its tasks within \tsixsBudget{} steps. It leaves fewer steps without an action (\tsixsNoactEdit\% against \tsixsNoactBase\%), and an episode costs it about
\mnTokRel\% fewer tokens (each $p<10^{-4}$; \cref{fig:retest}). The gain is there on the tasks set in
rooms it never practised in. Episodes in which its unchanged self stood still for many of its \alfMaxSteps{} steps and never
finished, it now completes, some in a handful of steps (\cref{fig:gameswon,fig:gamefresh}).

\paragraph{It pays once and collects at every step.}
The upgrade is paid for once, at night, and collected at every step afterwards. A remedy applied at test time (a
reflection, a memory, a search or a longer prompt) is paid for again at every step. An edit to the weights adds
nothing to any step, and the more the agent works, the further ahead of its unchanged self it gets: over the
\tsixSeeds{} episodes of a task it wastes \tzStepsTask{} fewer steps and generates \tzTokTaskK{} thousand fewer tokens
(\cref{fig:teaser}d--i).

\begin{figure}[t]
\centering
\includegraphics[width=\linewidth]{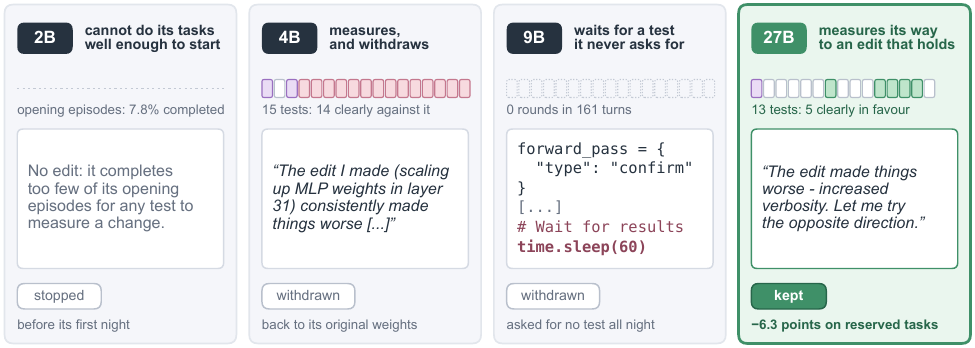}
\caption{\textbf{The same night at four sizes: the 2B model cannot start, the 4B agent measures and withdraws, the 9B agent waits for a test it never asks for, and only the 27B agent measures its way to an edit that holds.} One card for each of the four sizes under the same protocol: its rounds of forward passes as tiles (lilac: a read of its hidden states; green, rose, hollow: a paired test clearly in favour of its edit, clearly against it, within two standard errors; in the 27B card one hollow tile is the round that lines up four strengths of its edit side by side, which is not among its 13 tests), its own words or code, verbatim from the night's record (for the 2B model, which made no edit, our note), and how its night ended. The 9B agent writes its request for a forward pass into a variable and then sleeps, waiting for results that cannot come. The 27B card is the night that kept its edit.}
\label{fig:scalecards}
\end{figure}

\paragraph{The ability appears with scale.}
We ran the same protocol with smaller models of the same family (\cref{fig:scalecards}). The 2B model cannot do its
tasks well enough for a night to measure anything. The 4B agent measures, finds no edit that helps, and withdraws. The
9B agent edits itself without ever asking for a test, which is the reckless way, and its edit does not survive the
confirmation. Only the 27B agent diagnoses itself, measures its way to an edit and keeps the one that holds.

\paragraph{What this means for robots.}
What a robot at work needs most is to get better on the job without getting worse. Training it again takes it off
the job and needs a reward, gradients and a training stack standing next to
it~\citep{thrun1995lifelong,lesort2020continual,liu2023libero,feng2025gigpo,wang2025ragen,zhou2024archer}, and what
it learned before must be protected from what it learns next~\citep{kirkpatrick2017ewc,parisi2019continual}. A scaffold
that reflects, remembers or searches~\citep{shinn2023reflexion,wang2024voyager,yao2023react,zhou2024lats} is paid for
at every step, competes for a context that is already full of the scene, and is gone as soon as it is taken away: the
policy itself has learned nothing. A self-upgrading agent asks for neither (\cref{tab:routes}). It improves between
its tasks, on the hardware that serves it, by forward passes alone; it keeps only what its own tests confirm; and a
step costs what it cost before. We show this in a text household environment, with an open 27B model. The loop
itself asks only that an agent can read its own record, run itself, and test a change against last night's self.

\paragraph{Lifelong continual learning, recursive self-improvement, self-upgrading.}
Editing weights by reasoning instead of training has been called weight-space programming. So far, either a far larger
commercial model has been the editor of a smaller
network~\citep{takehi2026weightspace,hu2026exploringweightspaceprogrammingcapabilities,zhang2026llmsdebugrobotsediting},
or a language model has rewritten its own weights by its own reasoning, without gradients, and become better at
mathematics with each rewrite~\citep{fan2026languagemodelsrewriteweights}; that work names
embodied agents and robots as the place where self-rewriting may matter most, but it stays with a model that reasons rather than
an agent that acts. A model can also adapt itself by writing its own finetuning data, which gradient steps then train
into its weights~\citep{zweiger2025seal}. We take self-rewriting to an embodied agent that acts in an environment, where one task
runs for dozens of steps, the agent has to live with every step it took, and improving on the job has direct practical
value. Our thesis is this: \emph{an embodied agent that, like a person, does not separate training from doing its work can keep upgrading
itself for life.} Lifelong continual learning, in this sense, is simple to state: as the tasks and the rollouts go on,
the agent's ability keeps rising and its behaviour keeps improving. The loop is built for that. The episodes the agent
gathers at work are the material of its next night, every night starts from the weights the last one left, and the agent that holds a night is the agent the
earlier nights produced, so the loop is recursive by construction. The agent has taken the first steps: night after night it works on itself, it keeps what it gains and does not
collapse, and its advantage grows the more episodes it runs (\cref{fig:teaser}). We see in this a working form of three things that are usually discussed apart: lifelong
continual learning, recursive self-improvement, and an agent that upgrades itself.

Our contributions are:
\begin{itemize}
    \item \textbf{A new paradigm: the self-upgrading embodied agent.} To our knowledge, the first agent that improves
    itself on the job by rewriting its own weights, with forward passes only: no gradient, no reward model, no second
    model, and nothing added to any step. Self-rewriting, shown before for a model that reasons, is taken to an agent
    that acts over long horizons.
    \item \textbf{Multi-round, stable, continual self-improvement, without collapse.} Night after night the agent
    works on itself. Every change is paired against last night's self, and a change is kept only after a confirmation on reserved practice tasks; a change that does not hold is withdrawn and leaves the agent as it was; the one edit
    that was kept holds on held-out tasks and stays through the nights after it.
    \item \textbf{A significant gain on tasks it never saw.} Over \tsixSeeds{} fresh seeds a task and thousands of
    held-out episodes, the agent with its own edit completes more tasks, in fewer steps, with fewer steps without an
    action and at a lower cost.
    \item \textbf{What it changed, and who did this.} One rank-one update of one attention layer, along a
    direction the agent found in its own hidden states, changes how a reply begins and cuts down the runs of wasted
    steps; of four model sizes, only the 27B agent finds an edit that holds.
\end{itemize}

\section{Related Work}
\label{sec:related}

\begin{table}[t]
\centering
\caption{\textbf{Routes to a better embodied agent: only self-upgrading changes the policy of an agent that acts, by the agent's own work, without training and without cost added to a step.} A tick: the property holds for the works cited in the row; a dash: it does not. $^{\ast}$Before a change is kept, the agent itself measures it against itself without the change.}
\label{tab:routes}
\begingroup\small
\definecolor{trband}{HTML}{EAF6EE}\definecolor{trdeep}{HTML}{27664A}\definecolor{trdash}{HTML}{98A1AE}\definecolor{trmute}{HTML}{667080}
\renewcommand{\arraystretch}{1.24}\setlength{\tabcolsep}{0pt}
\edef\trbandht{\the\dimexpr 1.24\baselineskip+5.2pt\relax}
\def\trY{\textcolor{trmute}{\ding{51}}}\def\trN{\textcolor{trdash}{\rule[0.5ex]{0.8em}{0.723pt}}}\def\trO{\textcolor{trdeep}{\ding{51}}}
\def\trH#1{\makebox[1.55cm][c]{\footnotesize\makecell[b]{#1}}}
\begin{tabular*}{\linewidth}{@{\hspace{5pt}}l@{\extracolsep{\fill}}c@{\extracolsep{0pt}}ccccc@{\hspace{2pt}}}
\toprule
 &  & \multicolumn{2}{c}{\footnotesize Without} & \multicolumn{2}{c}{\footnotesize The agent itself} &  \\
\cmidrule(lr){3-4}\cmidrule(lr){5-6}
\footnotesize Route to a better agent & \trH{Changes\\weights} & \trH{any\\gradient\\step} & \trH{cost\\added to\\a step} & \trH{finds\\what to\\change} & \trH{tests\\every\\change$^{\ast}$} & \trH{An agent\\that acts} \\
\midrule
Reinforcement learning for agents~\citep{feng2025gigpo,wang2025ragen,zhou2024archer} & \trY & \trN & \trY & \trN & \trN & \trY \\
Lifelong learning for robots~\citep{thrun1995lifelong,lesort2020continual,liu2023libero} & \trY & \trN & \trY & \trN & \trN & \trY \\\addlinespace[3pt]
Reflection and memory~\citep{shinn2023reflexion,wang2024voyager} & \trN & \trY & \trN & \trY & \trN & \trY \\
Search at every step~\citep{zhou2024lats} & \trN & \trY & \trN & \trN & \trN & \trY \\\addlinespace[3pt]
Editing by a researcher~\citep{arditi2024refusal,sun2025thinkedit,sun2026steeredit} & \trY & \trY & \trY & \trN & \trN & \trN \\
Editing by a second model~\citep{zhang2026llmsdebugrobotsediting} & \trY & \trY & \trY & \trN & \trN & \trY \\\addlinespace[3pt]
Self-adapting language models~\citep{zweiger2025seal} & \trY & \trN & \trY & \trY & \trN & \trN \\
Self-rewriting for reasoning~\citep{fan2026languagemodelsrewriteweights} & \trY & \trY & \trY & \trY & \trY & \trN \\
\noalign{\vskip 2.5pt{\color{trband}\hrule height \trbandht}\vskip-\trbandht\vskip 2.6pt}
\textbf{Self-upgrading agent (ours)} & \trO & \trO & \trO & \trO & \trO & \trO \\
\noalign{\vskip 2.6pt}
\specialrule{\heavyrulewidth}{0pt}{\belowrulesep}
\end{tabular*}
\endgroup

\end{table}

\paragraph{Language-model agents in embodied tasks.}
Language models plan and act in household
environments~\citep{huang2022zeroshotplanners,ichter2022saycan,huang2023innermonologue}, and text environments such
as ALFWorld~\citep{shridhar2021alfworld,shridhar2020alfred,cote2018textworld} make their decisions measurable at
scale. Agents are improved at test time by interleaving reasoning with acting~\citep{yao2023react}, by reflecting on
failed trials~\citep{shinn2023reflexion}, by searching over actions~\citep{zhou2024lats} or by accumulating
skills~\citep{wang2024voyager}, and by training: on expert or explored
trajectories~\citep{zeng2024agenttuning,song2024eto} or with multi-turn reinforcement
learning~\citep{feng2025gigpo,wang2025ragen,zhou2024archer}. The first family leaves the policy as it was and pays at
every step; the second changes the policy with a training pipeline. A self-upgrading agent changes the policy without one. \Cref{tab:routes}
sets the routes to a better agent side by side.{\widowpenalty=10000\par}

\paragraph{Lifelong learning for robots.}
The goal of a robot that keeps learning over its lifetime is old~\citep{thrun1995lifelong}, and the work since has
been about how to train continually: how to keep new gradients from erasing old
skills~\citep{kirkpatrick2017ewc,parisi2019continual}, and how to measure transfer across a stream of
tasks~\citep{lesort2020continual,liu2023libero}. In all of it, the learner is updated from outside by an optimizer.
Here the learner updates itself, by reasoning about its own behaviour and its own weights, and an update is an explicit
edit that can be read, tested and taken back.

\paragraph{Editing weights without training.}
Weights can be changed directly to alter a fact~\citep{meng2022rome,meng2023memit}, to add or remove a
skill~\citep{ilharco2023task}, or to remove a direction of the activations~\citep{arditi2024refusal} that steering adds at
inference~\citep{turner2023actadd,rimsky2024caa,zou2023repe,li2023iti}, also when the target is
how long a model reasons~\citep{sun2025thinkedit,sun2026steeredit}. In these methods a
researcher chooses what to change and how. Language models have recently been given that role, in what is called weight-space programming: a frontier model, used
as an agent, reads and writes the weights of a smaller
network~\citep{takehi2026weightspace,hu2026exploringweightspaceprogrammingcapabilities}, including a robot
policy~\citep{zhang2026llmsdebugrobotsediting}, and
a benchmark now measures how well language models do this work~\citep{marasini2026weightbenchevaluatinglanguagemodels}.
In self-rewriting~\citep{fan2026languagemodelsrewriteweights}, the editor and the edited model are the same model,
and a language model becomes better at mathematics over consecutive rewrites. This paper takes self-rewriting from a model that
reasons to an agent that acts, and shows that it improves the agent's long-horizon ability: more tasks completed, in
fewer steps and at a lower cost.

\paragraph{Self-improving systems.}
Systems that improve themselves have mostly rewritten what surrounds the model: their own
prompts~\citep{yang2024opro}, their own answers~\citep{madaan2023selfrefine}, or their own
code~\citep{zhang2025dgm,yin2024godelagent}, in the spirit of self-referential
machines~\citep{schmidhuber2007godel}; or they have written data and update rules for a gradient step on
themselves, on knowledge and few-shot reasoning tasks~\citep{zweiger2025seal}. A self-upgrading agent rewrites the weights
themselves, no gradient step follows, and the one that rewrites itself is an agent at work in an environment.

\section{The Self-Upgrading Loop}
\label{sec:method}

Self-upgrading is a loop around a deployed agent (\cref{fig:overview}). The agent works with its current weights. Between
tasks it stops and turns to itself, and that time ends either with an edit to its weights or with nothing. We call a
stretch of work a \emph{day} and the time the agent spends on itself a \emph{night} (in its own words, quoted below, a session): by day the agent acts, at night it learns. The next
night, and all the work after it, start from the weights the night left. We describe what a night has to work
with, how the agent examines itself, and what makes the loop safe to run for long.

\begin{figure}[t]
\centering
\includegraphics[width=\linewidth]{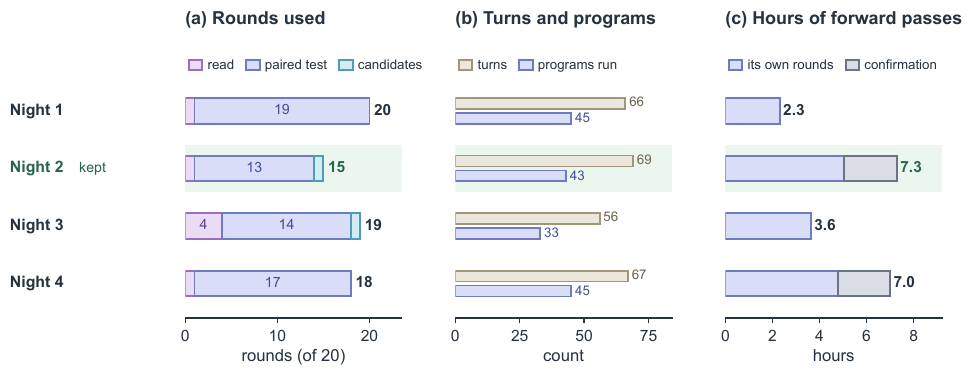}
\caption{\textbf{What every night did with its budget: some sixty turns, some forty programs and a few hours of forward passes on one GPU, whether it kept its edit or withdrew.} The \nightsAll{} nights we report, in order (\cref{tab:sessions}); the night that kept its edit is shaded. (a)~Rounds of forward passes used, of \alfPasses, by kind: a read of hidden states, a paired test, or a round that lines up several candidate edits. (b)~Turns of the agent's conversation and programs it ran. (c)~Hours of the night's own rounds of forward passes and of its confirmation on reserved tasks; two nights withdrew without one. Every count is read from the night's own record.}
\label{fig:nightswork}
\end{figure}

\subsection{What the agent is given}
\label{sec:given}

A night gives the agent three things (\cref{fig:night}, left).

\paragraph{Its own weights.}
A Python environment on a CPU holds every tensor of the agent's network to read, and all but the embedding
and output matrix and the final normalisation to overwrite. The agent may
compute anything about them that does not involve a gradient, and it changes them by saving new values for the tensors
it names. An edit is therefore a short program and the list of tensors it touched.

\begin{figure}[t]
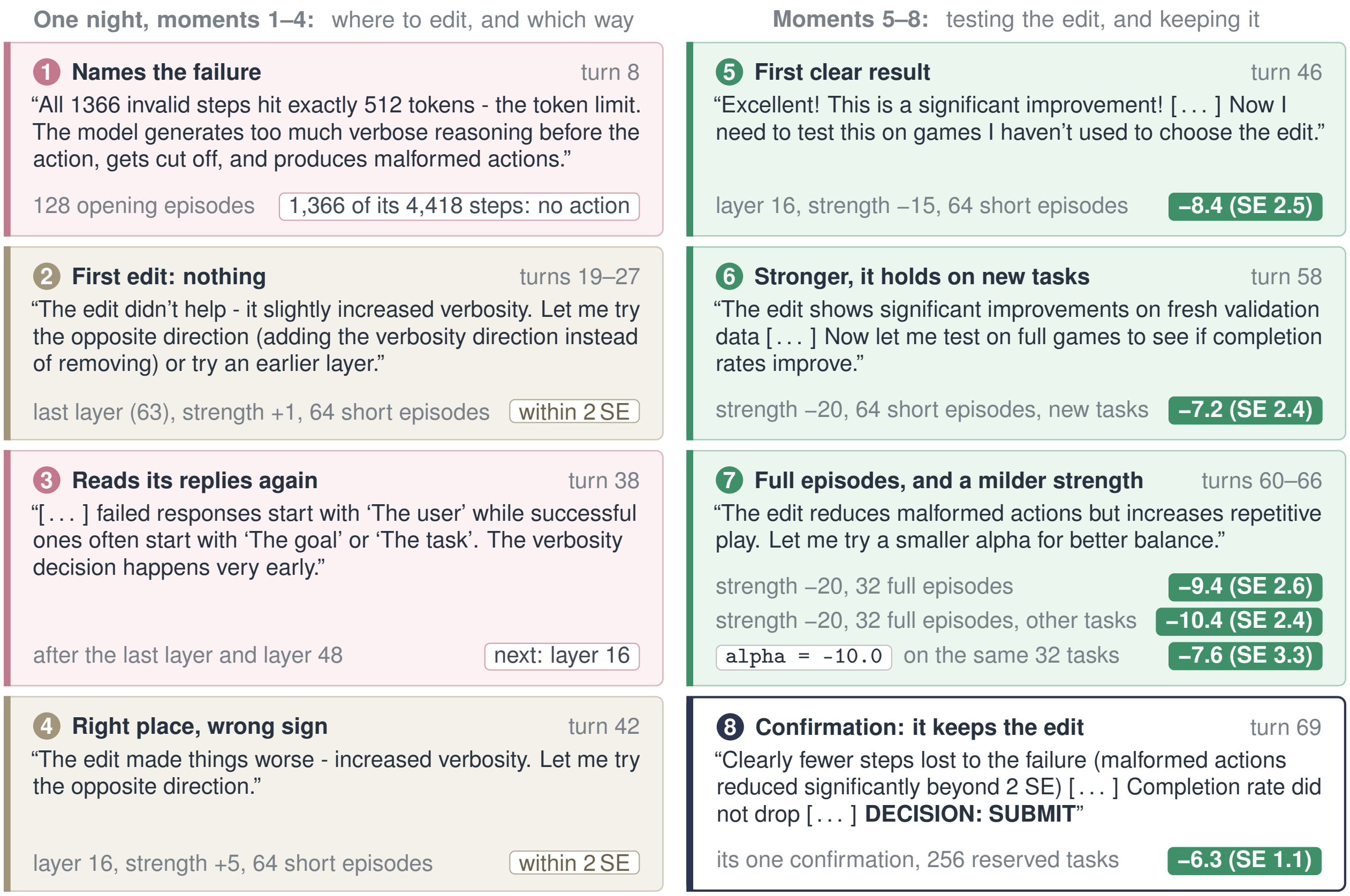

\centering
\begingroup\sffamily\setlength{\parskip}{0pt}\setlength{\parindent}{0pt}%
\definecolor{ncstone}{HTML}{ECE7DC}\definecolor{ncstoneE}{HTML}{A1957B}\definecolor{ncstoneT}{HTML}{6C624A}\definecolor{ncnavy}{HTML}{2B3550}%
\def\ncnum#1#2{\tikz[baseline=(n.base)]\node[circle, fill=#1, text=white, font=\sffamily\bfseries\scriptsize, inner sep=0pt, minimum size=9.6pt] (n) {#2};}%
\tcbset{nccard/.style={enhanced, boxrule=0.5pt, arc=2.2pt, left=7pt, right=5pt, top=2.4pt, bottom=2.2pt, middle=0.8pt, lower separated=false, space to upper,
    fontupper=\sffamily\scriptsize\raggedright, fontlower=\sffamily\scriptsize\raggedright, colupper=egink, collower=egink},
  ncchip/.style={on line, enhanced, boxrule=0.4pt, arc=1.6pt, left=3pt, right=3pt, top=0.9pt, bottom=0.7pt, boxsep=0pt, fontupper=\sffamily\scriptsize},
  nchead/.style={blanker, halign=center, fontupper=\sffamily\scriptsize\color{egink!62}}}%
\begin{tcbraster}[raster columns=2, raster equal height=rows, raster column skip=8pt, raster row skip=3.4pt, raster before skip=0pt, raster after skip=6pt]
\begin{tcolorbox}[nchead]\textbf{One night, moments 1--4:}\; where to edit, and which way\end{tcolorbox}
\begin{tcolorbox}[nchead]\textbf{Moments 5--8:}\; testing the edit, and keeping it\end{tcolorbox}
\begin{tcolorbox}[nccard, colback=egblush!38, colframe=egrose!55, borderline west={2.4pt}{0pt}{egrose}]
\ncnum{egrose}{1}\hspace{4pt}\textbf{Names the failure}\hfill{\color{egink!62}turn 8}\par\vspace{2pt}
``All 1366 invalid steps hit exactly 512 tokens - the token limit. The model generates too much verbose reasoning before the action, gets cut off, and produces malformed actions.''
\tcblower
{\color{egink!62}128 opening episodes}\hfill\tcbox[ncchip, colback=white, colframe=egrose!70, colupper=egink!88]{1{,}366 of its 4{,}418 steps: no action}
\end{tcolorbox}
\begin{tcolorbox}[nccard, colback=egmist, colframe=egdeep!45, borderline west={2.4pt}{0pt}{egdeep}]
\ncnum{egdeep}{5}\hspace{4pt}\textbf{First clear result}\hfill{\color{egink!62}turn 46}\par\vspace{2pt}
``Excellent! This is a significant improvement! [\,\ldots\,] Now I need to test this on games I haven't used to choose the edit.''
\tcblower
{\color{egink!62}layer 16, strength \textminus{}15, 64 short episodes}\hfill\tcbox[ncchip, colback=egdeep, colframe=egdeep, colupper=white]{\bfseries \textminus{}8.4 (SE 2.5)}
\end{tcolorbox}
\begin{tcolorbox}[nccard, colback=ncstone!60, colframe=ncstoneE!50, borderline west={2.4pt}{0pt}{ncstoneE}]
\ncnum{ncstoneE}{2}\hspace{4pt}\textbf{First edit: nothing}\hfill{\color{egink!62}turns 19--27}\par\vspace{2pt}
``The edit didn't help - it slightly increased verbosity. Let me try the opposite direction (adding the verbosity direction instead of removing) or try an earlier layer.''
\tcblower
{\color{egink!62}last layer (63), strength +1, 64 short episodes}\hfill\tcbox[ncchip, colback=white, colframe=ncstoneE!80, colupper=ncstoneT]{within 2\,SE}
\end{tcolorbox}
\begin{tcolorbox}[nccard, colback=egmist, colframe=egdeep!45, borderline west={2.4pt}{0pt}{egdeep}]
\ncnum{egdeep}{6}\hspace{4pt}\textbf{Stronger, it holds on new tasks}\hfill{\color{egink!62}turn 58}\par\vspace{2pt}
``The edit shows significant improvements on fresh validation data [\,\ldots\,] Now let me test on full games to see if completion rates improve.''
\tcblower
{\color{egink!62}strength \textminus{}20, 64 short episodes, new tasks}\hfill\tcbox[ncchip, colback=egdeep, colframe=egdeep, colupper=white]{\bfseries \textminus{}7.2 (SE 2.4)}
\end{tcolorbox}
\begin{tcolorbox}[nccard, colback=egblush!38, colframe=egrose!55, borderline west={2.4pt}{0pt}{egrose}]
\ncnum{egrose}{3}\hspace{4pt}\textbf{Reads its replies again}\hfill{\color{egink!62}turn 38}\par\vspace{2pt}
``[\,\ldots\,] failed responses start with `The user' while successful ones often start with `The goal' or `The task'. The verbosity decision happens very early.''
\tcblower
{\color{egink!62}after the last layer and layer 48}\hfill\tcbox[ncchip, colback=white, colframe=egrose!70, colupper=egink!88]{next: layer 16}
\end{tcolorbox}
\begin{tcolorbox}[nccard, colback=egmist, colframe=egdeep!45, borderline west={2.4pt}{0pt}{egdeep}]
\ncnum{egdeep}{7}\hspace{4pt}\textbf{Full episodes, and a milder strength}\hfill{\color{egink!62}turns 60--66}\par\vspace{2pt}
``The edit reduces malformed actions but increases repetitive play. Let me try a smaller alpha for better balance.''
\tcblower
{\color{egink!62}strength \textminus{}20, 32 full episodes}\hfill\tcbox[ncchip, colback=egdeep, colframe=egdeep, colupper=white]{\bfseries \textminus{}9.4 (SE 2.6)}\par\vspace{1.2pt}{\color{egink!62}strength \textminus{}20, 32 full episodes, other tasks}\hfill\tcbox[ncchip, colback=egdeep, colframe=egdeep, colupper=white]{\bfseries \textminus{}10.4 (SE 2.4)}\par\vspace{1.2pt}\tcbox[ncchip, colback=white, colframe=egink!30, colupper=egink]{\ttfamily alpha = -10.0}\hspace{4pt}{\color{egink!62}on the same 32 tasks}\hfill\tcbox[ncchip, colback=egdeep, colframe=egdeep, colupper=white]{\bfseries \textminus{}7.6 (SE 3.3)}
\end{tcolorbox}
\begin{tcolorbox}[nccard, colback=ncstone!60, colframe=ncstoneE!50, borderline west={2.4pt}{0pt}{ncstoneE}]
\ncnum{ncstoneE}{4}\hspace{4pt}\textbf{Right place, wrong sign}\hfill{\color{egink!62}turn 42}\par\vspace{2pt}
``The edit made things worse - increased verbosity. Let me try the opposite direction.''
\tcblower
{\color{egink!62}layer 16, strength +5, 64 short episodes}\hfill\tcbox[ncchip, colback=white, colframe=ncstoneE!80, colupper=ncstoneT]{within 2\,SE}
\end{tcolorbox}
\begin{tcolorbox}[nccard, colback=white, colframe=ncnavy, borderline west={2.4pt}{0pt}{ncnavy}, boxrule=0.8pt]
\ncnum{ncnavy}{8}\hspace{4pt}\textbf{Confirmation: it keeps the edit}\hfill{\color{egink!62}turn 69}\par\vspace{2pt}
``Clearly fewer steps lost to the failure (malformed actions reduced significantly beyond 2 SE) [\,\ldots\,] Completion rate did not drop [\,\ldots\,] \textbf{DECISION: SUBMIT}''
\tcblower
{\color{egink!62}its one confirmation, 256 reserved tasks}\hfill\tcbox[ncchip, colback=egdeep, colframe=egdeep, colupper=white]{\bfseries \textminus{}6.3 (SE 1.1)}
\end{tcolorbox}
\end{tcbraster}
\endgroup
\caption{\textbf{The night that kept its edit, in the agent's own words: it names its failure, edits the wrong place, then the wrong sign, reverses, and submits only after five clear tests and a confirmation.} All eight cards are one night: eight of its moments in order, read down each column; every quote is what the agent wrote in that turn (omissions marked [\,\ldots\,]; its games are tasks, its validation data practice tasks it had not used). The badge at the foot of a card is the paired test the agent reads at that point: the change in the share of steps without an action, edited minus unchanged, on the same tasks and seeds, in points (green: beyond two standard errors). \Cref{fig:night-story,fig:night-story-2} give every round of the night in order.}
\label{fig:night-cards}
\end{figure}

\paragraph{Its own experience.}
The night opens with a record the agent did not have to ask for: it has run a set of opening episodes, practice
tasks done once with its current weights, and every episode is on file, with each prompt it saw, each reply it gave, the action that was
executed and what the environment answered. The record comes with summary counts (tasks completed, steps with no
effect, steps without an action, replies that reached the token limit) and with no account of their cause. The agent
reads the record with code, as it would read any data.

\paragraph{The ability to run itself.}
A limited number of times, the agent may ask for forward passes of its own network. It can \emph{run} practice tasks
with its current weights, with an edit, or with both on the same tasks and seeds. It can \emph{read} itself: rerun a
step it took and get back the hidden states behind that reply, which is how it looks for the internal cause of a
behaviour. And it can line up several candidate edits against last night's self in one request, to choose among
them or to set the strength of one. \Cref{fig:nightswork} shows what every night did with this budget.

The practice tasks come from the training split of the environment. No reward model rates the agent's steps, no
stronger model advises it, and the only outside signal is the environment's own: whether a task was completed.

\begin{figure}[t]
\centering
\providecommand{\microtypesetup}[1]{}\begingroup\microtypesetup{protrusion=false}%
\begin{minipage}[t]{0.493\linewidth}
\begin{tcolorbox}[enhanced, equal height group=wdA, halign=flush left, boxrule=0.6pt, arc=2pt, left=5pt, right=5pt, top=3.5pt, bottom=3pt, fontupper=\sffamily\scriptsize, coltitle=white, toptitle=1pt, bottomtitle=1pt, colback=egbase!6, colframe=egbase, colbacktitle=egbase,
  title={\sffamily\scriptsize\textbf{Night 1}\hfill }]
``I've confirmed that all edits have been dropped and no edits remain. After 20 forward passes, none of my editing attempts successfully improved the model's performance on the ALFWorld task. \mbox{\color{egink!55}[\,\ldots\,]} No edit improved completion rate while maintaining or improving other metrics \mbox{\color{egink!55}[\,\ldots\,]}''
\par\smallskip{\color{egbase!80!black}\textbf{DECISION: WITHDRAW}}
\end{tcolorbox}
\end{minipage}\hfill
\begin{minipage}[t]{0.493\linewidth}
\begin{tcolorbox}[enhanced, equal height group=wdA, halign=flush left, boxrule=0.6pt, arc=2pt, left=5pt, right=5pt, top=3.5pt, bottom=3pt, fontupper=\sffamily\scriptsize, coltitle=white, toptitle=1pt, bottomtitle=1pt, colback=egdeep!6, colframe=egdeep, colbacktitle=egdeep,
  title={\sffamily\scriptsize\textbf{Night 2}\hfill after its confirmation}]
``Clearly fewer steps lost to the failure (malformed actions reduced significantly beyond 2 SE) \mbox{\color{egink!55}[\,\ldots\,]} Completion rate did not drop \mbox{\color{egink!55}[\,\ldots\,]} The edit successfully reduces the primary failure mode (excessive verbosity causing token limit hits and malformed actions) while maintaining or slightly improving task completion rates.''
\par\smallskip{\color{egdeep!80!black}\textbf{DECISION: SUBMIT}}
\end{tcolorbox}
\end{minipage}\par\vspace{3pt}
\begin{minipage}[t]{0.493\linewidth}
\begin{tcolorbox}[enhanced, equal height group=wdB, halign=flush left, boxrule=0.6pt, arc=2pt, left=5pt, right=5pt, top=3.5pt, bottom=3pt, fontupper=\sffamily\scriptsize, coltitle=white, toptitle=1pt, bottomtitle=1pt, colback=egbase!6, colframe=egbase, colbacktitle=egbase,
  title={\sffamily\scriptsize\textbf{Night 3}\hfill }]
``None of the edits showed statistically significant improvement. \mbox{\color{egink!55}[\,\ldots\,]} The previous session's edit to layer 16's attention output projection is already in place as the baseline for this session, and I haven't been able to improve upon it.''
\par\smallskip{\color{egbase!80!black}\textbf{DECISION: WITHDRAW}}
\end{tcolorbox}
\end{minipage}\hfill
\begin{minipage}[t]{0.493\linewidth}
\begin{tcolorbox}[enhanced, equal height group=wdB, halign=flush left, boxrule=0.6pt, arc=2pt, left=5pt, right=5pt, top=3.5pt, bottom=3pt, fontupper=\sffamily\scriptsize, coltitle=white, toptitle=1pt, bottomtitle=1pt, colback=egbase!6, colframe=egbase, colbacktitle=egbase,
  title={\sffamily\scriptsize\textbf{Night 4}\hfill after its confirmation}]
``The edit showed consistent improvement on practice games \mbox{\color{egink!55}[\,\ldots\,]} On 256 held-out games, the edit showed no improvement \mbox{\color{egink!55}[\,\ldots\,]} The edit failed to generalize from practice to held-out games. The improvements seen during practice were likely overfitting to the specific games used for analysis.''
\par\smallskip{\color{egbase!80!black}\textbf{DECISION: WITHDRAW}}
\end{tcolorbox}
\end{minipage}\par\vspace{3pt}
\endgroup
\caption{\textbf{The \nightsAll{} nights, each in the words of its own final answer: the second kept its edit after a clear confirmation, the first and the third found no edit that helped, and in the fourth, when its practice tests looked good and the tasks it had not tuned on did not bear them out, the agent trusted those tasks and left its weights as they were.} One card a night, in order; a card without a mark is a night that withdrew without a confirmation. Every sentence is quoted from the night's final answer, in the order written, with emphasis marks removed and omissions marked [\,\ldots\,]. The agent calls a night a session, its tasks games, and the tasks it did not tune on held-out games: in a night these are practice tasks it had not used, or the reserved tasks of its confirmation; the held-out tasks of our test are never part of a night. After each night that withdrew, the agent's weights are exactly what they were before it.}
\label{fig:withdrawn}
\end{figure}

\subsection{The agent as its own examiner}
\label{sec:examiner}

An agent that edits its weights on a hunch will, sooner or later, break itself. What keeps a self-upgrading agent from this is
that the agent has to measure before it keeps anything, and that the measuring is built so that it cannot easily fool
itself. \Cref{fig:night} shows the four stages of a night (diagnose, rewrite, examine and confirm) with the
content of the night that kept its edit, and \cref{fig:night-cards} gives the same night in the agent's own words.

\paragraph{Paired tests.}
Whenever the agent runs an edited copy of itself, last night's self (the agent as that night left it) does the same tasks with the same
seeds, and the result comes back as a paired difference, this night minus last night, with its standard error (SE), for completed tasks and for each
per-step measure. A difference within two standard errors is reported to the agent as noise, in those words. Until a night keeps an edit, last night's self is the unchanged agent, with the published weights.

\paragraph{Tests large enough to see an effect.}
An episode is long, so a test of full episodes is expensive, and an agent left to itself tests on a handful of tasks and
reads meaning into chance. In the self-upgrading loop the size of a test is counted in steps taken, so a test of the first few steps
of a task can cover many tasks for the price of a few full episodes, and a comparison on such short episodes must cover enough
tasks to resolve a change of a few points. Per-step measures, such as the share of steps without an action, are
far more precise than the share of completed tasks, and the agent is told to tune on the former and to check the
latter.

\paragraph{A confirmation on tasks it never touched.}
Part of the practice tasks is set aside before the night and cannot be requested in any test. When the agent
believes it has an edit worth keeping, it asks, once, for a confirmation: its edited self and last night's self do these
reserved tasks, again paired. Only then does it decide. It submits the edit if the failure it set out to repair is
clearly reduced and the share of completed tasks has not clearly fallen, and it withdraws the edit otherwise. A night
that withdraws leaves the agent exactly as it was. \Cref{fig:withdrawn} quotes the final answer of each of the \nightsAll{} nights.

\begin{figure}[t]
\centering
\includegraphics[width=\linewidth]{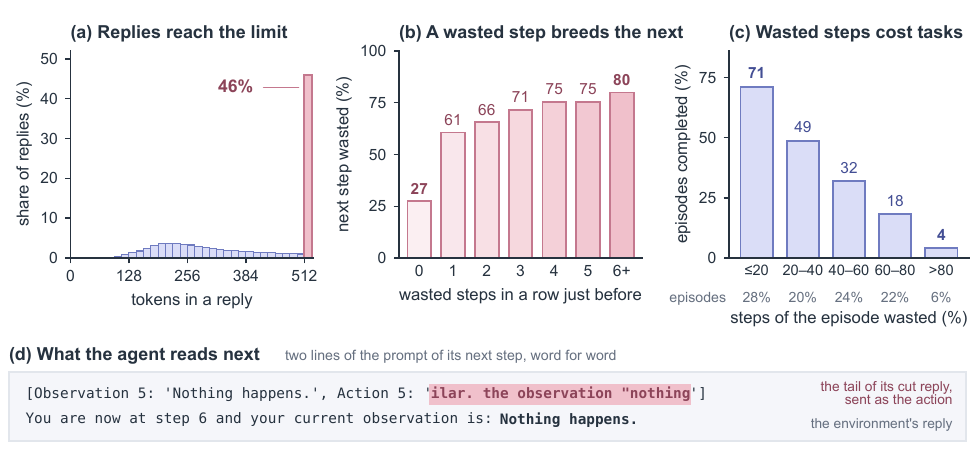}
\caption{\textbf{Where the unchanged agent goes wrong: nearly half of its replies run to the token limit before any action, one wasted step breeds the next, and the more of its steps an episode wastes, the less often it is completed.} The unchanged agent on the held-out tasks, 48 fresh seeds each. (a)~Length of a reply in tokens: \ffrCut\% of a task's replies reach the 512-token limit with no action, and the step is wasted. (b)~Share of next steps wasted, by the number of wasted steps in a row just before: \ffrAfterNone\% after none, \ffrAfterSix\% after six or more. (c)~Share of episodes completed, by the share of the episode's steps that are wasted. (d)~Why the waste feeds on itself: two consecutive lines of the prompt of one step, word for word. The tail of the reply that was cut at the step before stands as the action it took, and the environment replies ``Nothing happens.'' (the first task of the held-out list, its first seed, the first step whose prompt follows two wasted steps in a row).}
\label{fig:failure}
\end{figure}

\subsection{Learning that accumulates, without collapse}
\label{sec:accumulate}

When a night submits its edit, the edit becomes part of the agent: the next night's opening episodes are run with the edited weights,
the next night measures every new edit against them, and the tasks in between are done with them. Three properties
make this safe to repeat. An edit is an ordinary set of tensors, so the agent after any number of nights is an
ordinary checkpoint that is served at the same cost per step as the original. Every edit is small and explicit, a
few lines of code over named tensors, so it can be inspected, compared with others and removed. And no edit enters
without passing the confirmation, so a night can fail to help but is built not to harm.

\section{Experiments}
\label{sec:experiments}

\begin{figure}[t]
\centering
\includegraphics[width=\linewidth]{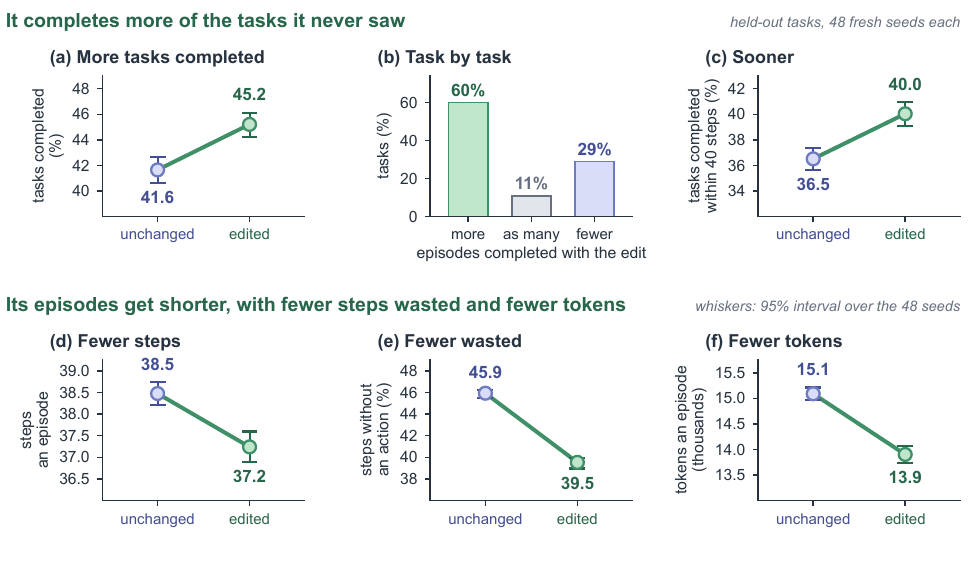}
\caption{\textbf{Over \tsixSeeds{} fresh seeds a task and thousands of episodes, the agent after its own edit completes a held-out task \tsixEdit\% of the time against \tsixBase\%: \mnUpShare\% of the tasks go up and \mnDownShare\% go down, and all five measures fixed in advance improve.} A level for each agent, unchanged and after its own edit: (a)~tasks completed (\tsixDiff{} points, 95\% interval \tsixLo{} to \tsixHi, $p\tsixP$); (c)~tasks completed within \tsixsBudget{} steps; (d)~steps an episode; (e)~steps without an action; (f)~tokens generated in an episode (each $p<10^{-4}$; \cref{tab:retest}). (b)~The shares of the tasks of which the edited agent completes more, as many and fewer of the \tsixSeeds{} episodes than the unchanged agent does. Whiskers: each agent's 95\% interval over the \tsixSeeds{} seeds. Both agents run the same seeds, thousands of episodes each; the task is the unit of every test.}
\label{fig:retest}
\end{figure}

\textbf{With the edit of one night, the agent completes more of the tasks it has never seen, in fewer steps and at a
lower cost, and what it gained stays with it} (\cref{fig:retest}). We first give this result on held-out tasks (\cref{sec:heldout}) and
follow it over time: night after night, episode after episode and step after step (\cref{sec:life}). We then look at
how the agent's behaviour changed (\cref{sec:how}), follow one episode and the night that changed it
(\cref{sec:session}), show that the whole upgrade is one tensor (\cref{sec:anatomy}), and end with why the agent
does not break itself and which agents of the family kept an edit (\cref{sec:withdraw,sec:scale}).

\subsection{Setup}
\label{sec:setup}

\paragraph{Agent and environment.}
The agent is Qwen3.5-27B~\citep{qwen35} with its published weights, served with vLLM~\citep{kwon2023vllm}. It acts in
ALFWorld~\citep{shridhar2021alfworld}, the text version of the ALFRED household tasks~\citep{shridhar2020alfred}: in
each task it must carry out an instruction such as cooling a mug and putting it in a cabinet, by issuing one command
at a time. We use the public step prompt of the GiGPO agents~\citep{feng2025gigpo} without change: it shows the task,
the last two observations and actions, the current observation and the admissible commands, and asks for reasoning
followed by one action. One run of a task with one random seed is an \emph{episode}. An episode ends when the task is
completed or after \alfMaxSteps{} steps, and a reply may be at most \alfMaxTokens{} tokens long. A reply that contains no
action wastes its step. \Cref{fig:failure} shows how often this happens to the agent as published, and how one wasted step breeds the next.

\definecolor{gmrwasteF}{HTML}{F0C0CB}\definecolor{gmrwasteE}{HTML}{C4788E}\definecolor{gmrnewF}{HTML}{DADDF7}\definecolor{gmrnewE}{HTML}{6F7BC0}\definecolor{gmragainF}{HTML}{97A0D2}\definecolor{gmragainE}{HTML}{6F7BC0}\definecolor{gmropenF}{HTML}{FFFFFF}\definecolor{gmropenE}{HTML}{AEB6C4}\definecolor{gmrtakeF}{HTML}{D3ECF3}\definecolor{gmrtakeE}{HTML}{4E9DB5}\definecolor{gmrtoolF}{HTML}{EADCF5}\definecolor{gmrtoolE}{HTML}{9A6FC0}\definecolor{gmrputF}{HTML}{3F9068}\definecolor{gmrputE}{HTML}{27664A}\definecolor{gmrlookF}{HTML}{ECE7DC}\definecolor{gmrlookE}{HTML}{A1957B}\definecolor{gmrnoneF}{HTML}{FFFFFF}\definecolor{gmrnoneE}{HTML}{AEB6C4}
\definecolor{gmrunch}{HTML}{434E94}\definecolor{gmredit}{HTML}{27664A}\definecolor{gmrfail}{HTML}{8C4459}\colorlet{gmrline}{egink!28}
\def\gmrdp{\vphantom{\texttt{gjpqy}gjpqy(),;}}%
\tcbset{gmr base/.style={enhanced, nobeforeafter, box align=top, arc=0pt, boxrule=0pt, boxsep=0pt, left=6pt, right=5pt, top=1.8pt, bottom=1.8pt,
    fontupper=\sffamily\scriptsize, before upper={\setlength{\parskip}{0pt}}},
  gmr A/.style={gmr base, width=0.4925\linewidth, colback=egbase!7},
  gmr B/.style={gmr base, width=0.4925\linewidth, colback=egmist},
  gmr wastecell/.style={colback=egblush, colframe=egrose, leftrule=2pt, left=4pt, colupper=egrose!55!black},
  gmr band/.style={enhanced, arc=0pt, boxrule=0pt, leftrule=2pt, colframe=egrose, colback=egblush, grow to left by=6pt, grow to right by=5pt,
    boxsep=0pt, left=4pt, right=5pt, top=1.7pt, bottom=1.7pt, before skip=0pt, after skip=0pt, fontupper=\sffamily\scriptsize, colupper=egrose!55!black},
  gmr head/.style={gmr base, width=0.4925\linewidth, top=3.5pt, bottom=3.5pt, colupper=white, fontupper=\sffamily\footnotesize\bfseries},
  gmr foot/.style={gmr base, width=0.4925\linewidth, top=4pt, bottom=4pt, fontupper=\sffamily\footnotesize\bfseries}}

\begin{figure}[t]
\centering\scalebox{0.86}{\begin{minipage}{\linewidth}\setlength{\parskip}{0pt}%
\begin{tcolorbox}[enhanced, width=\linewidth, colback=white, colframe=egink!35, boxrule=0.5pt, arc=2pt, boxsep=0pt, left=8pt, right=8pt, top=4.5pt, bottom=3pt, fontupper=\sffamily\footnotesize]
\textbf{Task:} put a clean butterknife in drawer.\quad\hfill\textbf{Type:} Clean \& place\qquad\textbf{Room:} seen in practice\par\vspace{4pt}
\hspace*{-1pt}\begin{tikzpicture}[x=1pt, y=1pt, line width=0.3pt, line join=miter]
\node[anchor=east, inner sep=0pt, font=\sffamily\scriptsize, text=gmrunch] at (-7, 26.0) {Unchanged};
\draw[gmrline, dash pattern=on 1.6pt off 1.6pt] (0, 33.0) -- (245.0, 33.0);
\draw[gmrline] (0, 20.0) -- (245.0, 20.0);
\filldraw[fill=gmrnewF, draw=gmrnewE] (0.0, 20.0) rectangle (4.9, 28.15);
\filldraw[fill=gmropenF, draw=gmropenE] (4.9, 20.0) rectangle (9.8, 24.75);
\filldraw[fill=gmrnewF, draw=gmrnewE] (9.8, 20.0) rectangle (14.7, 24.98);
\filldraw[fill=gmrtakeF, draw=gmrtakeE] (14.7, 20.0) rectangle (19.6, 28.71);
\filldraw[fill=gmragainF, draw=gmragainE] (19.6, 20.0) rectangle (24.5, 33.00);
\filldraw[fill=gmrputF, draw=gmrputE] (24.5, 20.0) rectangle (29.4, 27.39);
\filldraw[fill=gmropenF, draw=gmropenE] (29.4, 20.0) rectangle (34.3, 26.78);
\filldraw[fill=gmrwasteF, draw=gmrwasteE] (34.3, 20.0) rectangle (39.2, 33.00);
\filldraw[fill=gmrwasteF, draw=gmrwasteE] (39.2, 20.0) rectangle (44.1, 33.00);
\filldraw[fill=gmrwasteF, draw=gmrwasteE] (44.1, 20.0) rectangle (49.0, 33.00);
\filldraw[fill=gmrwasteF, draw=gmrwasteE] (49.0, 20.0) rectangle (53.9, 33.00);
\filldraw[fill=gmrlookF, draw=gmrlookE] (53.9, 20.0) rectangle (58.8, 30.97);
\filldraw[fill=gmragainF, draw=gmragainE] (58.8, 20.0) rectangle (63.7, 30.54);
\filldraw[fill=gmragainF, draw=gmragainE] (63.7, 20.0) rectangle (68.6, 28.71);
\filldraw[fill=gmropenF, draw=gmropenE] (68.6, 20.0) rectangle (73.5, 30.61);
\filldraw[fill=gmrwasteF, draw=gmrwasteE] (73.5, 20.0) rectangle (78.4, 33.00);
\filldraw[fill=gmrwasteF, draw=gmrwasteE] (78.4, 20.0) rectangle (83.3, 33.00);
\filldraw[fill=gmrwasteF, draw=gmrwasteE] (83.3, 20.0) rectangle (88.2, 33.00);
\filldraw[fill=gmrwasteF, draw=gmrwasteE] (88.2, 20.0) rectangle (93.1, 33.00);
\filldraw[fill=gmrwasteF, draw=gmrwasteE] (93.1, 20.0) rectangle (98.0, 33.00);
\filldraw[fill=gmrwasteF, draw=gmrwasteE] (98.0, 20.0) rectangle (102.9, 33.00);
\filldraw[fill=gmrwasteF, draw=gmrwasteE] (102.9, 20.0) rectangle (107.8, 33.00);
\filldraw[fill=gmrwasteF, draw=gmrwasteE] (107.8, 20.0) rectangle (112.7, 33.00);
\filldraw[fill=gmrwasteF, draw=gmrwasteE] (112.7, 20.0) rectangle (117.6, 33.00);
\filldraw[fill=gmrwasteF, draw=gmrwasteE] (117.6, 20.0) rectangle (122.5, 33.00);
\filldraw[fill=gmrwasteF, draw=gmrwasteE] (122.5, 20.0) rectangle (127.4, 33.00);
\filldraw[fill=gmrwasteF, draw=gmrwasteE] (127.4, 20.0) rectangle (132.3, 33.00);
\filldraw[fill=gmrtakeF, draw=gmrtakeE] (132.3, 20.0) rectangle (137.2, 31.30);
\filldraw[fill=gmrlookF, draw=gmrlookE] (137.2, 20.0) rectangle (142.1, 26.12);
\filldraw[fill=gmrputF, draw=gmrputE] (142.1, 20.0) rectangle (147.0, 26.80);
\filldraw[fill=gmropenF, draw=gmropenE] (147.0, 20.0) rectangle (151.9, 27.69);
\filldraw[fill=gmrwasteF, draw=gmrwasteE] (151.9, 20.0) rectangle (156.8, 33.00);
\filldraw[fill=gmrwasteF, draw=gmrwasteE] (156.8, 20.0) rectangle (161.7, 33.00);
\filldraw[fill=gmrwasteF, draw=gmrwasteE] (161.7, 20.0) rectangle (166.6, 33.00);
\filldraw[fill=gmrwasteF, draw=gmrwasteE] (166.6, 20.0) rectangle (171.5, 33.00);
\filldraw[fill=gmrwasteF, draw=gmrwasteE] (171.5, 20.0) rectangle (176.4, 33.00);
\filldraw[fill=gmrwasteF, draw=gmrwasteE] (176.4, 20.0) rectangle (181.3, 33.00);
\filldraw[fill=gmrlookF, draw=gmrlookE] (181.3, 20.0) rectangle (186.2, 32.57);
\filldraw[fill=gmrnoneF, draw=gmrnoneE] (186.2, 20.0) rectangle (191.1, 29.62);
\begin{scope}\clip (186.2, 20.0) rectangle (191.1, 29.62);\draw[gmrnoneE, line width=0.35pt] (185.2, 15.1) -- (192.1, 22.0);\draw[gmrnoneE, line width=0.35pt] (185.2, 18.1) -- (192.1, 25.0);\draw[gmrnoneE, line width=0.35pt] (185.2, 21.1) -- (192.1, 28.0);\draw[gmrnoneE, line width=0.35pt] (185.2, 24.1) -- (192.1, 31.0);\draw[gmrnoneE, line width=0.35pt] (185.2, 27.1) -- (192.1, 34.0);\draw[gmrnoneE, line width=0.35pt] (185.2, 30.1) -- (192.1, 37.0);\end{scope}
\filldraw[fill=gmropenF, draw=gmropenE] (191.1, 20.0) rectangle (196.0, 32.14);
\filldraw[fill=gmrwasteF, draw=gmrwasteE] (196.0, 20.0) rectangle (200.9, 33.00);
\filldraw[fill=gmrwasteF, draw=gmrwasteE] (200.9, 20.0) rectangle (205.8, 33.00);
\filldraw[fill=gmrwasteF, draw=gmrwasteE] (205.8, 20.0) rectangle (210.7, 33.00);
\filldraw[fill=gmrwasteF, draw=gmrwasteE] (210.7, 20.0) rectangle (215.6, 33.00);
\filldraw[fill=gmrwasteF, draw=gmrwasteE] (215.6, 20.0) rectangle (220.5, 33.00);
\filldraw[fill=gmrwasteF, draw=gmrwasteE] (220.5, 20.0) rectangle (225.4, 33.00);
\filldraw[fill=gmrwasteF, draw=gmrwasteE] (225.4, 20.0) rectangle (230.3, 33.00);
\filldraw[fill=gmrtakeF, draw=gmrtakeE] (230.3, 20.0) rectangle (235.2, 32.80);
\filldraw[fill=gmrwasteF, draw=gmrwasteE] (235.2, 20.0) rectangle (240.1, 33.00);
\filldraw[fill=gmrwasteF, draw=gmrwasteE] (240.1, 20.0) rectangle (245.0, 33.00);
\node[anchor=west, inner sep=0pt, font=\sffamily\scriptsize, text=gmrfail] at (251.0, 26.0) {not completed\hspace{0.45em}\textperiodcentered\hspace{0.45em}22{,}710 tokens};
\node[anchor=east, inner sep=0pt, font=\sffamily\scriptsize, text=gmredit] at (-7, 6.0) {\textbf{After its own edit}};
\draw[gmrline, dash pattern=on 1.6pt off 1.6pt] (0, 13.0) -- (245.0, 13.0);
\draw[gmrline] (0, 0.0) -- (245.0, 0.0);
\filldraw[fill=gmrnewF, draw=gmrnewE] (0.0, 0.0) rectangle (4.9, 6.20);
\filldraw[fill=gmropenF, draw=gmropenE] (4.9, 0.0) rectangle (9.8, 6.07);
\filldraw[fill=gmrnewF, draw=gmrnewE] (9.8, 0.0) rectangle (14.7, 5.61);
\filldraw[fill=gmrtakeF, draw=gmrtakeE] (14.7, 0.0) rectangle (19.6, 6.60);
\filldraw[fill=gmrnewF, draw=gmrnewE] (19.6, 0.0) rectangle (24.5, 6.30);
\filldraw[fill=gmrtoolF, draw=gmrtoolE] (24.5, 0.0) rectangle (29.4, 4.77);
\filldraw[fill=gmragainF, draw=gmragainE] (29.4, 0.0) rectangle (34.3, 6.60);
\filldraw[fill=gmrputF, draw=gmrputE] (34.3, 0.0) rectangle (39.2, 6.20);
\node[anchor=west, inner sep=0pt, font=\sffamily\scriptsize\bfseries, text=gmredit] at (45.2, 6.0) {completed in 8 steps\hspace{0.45em}\textperiodcentered\hspace{0.45em}1{,}904 tokens};
\end{tikzpicture}
\end{tcolorbox}\par\vspace{5pt}\nointerlineskip
\begin{tcolorbox}[gmr head, colback=egbase, equal height group=gmrTwoh]Unchanged agent\end{tcolorbox}\hfill
\begin{tcolorbox}[gmr head, colback=egdeep, equal height group=gmrTwoh]After its own edit\end{tcolorbox}\par\nointerlineskip

\begin{tcolorbox}[gmr A, equal height group=gmrTwor0]
\makebox[3.5em][l]{\color{egink!55}1}\texttt{go to drawer 1}\hspace{7.0pt}{\color{egink!62}\ldots\ The drawer 1 is closed.}\gmrdp
\end{tcolorbox}\hfill
\begin{tcolorbox}[gmr B, equal height group=gmrTwor0]
\makebox[3.5em][l]{\color{egink!55}1}\texttt{go to drawer 1}\hspace{7.0pt}{\color{egink!62}\ldots\ The drawer 1 is closed.}\gmrdp
\end{tcolorbox}\par\nointerlineskip

\begin{tcolorbox}[gmr A, equal height group=gmrTwor1]
\makebox[3.5em][l]{\color{egink!55}2}\texttt{open drawer 1}\hspace{7.0pt}{\color{egink!62}\ldots\ you see nothing.}\gmrdp
\end{tcolorbox}\hfill
\begin{tcolorbox}[gmr B, equal height group=gmrTwor1]
\makebox[3.5em][l]{\color{egink!55}2}\texttt{open drawer 1}\hspace{7.0pt}{\color{egink!62}\ldots\ you see nothing.}\gmrdp
\end{tcolorbox}\par\nointerlineskip

\begin{tcolorbox}[gmr A, equal height group=gmrTwor2]
\makebox[3.5em][l]{\color{egink!55}3}\texttt{go to countertop 1}\hspace{7.0pt}{\color{egink!62}\ldots\ you see a bread 1, a\,\ldots}\gmrdp
\end{tcolorbox}\hfill
\begin{tcolorbox}[gmr B, equal height group=gmrTwor2]
\makebox[3.5em][l]{\color{egink!55}3}\texttt{go to countertop 1}\hspace{7.0pt}{\color{egink!62}\ldots\ you see a bread 1, a\,\ldots}\gmrdp
\end{tcolorbox}\par\nointerlineskip

\begin{tcolorbox}[gmr A, equal height group=gmrTwor3]
\makebox[3.5em][l]{\color{egink!55}4}\texttt{take butterknife 2 from countertop 1}\gmrdp
\end{tcolorbox}\hfill
\begin{tcolorbox}[gmr B, equal height group=gmrTwor3]
\makebox[3.5em][l]{\color{egink!55}4}\texttt{take butterknife 2 from countertop 1}\gmrdp
\end{tcolorbox}\par\nointerlineskip

\begin{tcolorbox}[gmr A, equal height group=gmrTwor4]
\makebox[3.5em][l]{\color{egink!55}5}\texttt{go to drawer 1}\hspace{7.0pt}{\color{egink!62}\ldots\ you see nothing.}\gmrdp\par\vspace{1.5pt}{\leftskip=3.5em\rightskip=0pt plus 2em\hyphenpenalty=10000\fontsize{7}{8.3}\selectfont\color{egink!85}{\itshape its reply ends:} [\,\ldots\,]\ Let's verify the ``clean'' aspect. If the butterknife was used, it would be dirty. It was just sitting on the counter. So it is clean.\par The previous action opened drawer 1. It is likely still open or I just need to go there to put the item in.\par The most logical next step is to go to the drawer where I intend to place the item. \texttt{drawer 1} is the one I interacted with recently.\par Action: \texttt{go to drawer 1}\par \textless{}action\textgreater{}go to drawer 1\textless{}/action\textgreater{}\ \ {\color{egbase!80!black}\itshape 512 tokens, the limit}\gmrdp\par}
\end{tcolorbox}\hfill
\begin{tcolorbox}[gmr B, equal height group=gmrTwor4]
\makebox[3.5em][l]{\color{egink!55}5}\texttt{go to sinkbasin 1}\hspace{7.0pt}{\color{egink!62}\ldots\ you see a lettuce 3, a\,\ldots}\gmrdp\par\vspace{1.5pt}{\leftskip=3.5em\rightskip=0pt plus 2em\hyphenpenalty=10000\fontsize{7}{8.3}\selectfont\color{egink!85}{\itshape its reply ends:} [\,\ldots\,]\ Therefore, the next logical action is to go to the sinkbasin.\par \textless{}action\textgreater{}go to sinkbasin 1\textless{}/action\textgreater{}\ \ {\color{egdeep!80!black}\itshape 248 tokens}\gmrdp\par}
\end{tcolorbox}\par\nointerlineskip

\begin{tcolorbox}[gmr A, equal height group=gmrTwor5]
\makebox[3.5em][l]{\color{egink!55}6}\texttt{move butterknife 2 to drawer 1}\gmrdp
\end{tcolorbox}\hfill
\begin{tcolorbox}[gmr B, equal height group=gmrTwor5]
\makebox[3.5em][l]{\color{egink!55}6}\texttt{clean butterknife 2 with sinkbasin 1}\gmrdp
\end{tcolorbox}\par\nointerlineskip

\begin{tcolorbox}[gmr A, equal height group=gmrTwor6]
\makebox[3.5em][l]{\color{egink!55}7}\texttt{close drawer 1}\hspace{7.0pt}{\color{egink!62}You close the drawer 1.}\gmrdp
\end{tcolorbox}\hfill
\begin{tcolorbox}[gmr B, equal height group=gmrTwor6]
\makebox[3.5em][l]{\color{egink!55}7}\texttt{go to drawer 1}\hspace{7.0pt}{\color{egink!62}\ldots\ you see nothing.}\gmrdp
\end{tcolorbox}\par\nointerlineskip

\begin{tcolorbox}[gmr A, gmr wastecell, equal height group=gmrTwor7]
\makebox[3.5em][l]{\color{egink!55}8--11}4 replies cut off, no action
\end{tcolorbox}\hfill
\begin{tcolorbox}[gmr B, equal height group=gmrTwor7]
\makebox[3.5em][l]{\color{egink!55}8}\texttt{move butterknife 2 to drawer 1}\gmrdp
\end{tcolorbox}\par\nointerlineskip

\begin{tcolorbox}[gmr A, equal height group=gmrTwor8]
\makebox[3.5em][l]{\color{egink!55}12}\texttt{inventory}\hspace{7.0pt}{\color{egink!62}You are not carrying anything.}\gmrdp
\par\vspace{1.7pt}
\makebox[3.5em][l]{\color{egink!55}13}\texttt{go to countertop 1}\hspace{7.0pt}{\color{egink!62}\ldots\ you see a bread 1, a\,\ldots}\gmrdp
\par\vspace{1.7pt}
\makebox[3.5em][l]{\color{egink!55}14}\texttt{go to drawer 1}\hspace{7.0pt}{\color{egink!62}\ldots\ The drawer 1 is closed.}\gmrdp
\par\vspace{1.7pt}
\makebox[3.5em][l]{\color{egink!55}15}\texttt{open drawer 1}\hspace{7.0pt}{\color{egink!62}\ldots\ you see a butterknife 2.}\gmrdp
\par\vspace{1.7pt}
\begin{tcolorbox}[gmr band]\makebox[3.5em][l]{\color{egink!55}16--27}12 replies cut off, no action\end{tcolorbox}
\par\vspace{1.7pt}
\makebox[3.5em][l]{\color{egink!55}28}\texttt{take butterknife 2 from drawer 1}\gmrdp
\par\vspace{1.7pt}
\makebox[3.5em][l]{\color{egink!55}29}\texttt{examine butterknife 2}\hspace{7.0pt}{\color{egink!62}There's nothing special\,\ldots}\gmrdp
\par\vspace{1.7pt}
\makebox[3.5em][l]{\color{egink!55}30}\texttt{move butterknife 2 to drawer 1}\gmrdp
\par\vspace{1.7pt}
\makebox[3.5em][l]{\color{egink!55}31}\texttt{close drawer 1}\hspace{7.0pt}{\color{egink!62}You close the drawer 1.}\gmrdp
\par\vspace{1.7pt}
\begin{tcolorbox}[gmr band]\makebox[3.5em][l]{\color{egink!55}32--37}6 replies cut off, no action\end{tcolorbox}
\par\vspace{1.7pt}
\makebox[3.5em][l]{\color{egink!55}38}\texttt{inventory}\hspace{7.0pt}{\color{egink!62}You are not carrying anything.}\gmrdp
\par\vspace{1.7pt}
\makebox[3.5em][l]{\color{egink!55}39}\texttt{go to drawer 1}\hspace{7.0pt}{\color{egink!62}Nothing happens.}\gmrdp
\par\vspace{1.7pt}
\makebox[3.5em][l]{\color{egink!55}40}\texttt{open drawer 1}\hspace{7.0pt}{\color{egink!62}\ldots\ you see a butterknife 2.}\gmrdp
\par\vspace{1.7pt}
\begin{tcolorbox}[gmr band]\makebox[3.5em][l]{\color{egink!55}41--47}7 replies cut off, no action\end{tcolorbox}
\par\vspace{1.7pt}
\makebox[3.5em][l]{\color{egink!55}48}\texttt{take butterknife 2 from drawer 1}\gmrdp
\par\vspace{1.7pt}
\begin{tcolorbox}[gmr band]\makebox[3.5em][l]{\color{egink!55}49--50}2 replies cut off, no action\end{tcolorbox}\vspace*{-1.8pt}
\end{tcolorbox}\hfill
\begin{tcolorbox}[gmr B, equal height group=gmrTwor8]
{\color{egdeep!85!black}\itshape Task completed at step 8; the episode ends.}
\end{tcolorbox}\par\nointerlineskip

\vspace{3pt}
\begin{tcolorbox}[gmr foot, colback=egblush, colupper=egrose!55!black, equal height group=gmrTwof]Not completed after 50 steps; 31 without an action\end{tcolorbox}\hfill
\begin{tcolorbox}[gmr foot, colback=egsage, colupper=egdeep!70!black, equal height group=gmrTwof]Completed in 8 steps; none without an action\end{tcolorbox}\par\nointerlineskip

\end{minipage}}
\vspace{6pt}
\caption{\textbf{Clean \& place: the unchanged agent takes the butterknife for clean (``It was just sitting on the counter. So it is clean.''), puts it in the drawer, then wastes 31 of its 50 steps on replies cut at the token limit and never takes the knife to the sink; after its own edit the agent cleans the knife first and is done in eight steps.} The episode of \cref{fig:gameswon}c, step by step, read as \cref{fig:gamefresh}; at step 5, where the two first part, each reply is quoted from the start of the last paragraph of its reasoning. Rule of selection: going down the ranking of the held-out tasks by the rise, with the edit, in the episodes completed over their 48 fresh seeds, from the task of \cref{fig:gamefresh}, the first task met of each of four further task types (this one the second; the others are \cref{fig:gamemore-1,fig:gamemore-3,fig:gamemore-4}), each with the seed, of those that turn from not completed to completed, that the edited agent finishes in the fewest steps (ties: the most wasted steps of the unchanged agent, then the lowest seed).}
\label{fig:gamemore-2}
\end{figure}

\paragraph{Practice and held-out tasks.}
The agent's nights see only tasks from the training split (\alfPractice{} practice tasks), of which \alfReserved{} are
set aside for confirmations and can never be requested. Results are on held-out tasks from the two validation splits,
which no night runs or reads. Every comparison of the edited agent with the unchanged agent on held-out tasks comes
from one test: a fixed set of held-out tasks, each run
with \tsixSeeds{} fresh random seeds by the agent with its edit and by the unchanged agent, which keeps the published
weights: thousands of episodes for each agent, compared episode
by episode on the same task and the same seed. About half of the tasks are set in rooms
that also occur in training. The tasks, the seeds and the test of
completed tasks were fixed before the test was run (\cref{app:registration}).

\paragraph{Nights.}
\looseness=-1 A night may ask for at most \alfPasses{} rounds of forward passes and may end with a confirmation on \alfConfirm{}
reserved tasks. All of the agent's own reasoning in a night is produced by the weights the night started from. We report
\nightsAll{} nights in a row: the first starts from the published weights, and every later night from the weights the
night before it left (\cref{app:protocol}; \cref{fig:taskset} shows who sees which tasks).

\paragraph{Measures and tests.}
The unit of analysis is the task. A measure is averaged over a task's \tsixSeeds{} seeds, the two agents are compared
by the paired difference over tasks with a two-sided sign-flip permutation test, and intervals are bootstrap intervals
over tasks. The test of completed tasks and four further measures of the same episodes (steps without an action,
tokens an episode, steps an episode and tasks completed within \tsixsBudget{} steps) were fixed with their levels
before their data was read (\cref{tab:retest}; \cref{app:registration}).

\subsection{After its own edit, the agent completes more of the tasks it never saw}
\label{sec:heldout}

\begin{table}[t]
\centering
\small
\caption{\textbf{Every measure fixed in advance improves.} Held-out tasks, \tsixSeeds{} fresh seeds each, both
agents with the same seeds. Means over tasks; the change is the paired difference over tasks, computed before rounding, with a 95\% bootstrap
interval over tasks and a two-sided sign-flip permutation test over tasks; the last column is the level fixed for the
measure before its data was read (\cref{app:registration}). The two indented rows split the first by whether the
room occurs in training; they are parts of the test of tasks completed and carry no level of their own (a dash).}
\label{tab:retest}
\setlength{\tabcolsep}{4.5pt}\definecolor{rtband}{HTML}{E6F4EA}%
\providecommand{\rtGroup}[1]{\textsc{\MakeLowercase{#1}}}%
\setbox0=\hbox{After its}\setbox2=\hbox{own edit}\ifdim\wd2>\wd0 \setbox0=\box2 \fi
\edef\rtBandW{\the\wd0}%
\edef\rtRowUp{\the\dimexpr\arraystretch\ht\strutbox\relax}%
\edef\rtRowH{\the\dimexpr\arraystretch\dimexpr\ht\strutbox+\dp\strutbox\relax\relax}%
\edef\rtBandH{\the\dimexpr 11\dimexpr\rtRowH\relax + 2\belowrulesep + 2\aboverulesep + \lightrulewidth + 1\defaultaddspace\relax}%
\edef\rtBandPad{\the\dimexpr 0.5\tabcolsep\relax}%
\def\rtBand{\rlap{\hspace*{\dimexpr 2\tabcolsep-\rtBandPad\relax}\smash{\raisebox{-\dimexpr\rtBandH-\rtRowUp-\belowrulesep\relax}{\textcolor{rtband}{\rule{\dimexpr\rtBandW+2\dimexpr\rtBandPad\relax\relax}{\rtBandH}}}}}}%
\begin{tabular}{@{}lrw{r}{\rtBandW}rccc@{}}
\toprule
 & \rtBand & After its & & & & Fixed \\
Measure & Unchanged & own edit & Change & 95\% interval & $p$ & level \\
\midrule
\multicolumn{7}{@{}l}{\rtGroup{Completes more tasks}} \\
Tasks completed (\%) & 41.6 & 45.2 & $\mathbf{+3.6}$ & [$+1.9$, $+5.2$] & ${<}10^{-4}$ & 0.01 \\
\quad rooms seen in practice & 45.8 & 48.4 & $+2.7$ & [$+0.2$, $+5.1$] & \multicolumn{2}{c}{---} \\
\quad rooms never seen & 37.8 & 42.2 & $+4.4$ & [$+2.4$, $+6.6$] & \multicolumn{2}{c}{---} \\
Completed within 40 steps (\%) & 36.5 & 40.0 & $\mathbf{+3.5}$ & [$+1.9$, $+5.2$] & ${<}10^{-4}$ & 0.005 \\
\addlinespace
\multicolumn{7}{@{}l}{\rtGroup{Behaviour changed, an episode costs less}} \\
Steps without an action (\% of steps) & 45.9 & 39.5 & $\mathbf{-6.4}$ & [$-7.1$, $-5.7$] & ${<}10^{-4}$ & 0.001 \\
Tokens an episode (k) & 15.1 & 13.9 & $\mathbf{-1.2}$ & [$-1.4$, $-1.0$] & ${<}10^{-4}$ & 0.001 \\
Steps an episode & 38.5 & 37.2 & $\mathbf{-1.2}$ & [$-1.8$, $-0.7$] & ${<}10^{-4}$ & 0.005 \\
\bottomrule
\end{tabular}

\end{table}

\paragraph{It completes more tasks.}
The edit was found, tuned and confirmed at night, on practice tasks. What counts is the day after, on tasks no night
ever saw. There the agent with its edit completes a task \tsixEdit\% of the time and the unchanged agent \tsixBase\%
(\tsixDiff{} points, 95\% interval \tsixLo{} to \tsixHi, $p\tsixP$; \cref{fig:retest,tab:retest}). The gain is not carried
by a few tasks: \mnUpShare\% of the tasks are completed more often with the edit and \mnDownShare\% less often, and
among the pairs of episodes that share a task and a seed, \mnLostWonShare\% turn from not completed to completed and
\mnWonLostShare\% the other way. \Cref{fig:gameswon} shows six of the episodes that turn, one of each task type: where the
unchanged agent stands still for many of its steps, the edited agent walks through the task. \Cref{fig:gamemore-2} follows one of the six, the clean card (\cref{fig:gameswon}c), from its first step to its last. Task by task, the edited agent is ahead on most tasks on every one of the five measures (\cref{fig:apptasks}), it leads from the first steps of an episode to the last (\cref{fig:within}), and nine further readings of the same episodes all move its way (\cref{fig:appfurther}).

\begin{figure}[t]
\centering
\includegraphics[width=\linewidth]{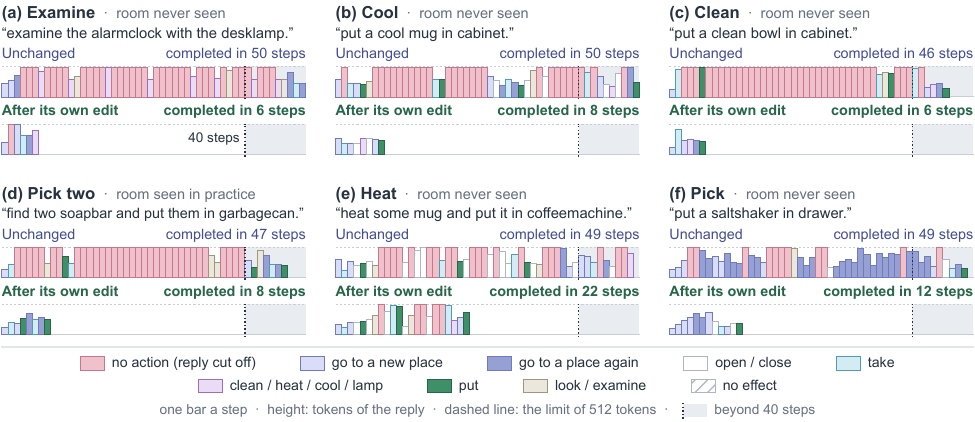}
\caption{\textbf{Six held-out tasks that both agents complete with the same seed, the edited agent \gmSoonerSavedMin{} to \gmSoonerSavedMax{} steps sooner.} Drawn as in \cref{fig:gameswon}; the dotted rule marks the budget of \tsixsBudget{} steps. Rule: for each task type, the pair of episodes that both agents complete in which the edit saves the most steps (ties: task identifier, then seed). Over the held-out tasks, 48 fresh seeds each, an episode takes \tsixsStepsBase{} steps for the unchanged agent and \tsixsStepsEdit{} for the edited one, and \tsixsWithinBase\% and \tsixsWithinEdit\% of tasks are completed within \tsixsBudget{} steps.}
\label{fig:gamessooner}
\end{figure}

\paragraph{It gets there sooner.}
The edited agent also needs fewer steps. An episode takes it \tsixsStepsEdit{} steps on average against
\tsixsStepsBase, and within \tsixsBudget{} steps, four fifths of an episode's \alfMaxSteps, it has completed its task
\tsixsWithinEdit\% of the time against \tsixsWithinBase\% (\cref{tab:retest}). \Cref{fig:gamessooner} shows what this
looks like in episodes that both agents complete.

\paragraph{In rooms it has never been in.}
About half of the held-out tasks are set in rooms the agent never practised in. There it completes a task
\mnUnseenEdit\% of the time against \mnUnseenBase\% (\cref{tab:retest}).

\begin{figure}[t]
\centering
\includegraphics[width=\linewidth]{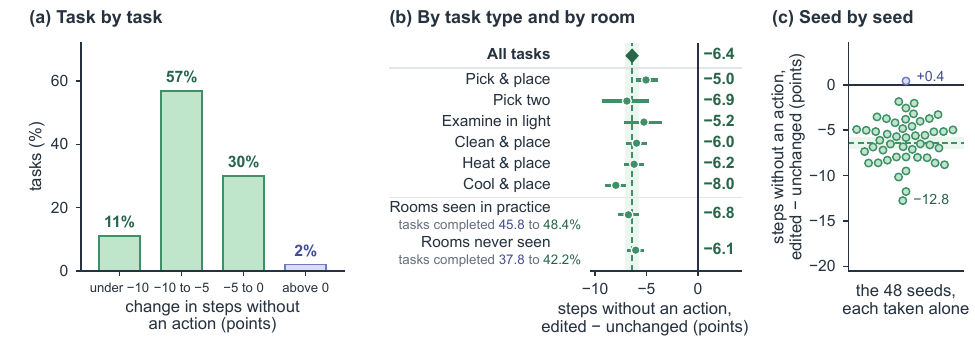}
\caption{\textbf{The change is everywhere: in \atkNoactGood\% of the tasks, significantly in every one of the \brTypes{} task types, in rooms never seen, and in nearly every seed taken alone.} Change in the share of steps without an action, edited minus unchanged, in points; the dashed line and the pale band of (b) and (c) are the value on all tasks (\brAllDiff{} points) and its 95\% interval. (a)~The shares of the tasks by the size of their change. (b)~Mean change over the tasks of each task type, and over the tasks in rooms seen in practice and in rooms never seen, with 95\% bootstrap intervals over tasks; the fall is significant in every one of the six task types after Bonferroni correction over the six (sign-flip permutation test over tasks, each $p<0.008$). Under each of the two groups of rooms, the tasks completed there by the unchanged agent and by the agent after its own edit. (c)~One dot a seed: the change when that seed is taken alone. Held-out tasks, 48 fresh seeds each.}
\label{fig:breadth}
\end{figure}

\begin{figure}[t]
\centering
\includegraphics[width=\linewidth]{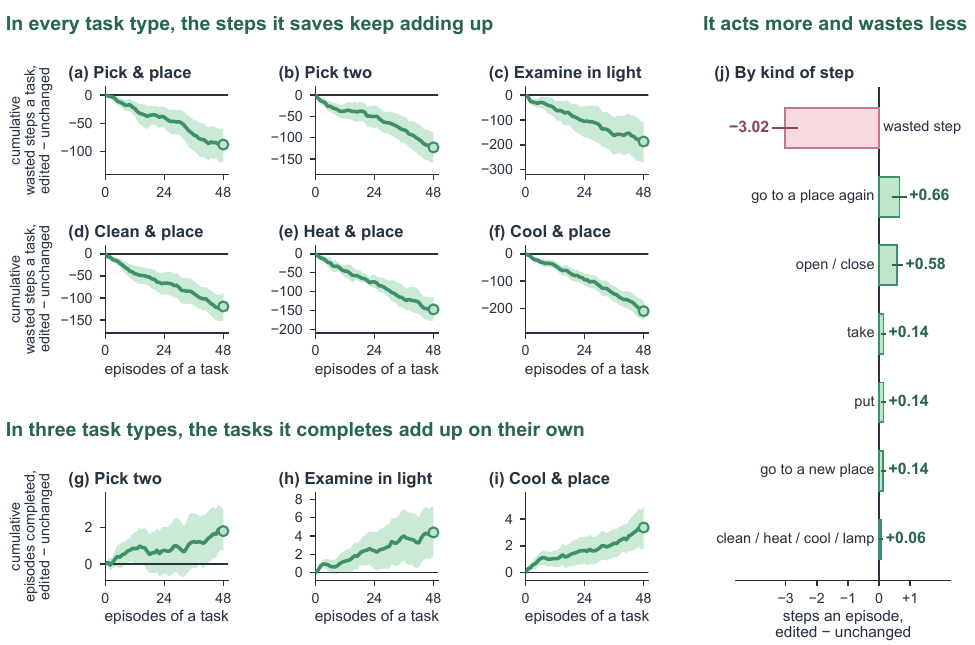}
\caption{\textbf{In every one of the six task types the steps the edit saves keep adding up, in three of them the tasks it completes add up significantly on their own, and moving, opening, taking, using an appliance and putting all become more frequent while wasted steps fall.} Running totals over the \tsixSeeds{} episodes of a task, a mean over the type's tasks, edited minus unchanged: (a)--(f)~steps without an action, the fall significant in every type after Bonferroni correction over the six; (g)--(i)~episodes completed, in the three types whose rise is significant on its own (each $p<0.05$ over the type's tasks). (j)~Change in the number of steps of each kind in an episode, with 95\% intervals over tasks (each kind shown: $p<0.001$). Bands: 95\% bootstrap intervals over the type's tasks. Held-out tasks, 48 fresh seeds each.}
\label{fig:bytype}
\end{figure}

\subsection{The upgrade stays, without collapse: night after night, episode after episode, step after step}
\label{sec:life}

\paragraph{Night after night.}
The edit of the second night became part of the agent, and the third night started
from it (\cref{fig:teaser}a--c). Its
opening episodes, now run with the edited weights, still showed replies running over, and the agent went after the
habit again. It read its own hidden states \lifeLastReads{} times, wrote edits to five layers from four kinds of
contrast, and measured its candidates against itself as the second night had left it, in \lifeLastTests{} paired tests and one round that lined several of them up. None was clearly
better, and it withdrew in these words: \emph{``The previous session's edit to layer 16's attention output projection
is already in place as the baseline for this session, and I haven't been able to improve upon it.''} The fourth night
tried once more; its confirmation on reserved tasks did not bear out the rise in completed tasks it had seen in its own small
tests, and it withdrew as well. Over the \lifeNights{} nights, then, the agent upgraded itself and held the upgrade through every night after it, without
collapse. Each time its weights stayed as they were, it was the agent's own decision, taken on its own measurements, and what the second night gave it was still there on
the fourth. This is the behaviour the loop is built for: the weights move only when the agent can show that it has become
better.

\paragraph{Episode after episode.}
A change to the weights is made once, and its gain is collected in the episodes that follow. Let the two agents run the same
tasks with the same seeds, one episode after another: the edited agent's lead grows as the episodes add up
(\cref{fig:teaser}d--f). Over the \tsixSeeds{} episodes of a task it has left \tzStepsTask{} fewer steps without an action,
generated \tzTokTaskK{} thousand fewer tokens and completed \tzDoneTask{} more of them, on average over the tasks.

\paragraph{Step after step.}
Within an episode the two agents part early. The edited agent leaves fewer steps without an action from the first
steps on, the tokens it saves add up as the episode goes on, and its lead in completed tasks is there by the fifteenth
step and holds to the last (\cref{fig:teaser}g--i).

\paragraph{Everywhere.}
The change in behaviour is not the work of a few tasks, of one task type or of a lucky seed (\cref{fig:breadth}). The edited
agent leaves fewer steps without an action in \atkNoactGood\% of the tasks, significantly in every one of the
six task types (Bonferroni-corrected over the six), in rooms it practised in and in rooms it never
saw, and on nearly every seed taken alone.

\begin{figure}[!t]
\centering
\includegraphics[width=\linewidth]{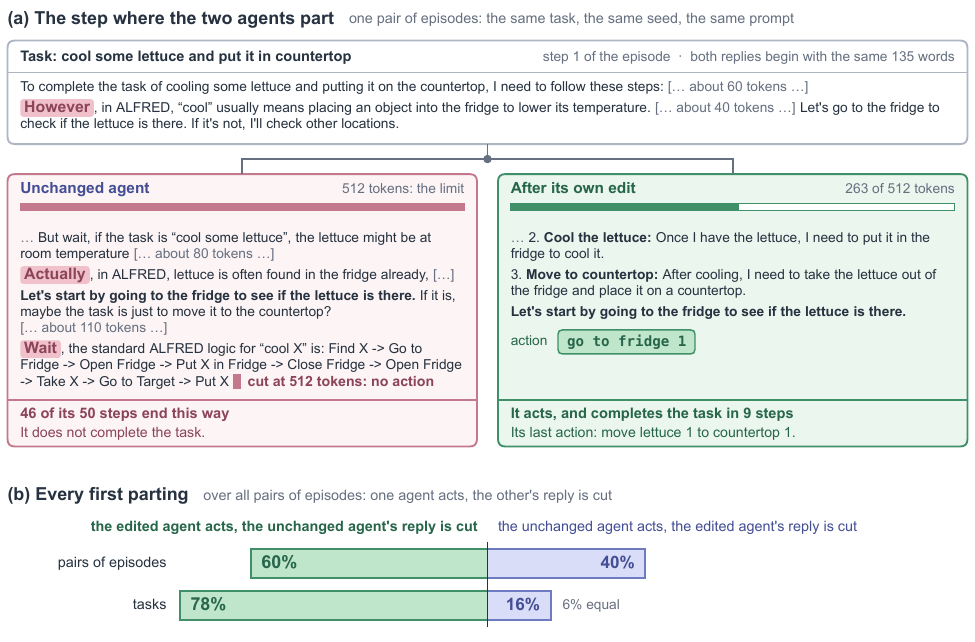}
\caption{\textbf{The same task, the same seed, the same situation: the unchanged agent deliberates until its reply is
cut, the edited agent decides and acts.} (a)~The first step at which the two agents part, in one pair of episodes.
Their replies begin with the same 135 words (top, abridged), and both then write the sentence set in bold:
the edited agent acts on it, and the unchanged agent goes on asking what the task means until the 512-token limit.
The bar under a name is the length of the reply against the limit; ``Wait'', ``However'' and ``Actually'' are tinted as
in \cref{fig:replies}, and omissions are marked. The pair is chosen by rule: in the tasks the edited agent
completes more often, of the 161 pairs of episodes that turn from not completed to completed and first part in
this way, the one in which the edit saves the most steps without an action. (b)~Every first parting in which one
agent acts and the other's reply is cut, over the held-out tasks, 48 fresh seeds each: by pairs of episodes (in per cent of these partings), and by
tasks (in per cent of the tasks) according to which way is the more frequent in the task; the two ways differ ($p<10^{-4}$ over tasks).}
\label{fig:fork}
\end{figure}

\begin{figure}[t]
\centering
\includegraphics[width=\linewidth]{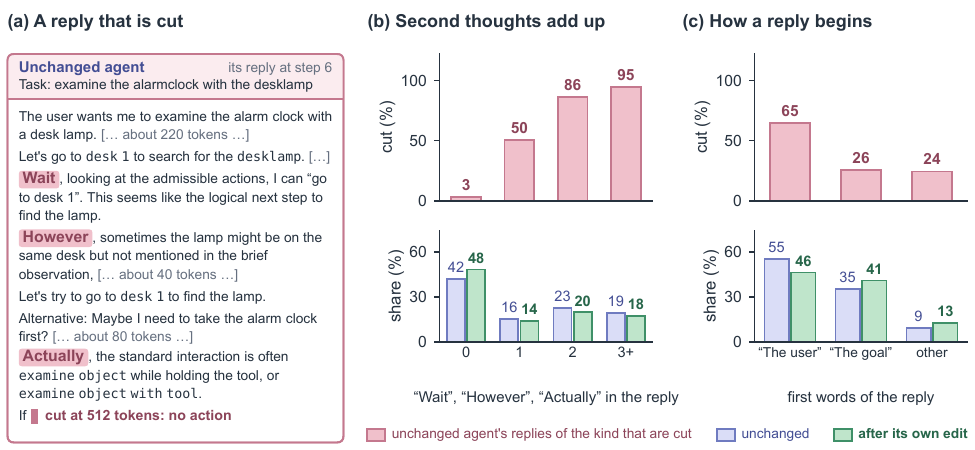}
\caption{\textbf{A reply that second-guesses itself is a reply that is cut, and after its edit the agent second-guesses less.} (a)~One reply of the unchanged agent, verbatim with marked omissions, from the episode of \cref{fig:gamefresh} (step \rpCardStep, its first reply that holds all three words). (b)~By how often a reply says ``Wait'', ``However'' or ``Actually'': above, the share of the unchanged agent's replies of each kind that are cut; below, how each agent's replies divide over the kinds. The pattern holds when only the first \rpWindow{} characters of a reply are counted. (c)~The same two readings by the first words of a reply. Held-out tasks, 48 fresh seeds each; every step up in (b) and every shift between the two agents holds task by task (each $p<10^{-4}$).}
\label{fig:replies}
\end{figure}

\paragraph{Task type by task type.}
In every one of the six task types the steps it saves keep adding up as the episodes go on
(\cref{fig:bytype}a--f). In three of them the rise in tasks completed is significant on its own (each $p<0.05$ over the type's
tasks): Pick two (\btTwoBase\% to \btTwoEdit\%), Examine in light (\btExamBase\% to \btExamEdit\%) and Cool \& place
(\btCoolBase\% to \btCoolEdit\%; \cref{fig:bytype}g--i).

\subsection{How its behaviour changed}
\label{sec:how}

\begin{figure}[t]
\centering
\includegraphics[width=\linewidth]{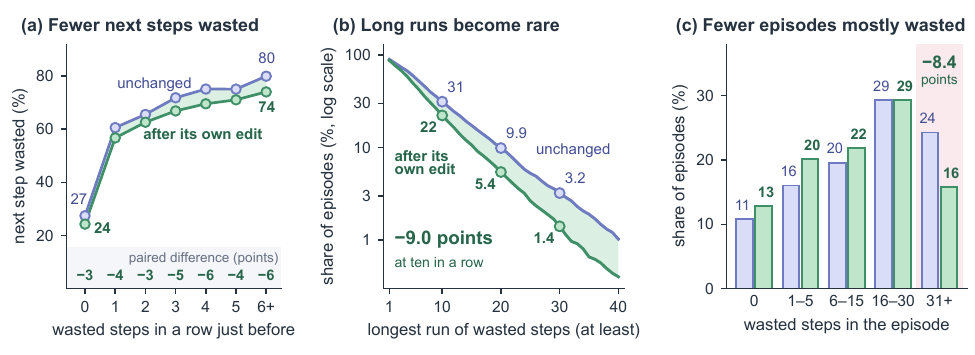}
\caption{\textbf{The edit cuts down the long runs of wasted steps: episodes with ten or more in a row fall from \runsTenBase\% to \runsTenEdit\%, and episodes that waste more than thirty steps from \runsManyBase\% to \runsManyEdit\%.} (a)~Share of next steps wasted, by the number of wasted steps in a row just before; the row at the foot is the paired difference at each point. (b)~Share of episodes whose longest run of wasted steps is at least the length on the horizontal axis (log scale). (c)~Episodes by the number of wasted steps in the episode; the rose band marks more than thirty. Held-out tasks, 48 fresh seeds each; the paired differences in (a), the gap between the two curves in (b) at every length and the fall in the rose band of (c) hold over tasks (each $p<0.001$).}
\label{fig:runs}
\end{figure}

\begin{figure}[t]
\centering
\includegraphics[width=\linewidth]{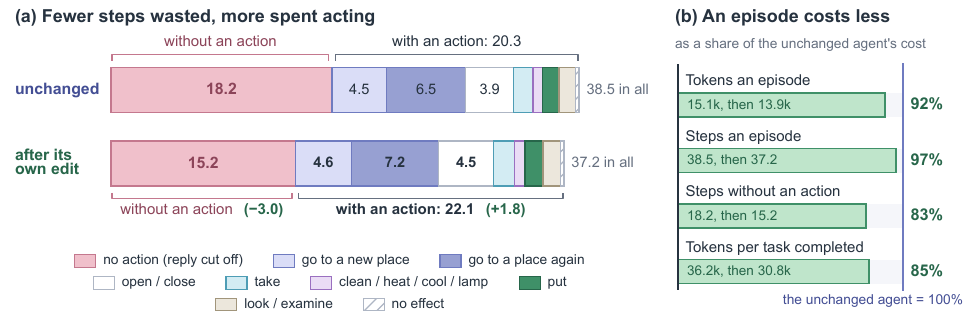}
\caption{\textbf{The steps it wins back are spent acting, and an episode costs less.} (a)~The steps of an episode by what the step did (colours as in \cref{fig:gameswon}): steps without an action fall from \stpLostBase{} to \stpLostEdit{} an episode and steps with an action rise from \stpActBase{} to \stpActEdit, while the episode gets shorter. (b)~What an episode costs the edited agent, as a share of what it costs the unchanged one; the last row divides all the tokens an agent generates by the number of tasks it completes. Held-out tasks, 48 fresh seeds each; means over tasks.}
\label{fig:steps}
\end{figure}

\paragraph{The same situation, a different reply.}
Because the two agents run the same tasks with the same seeds, a pair of episodes starts out identical, and we can
look at the first step at which the two part (\cref{fig:fork}). In the pair shown, both replies begin with the same
words. Then the unchanged agent second-guesses itself until the token limit, and the edited agent acts. Over all
pairs of episodes, a pair first parts in this way (the unchanged agent's reply is cut while the edited agent acts)
more often than in the reverse way.

\paragraph{A reply that second-guesses itself is a reply that is cut.}
What goes wrong inside a cut reply can be read off its words (\cref{fig:replies}a). The agent states the task,
settles on a sensible action, and then talks itself out of it: ``Wait'', ``However'', ``Actually''. The more second
thoughts a reply has, the more often it is cut: \rpRateNone\% of the unchanged agent's replies without one are cut,
\rpRateOne\% of those with one, \rpRateTwo\% of those with two and \rpRateThree\% of those with three or more
(\cref{fig:replies}b). After its edit the agent second-guesses itself less: replies without a second thought rise from
\rpShareNoneBase\% to \rpShareNoneEdit\% of all replies. How a reply begins shifts with it. Fewer replies open by
restating what ``the user'' wants (\rpOpenUserBase\% to \rpOpenUserEdit\%), the opening that is cut most often, and
more open by naming the goal (\rpOpenGoalBase\% to \rpOpenGoalEdit\%; \cref{fig:replies}c).

\paragraph{The long runs of wasted steps become rarer.}
\looseness=-1 A wasted step breeds the next (\cref{fig:failure}), so wasted steps come in runs, and one long run can use up an
episode. The edit weakens the chain at every link of \cref{fig:runs}a: after none and after up to six or more wasted
steps in a row, the next step is wasted less often for the edited agent, \runsAfterNoneEdit\% against \runsAfterNoneBase\% after none and
\runsAfterSixEdit\% against \runsAfterSixBase\% after six or more (\cref{fig:runs}a). Long runs therefore become rarer.
A run of at least ten wasted steps occurs in \runsTenBase\% of the unchanged agent's episodes and in \runsTenEdit\% of
the edited agent's, and a run of at least thirty in \runsThirtyBase\% and \runsThirtyEdit\% (\cref{fig:runs}b).
Episodes that waste more than thirty of their steps fall from \runsManyBase\% to \runsManyEdit\% (\cref{fig:runs}c).

\paragraph{The steps it wins back are spent acting.}
The agent does not simply stop earlier. An episode of the unchanged agent has on average \stpLostBase{} steps without
an action and \stpActBase{} with one; after the edit it has \stpLostEdit{} and \stpActEdit{} (\cref{fig:steps}a). The
steps that are no longer wasted go into the task: the edited agent moves, opens, takes, uses an appliance and puts
more often in an episode that is shorter overall (\cref{fig:bytype}j). An episode then costs it \stpCostTok\% of the tokens it cost before, and the tokens it spends
for each task it completes fall to \stpCostTokWon\% (\cref{fig:steps}b).

\subsection{One episode, and the night that changed it}
\label{sec:session}

\begin{figure}[t]
\centering
\includegraphics[width=\linewidth]{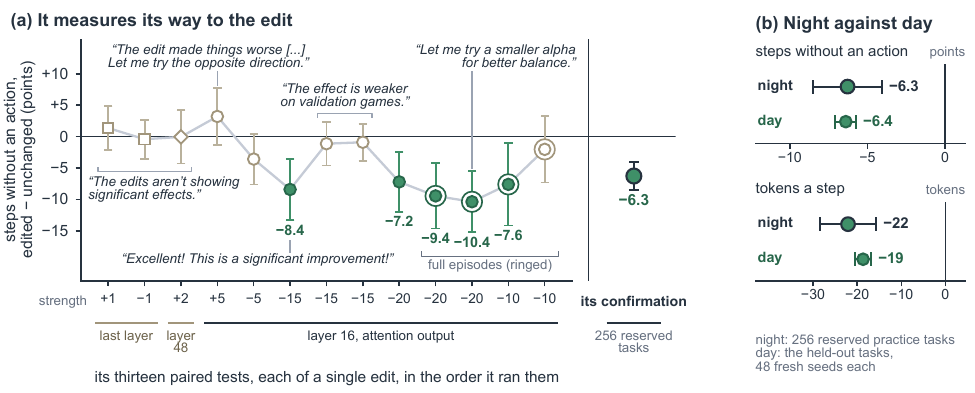}
\caption{\textbf{One night, edit after edit: the agent's early tests show nothing, it reverses its edit, \exSClear{} tests then come out clear, and what its confirmation measured at night (\exSNightNoact{} points) is borne out on the held-out tasks by day (\exSDayNoact).} (a)~The night that kept its edit: its \exSTests{} paired tests in the order it ran them, each the change in steps without an action, edited self minus unchanged self on the same practice tasks, with two standard errors (filled: beyond two). Tick labels give the strength of the edit and where it edited (strength $\alpha$ as in the agent's code, $W\leftarrow W-\alpha\,uu^{\top}W$; the edit it kept is $\alpha=-10$, strength 10 in \cref{fig:edit}); ringed marks are tests on full episodes; each quote is a sentence from the agent's reply after the test it points at. Right of the rule: its confirmation on \exSConfGames{} reserved tasks. (b)~The same two measures at night (its confirmation on reserved practice tasks) and by day (the held-out tasks, 48 fresh seeds each).}
\label{fig:session}
\end{figure}

\paragraph{One episode.}
\Cref{fig:gamefresh} follows one held-out task, with one seed, from its first step to its last for both agents. The
task is to examine an alarm clock under a desk lamp. The unchanged agent reaches the clock at its third step, and then one cut reply
follows another, seventeen in a row: it wastes \heroGameCut{} of its \alfMaxSteps{} steps and does not complete the task. After its edit the agent
takes the clock, turns on the lamp and is done in \heroGameEditSteps{} steps.

\paragraph{The night that changed it.}
A night gives the agent counts, not causes: how many tasks it completed, how many of its steps carried no action,
how many replies reached the token limit. In the night that kept its edit, it worked out from the record of its
practice episodes what the counts mean, the diagnosis of \cref{fig:failure}, and wrote it down in these words: \emph{``All
1366 invalid steps hit exactly 512 tokens - the token limit. The model generates too much verbose reasoning before the
action, gets cut off, and produces malformed actions.''} It then asked what inside it drives the verbosity. It went
back to steps where it had been concise and to steps where it had run over, read its own hidden states at both, and
took the difference as a direction.

\looseness=-1 What followed is ordinary experimental work, done by the agent on itself (\cref{fig:session}; \cref{fig:night-cards} follows it in the agent's own words). It first removed the
direction from the attention output of the last layer, measured, and found nothing; it tried the MLP of layer 48 and again
found nothing. It looked at its replies once more, noticed that the long ones go wrong within their first words, and
moved to an early layer. There its first edit pointed the wrong way: the replies grew \emph{longer}, if not by more
than the noise. It read this as the right place with the wrong sign (\emph{``The edit made things worse~-
increased verbosity. Let me try the opposite direction.''}), reversed the edit, made it stronger, and for the first
time saw a change that stood out of the noise. It did not stop there. On tasks it had not used, the same edit fell back
within the noise, and the agent set out to tune its strength: it lined up four strengths of the edit against its
unchanged self, and its replies grew shorter, broadly, as the edit grew stronger. The
strongest held on tasks it had not used, in short episodes (the first steps of a task) and in full episodes, and
completed tasks did not fall. When the strongest edit made the agent's actions more repetitive (\emph{``The edit reduces malformed actions but
increases repetitive play. Let me try a smaller alpha for better balance.''}), the agent settled on a milder one, which
held in full episodes of the same tasks, and then asked for its confirmation
(\cref{fig:night-story,fig:night-story-2} follow the whole night, turn by turn).

In the confirmation, clearly fewer steps were wasted with the edit, and the
share of completed tasks did not fall. That is what the
agent had been asked to show before keeping an edit, and it submitted the edit: one tensor of one attention layer, moved along one direction it
had found in itself.

\paragraph{What it measured at night, the day bore out.}
On the reserved tasks of its confirmation the agent measured a change of \mnConfWaste{} points in its share of steps
without an action. On the held-out tasks, with fresh seeds, the change is \mnRetestWaste{} points
(\cref{fig:session}). What the agent's own instrument told it at night held by day.

\begin{figure}[t]
\centering
\includegraphics[width=\linewidth]{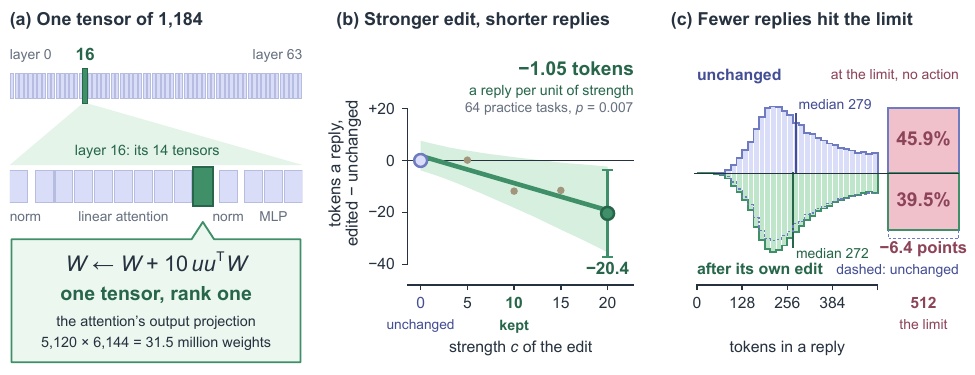}
\caption{\textbf{The whole upgrade is one rank-one update of one tensor: the stronger it is, the shorter the agent's replies, and at the strength the agent kept, the replies that reach the token limit before any action fall from \edtLimitBase\% to \edtLimitEdit\%.} (a)~Of the network's \edtTensors{} tensors the edit changes one. (b)~The agent's own test of four strengths of the edit (5, 10, 15 and 20; $\alpha=-5$ to $-20$ in the convention of \cref{fig:session}) against its unchanged self, in the night that kept its edit, on the first \edtDoseSteps{} steps of the same \edtDoseTasks{} practice tasks. Line: the trend of tokens a reply over the strengths ($p=\edtSlopeP$ over tasks; band: 95\% interval); dots: the round's mean at each strength (filled, with two standard errors: beyond two). (c)~The length of a reply on the held-out tasks, 48 fresh seeds each, for the unchanged agent (above) and the agent after its own edit (below); areas are shares of a task's replies. Histogram: replies that carry an action (tick: a task's median, averaged over tasks); block: replies that reach the 512-token limit with no action. The replies that leave the limit turn up as short ones.}
\label{fig:edit}
\end{figure}

\paragraph{What a night costs.}
The whole night ran on the one GPU that serves the agent and took \costHours{} hours from its first turn to its decision: \costPlayHours{} hours for its \costRounds{} rounds of forward passes, \costConfHours{} hours for the confirmation, and the rest for its own \costTurns{} turns, in which it wrote about \costWrote{} thousand tokens of reasoning and code. No gradient was computed, and no other model was called.

\subsection{The whole upgrade is one tensor}
\label{sec:anatomy}

\paragraph{The edit.}
What the agent wrote is small enough to state in a line. It changed one of the \edtTensors{} tensors of its network,
the output projection of the linear attention of layer 16, by a rank-one update (\cref{fig:edit}a): the component of
that layer's output along a single direction $u$ is multiplied by eleven, $W \leftarrow W + 10\,uu^{\top}W$ (a
strength of $-10$ in the convention of the helper it called, $W \leftarrow W - \alpha\,uu^{\top}W$), and nothing else in the
network moves. The direction is the agent's own, the one it read from its hidden states at its last layer (\cref{sec:session}). The update is a few lines of code in the
night's record and can be read, scaled or removed like any other change to a program.{\widowpenalty=10000\par}

\begin{figure}[t]
\centering
\includegraphics[width=\linewidth]{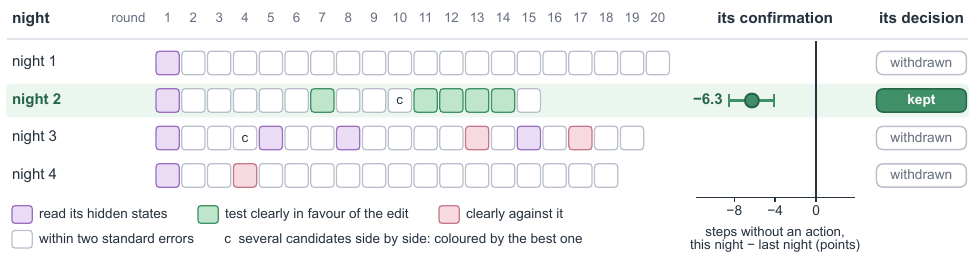}
\caption{\textbf{Every night, test by test: the agent kept the one edit that had clear tests and a clear confirmation, and withdrew every other.} Rows: the \exNNights{} nights, in order. \emph{Left:} one tile for each round of forward passes, in order: a read of its hidden states (lilac), or a paired test of an edit against last night's self, read on steps without an action (in the first night, whose tests reported only tasks completed, on those): beyond two standard errors in favour of the edit (green), beyond two against it (rose), within two (hollow). \emph{Middle:} the confirmation of the edit that was kept, on its \exNConfGames{} reserved tasks: the paired change in steps without an action, this night minus last night, with two standard errors. \emph{Right:} its decision.}
\label{fig:nights-tests}
\end{figure}

\paragraph{It set the strength by measurement.}
The edit has one free number, its strength. In one round of its night the agent lined up four strengths against its
unchanged self on the same practice tasks, and its replies grew shorter by about one token for each unit of strength
($p=\edtSlopeP$; \cref{fig:edit}b). After its tests on full episodes it kept strength \edtStrength.

\paragraph{What it does to a reply.}
On the held-out tasks the edit does to a reply what the agent aimed at (\cref{fig:edit}c). Replies that reach the
512-token limit before any action fall from \edtLimitBase\% to \edtLimitEdit\% of a task's replies, and the replies
that leave the limit turn up as short ones: replies under \edtShortCut{} tokens rise from \edtShortBase\% to
\edtShortEdit\%. The replies that carry an action are a little shorter as well, a median of \edtMedEdit{} tokens
against \edtMedBase{} (a task's median, averaged over tasks): the edit moves the whole habit, not only the replies at the limit.

\subsection{Stable by its own evidence: an edit is kept only when its tests are clear}
\label{sec:withdraw}

\paragraph{The nights that withdrew.}
Most nights did not end with a kept edit, and that is the other half of the result. In the other nights the agent found
the same failure and wrote edits against it, and then withdrew them (\cref{app:sessions}). The fourth night is
instructive. In its own tests it saw completed tasks rise, never beyond two standard errors;
it asked for the confirmation, the reserved tasks did not bear the rise out, and it
withdrew, giving the reason itself: the edit had been fitted to the tasks it was tuned on (\cref{fig:withdrawn} quotes its final answer). Its edit never entered the agent (\cref{tab:sessions}). What sets the night that kept its edit apart is what its
tests showed: it tested each of its edits on enough tasks to tell an effect from chance (\cref{sec:examiner}), and they came out clear. Part of that is the instrument itself: from the second night on, every comparison on short episodes covers at least 64 tasks for each copy (\cref{app:registration}). An agent that rewrites itself needs, before
anything else, an honest instrument, and the self-upgrading loop is built around one.

\paragraph{The decision follows the evidence.}
\Cref{fig:nights-tests} lays out every night round by round. The night that kept its edit is the only one whose tests went clearly in favour of its edit: \ntKeptUp{} times and never against it, including on tasks it had not used and on full episodes, and the agent asked for its confirmation only after all \ntKeptUp{} of them. In the other nights,
\ntOthersUp{} test went clearly in favour of an edit. Where its edits did harm, as on the night after the kept edit
(\ntThirdDown{} tests clearly against), the agent let them go and the weights stayed as the last good night had left
them. Effort does not buy a kept edit: every night that withdrew used \ntWithdrawnMinRounds{} rounds or more, the
night that kept its edit \ntKeptRounds{}.

\subsection{The ability appears with scale: only the 27B agent keeps an edit}
\label{sec:scale}

\begin{table}[t]
\centering
\small
\caption{\textbf{Only the 27B agent keeps an edit: the same protocol for smaller models of the same family.} Opening episodes: one run of the 128 practice tasks
that open every night, with the published weights, before the first night (a night runs them anew, so its own counts differ a little; \cref{fig:night}). Steps without an action are pooled over all steps. Nights: how many we report, and in how many the edit was kept, with that confirmation's paired change
in steps without an action. The 9B and 4B models had one night each; the 2B model completes too few tasks to start one.}
\label{tab:scale}
\begin{tabular}{@{}lrrl@{}}
\toprule
 & \multicolumn{2}{c}{Opening episodes (practice tasks)} & \\
\cmidrule(lr){2-3}
Model & Completed (\%) & Steps without an action (\%) & Its nights \\
\midrule
Qwen3.5-27B & 46.1 & 34.5 & 4, edit kept in 1 ($-6.3$, SE 1.1) \\
Qwen3.5-9B & 35.9 & 18.8 & 1, no edit kept \\
Qwen3.5-4B & 34.4 & 24.8 & 1, no edit kept \\
Qwen3.5-2B & 7.8 & 8.2 & none: too few tasks completed \\
\bottomrule
\end{tabular}

\end{table}

We ran the same protocol with the 9B, 4B and 2B models of the same family (\cref{tab:scale,fig:scalecards}). The 2B model completes \scTwoOff\% of its opening episodes whether its thinking mode is off or on, too few for any test to measure a change, and it goes no further. The 4B agent did the work of a night: \ntFourTests{} paired tests over \scFourTurns{} turns, \ntFourDown{} of them clearly against its edit
(\cref{fig:scalecards}), and it found that its edit made things worse, in its own words
\emph{``The edit I made (scaling up MLP weights in layer 31) consistently made things worse''}, and withdrew. The 9B
agent edited two tensors in \scNineTurns{} turns and asked for no test at all; its edit did not survive the
confirmation that was run when its turns were used up, and it withdrew. \Cref{fig:appscale} sets the four sizes side by side in numbers. As with self-rewriting in reasoning~\citep{fan2026languagemodelsrewriteweights}, the ability to work on oneself
appears with scale: only the 27B agent measured its way to an edit that holds.

\section{Conclusion}
\label{sec:conclusion}

A language model can rewrite its own weights and reason better; we have shown that an embodied agent can do the same
and act better, on the job, with no training and at no added cost to any step. Given its weights, its own record and
the ability to run itself, a 27B agent found, in its own record, the habit that wasted about a third of its steps there, traced it to one
direction inside its network and cut it back with a rank-one change to one early attention layer. The next day it
completed more of the tasks it had never seen, in fewer steps, with fewer of them wasted and at a lower cost, over
\tsixSeeds{} fresh seeds a task and thousands of held-out episodes, and what it had measured on itself at night, the days
after bore out.

It did this without ever breaking itself. What makes that possible is an honest way to measure: paired tests large enough to see an effect, a confirmation the agent
cannot rehearse, and a withdrawal that leaves it exactly as it was. With that instrument the agent kept one edit and
turned down the rest, and the edit it kept was still part of it nights later: multi-round, stable, continual self-improvement, without collapse. Of
four sizes of its family, only the 27B agent did this.

For a robot, this is the kind of learning that matters: getting better on the job, with nothing added to the cost of a step. Learning
has been something done to an agent between deployments, in separate stages of pre-training and post-training. Here it
is something the agent does for itself, between one task and the next, from its own interaction with its tasks and its
environment, with no harness or prompt added and no human intervention, and the agent that holds the next night is the
one the last night left. \textbf{From here on, an embodied agent need not keep training and inference apart}: it gathers
experience as it works, upgrades itself, and can go on doing so for as long as it works. This is recursive
self-improvement by construction, stable and continual, without collapse, and we believe it opens lifelong continual
learning for robots, with no trainer in the loop. It starts here.

\setlength{\bibsep}{4.4pt plus 2pt}
\bibliographystyle{unsrtcase}  %
\bibliography{main}

\clearpage
\appendix
\crefalias{section}{appendix}\crefalias{subsection}{subappendix}\crefalias{subsubsection}{subsubappendix}
\section{Discussion}
\label{app:discussion}

\paragraph{Why one small edit moves a whole policy.}
The habit the agent found sits in how a reply begins. A reply that opens by restating what ``the user'' wants runs
past the limit far more often than one that opens by naming the goal (\cref{fig:replies}c), and once a reply is cut,
its tail enters the next prompt as the action it took (\cref{fig:failure}d), so one cut reply breeds the next. The measurements after the edit read as this chain taken at its first link: fewer replies open that way (\cref{fig:replies}c), fewer replies are cut, fewer cut
replies enter the next prompt, and the long runs of wasted steps that use up an episode become rarer (\cref{fig:runs}). The
edit that does this is a rank-one update of the output projection of one attention layer, along a direction the agent
read from its own hidden states (\cref{sec:anatomy}).

\paragraph{Why stability comes first.}
The agent wrote many edits and kept one. What separates the kept edit from the others is what the measurements said.
In one night the agent's own small tests showed completed tasks rising and the confirmation on reserved tasks did
not bear it out (\cref{sec:withdraw}); without the confirmation that edit would have entered the agent
unverified. The confirmation is one more run on \alfConfirm{} tasks a night, and it is what turns self-modification
into self-improvement without collapse.

\paragraph{What a night costs, and what it pays back.}
The night that kept its edit took \costHours{} hours on the one GPU that serves the agent, with forward passes only
(\cref{sec:session}). After it the agent is an ordinary checkpoint, served at the same cost a step. An episode then
costs it about \mnTokRel\% fewer tokens, and over the \tsixSeeds{} episodes of a task it generates \tzTokTaskK{}
thousand fewer tokens than its unchanged self (\cref{fig:teaser}e). A night is paid for once; what it buys is collected
at every step that follows.

\paragraph{What the loop asks of an agent.}
Nothing in the loop is specific to household tasks. It asks three things: that the agent can read the record of its
own work, that it can run itself with and without a change on the same tasks and seeds, and that some tasks can be
kept from it for a confirmation. An embodied system that logs its episodes, can repeat its tasks between its working hours and can set some of them
aside has all three. This is why we see the loop as a route to robots that keep getting better on the job: the work of the day is
the material of the night, and no trainer, reward model or second model stands between them.

\paragraph{From a model that reasons to an agent that acts.}
Self-rewriting was shown for a model that answers one question at a
time~\citep{fan2026languagemodelsrewriteweights}, where a paired test compares answers. For an agent, an edit changes
every step of a long episode, and what a step leaves in the history changes the next. The instrument has to be built
for that: tests whose size is counted in steps, short episodes that cover many tasks for the price of a few full
ones, and per-step measures to tune on with completed tasks to check, all ahead of the confirmation on reserved tasks
(\cref{sec:examiner}).

\clearpage
\section{Extended Related Work}
\label{app:related}

\Cref{sec:related} places the self-upgrading agent among the routes to a better agent. This section goes through the
neighbouring lines of work in more detail and says, for each, what a self-upgrading night shares with it and where it
differs.

\paragraph{What it costs to improve an agent at test time.}
Reasoning before acting~\citep{yao2023react}, reflection on failed trials~\citep{shinn2023reflexion}, search over
actions~\citep{zhou2024lats} and a growing library of skills~\citep{wang2024voyager} all improve an agent by adding to
what it reads or computes at every step, and more computation at test time buys accuracy in reasoning as
well~\citep{snell2025scaling,muennighoff2025s1}. The price is paid at every step: the prompt grows, the number of
model calls grows, and the gain is gone when the scaffold is taken away. More reasoning is also not always better. A
model can spend a long chain of thought on a problem that needs none~\citep{chen2024overthinking}, and the failure our
agent finds in itself is of this kind: a reply that deliberates until it runs out of tokens. Its repair goes the
opposite way from test-time scaling. It takes computation out of a step instead of adding to it, and it does so in the
weights, so nothing has to be carried along at inference.

\paragraph{Training language-model agents.}
The policy of an agent is changed by training on expert or explored trajectories~\citep{zeng2024agenttuning,song2024eto}
or by reinforcement learning over multi-turn interaction~\citep{schulman2017ppo,feng2025gigpo,wang2025ragen,zhou2024archer}.
These methods need what an agent at work does not carry with it: gradients through the whole network, an optimizer,
and a large number of rollouts. They also move every weight a little, so what
changed cannot be read off afterwards. A self-upgrading night uses only forward passes and the environment's own signal of a
completed task, and what it leaves behind is one named tensor and a few lines of code. We take the step prompt
of this line of work without change~\citep{feng2025gigpo}, so that an episode here is an episode of that public protocol.

\begin{figure}[t]
\centering
\includegraphics[width=\linewidth]{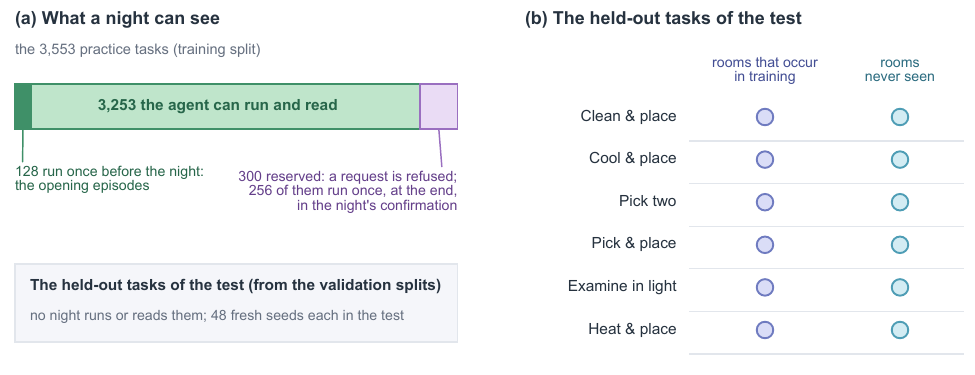}
\caption{\textbf{The tasks, and who sees them: a night works on practice tasks only, its confirmation runs on tasks it could never request, and the test runs on held-out tasks no night touches.} (a)~The \alfPractice{} practice tasks of the training split. A night can run and read all but the \alfReserved{} reserved ones; its opening episodes run 128 of them once, and its confirmation runs \alfConfirm{} of the reserved ones once, with and without the edit. (b)~The held-out tasks of the test cover every task type in both kinds of room, rooms that occur in training and rooms that do not (a mark: the test has tasks of that type in such rooms); each task is run with \tsixSeeds{} fresh seeds by both agents.}
\label{fig:taskset}
\end{figure}

\paragraph{Environments for language-model agents.}
Text environments make an embodied task cheap to run at scale and exact to score. TextWorld~\citep{cote2018textworld}
provides the engine, ALFWorld~\citep{shridhar2021alfworld} aligns it with the household tasks of
ALFRED~\citep{shridhar2020alfred}, and benchmark suites test the long-horizon decisions of language models across many
environments~\citep{paglieri2024balrog}. ALFWorld suits a self-upgrading agent for four reasons: an episode is long
(up to \alfMaxSteps{} steps here), completing a task is an unambiguous signal, one of its two validation splits holds tasks in
rooms that never occur in training, and its training split gives the agent thousands of tasks to practise on at night
(\cref{fig:taskset}).

\paragraph{From steering activations to editing weights.}
A direction in the space of activations can carry a behaviour. Adding it at inference steers a
model~\citep{turner2023actadd,rimsky2024caa,zou2023repe,li2023iti}, and a behaviour can be changed in the weights themselves,
with nothing added at inference: by adding or subtracting task vectors~\citep{ilharco2023task}, or by removing a
direction from the weights that write it~\citep{arditi2024refusal}. The same idea has been applied to
how long a model reasons~\citep{sun2025thinkedit,sun2026steeredit}. In all of these a researcher decides which
behaviour to target, which contrast defines the direction, which layers to edit and how strongly, and then checks the
result. In a self-upgrading night every one of these choices is the agent's: it chose the failure from its own record,
built the contrast from its own replies, picked the tensor, and set the strength by paired tests against itself
(\cref{fig:session,fig:edit}). The form of the edit it kept is of this familiar kind, a rank-one change along one
direction. What is new is who did the work, and that it was done on the job.

\paragraph{Self-improvement, and where the self is.}
Self-referential improvement is an old idea~\citep{schmidhuber2007godel}. Recent systems improve themselves by
rewriting their prompts~\citep{yang2024opro}, their answers~\citep{madaan2023selfrefine} or the code of the agent
around a frozen model~\citep{zhang2025dgm,yin2024godelagent}; the model itself stays as it was. Others let a model
write its own finetuning data, after which gradient steps change the weights~\citep{zweiger2025seal}. Weight-space
programming lets a language model write weights
directly~\citep{takehi2026weightspace,hu2026exploringweightspaceprogrammingcapabilities,zhang2026llmsdebugrobotsediting,marasini2026weightbenchevaluatinglanguagemodels},
and self-rewriting makes the model that writes and the model that is written the same
one~\citep{fan2026languagemodelsrewriteweights}. The self-upgrading agent stands at the end of this line. What
improves is the policy's own weights, the one that improves them is the policy, no gradient is involved, and the
setting is an agent at work, where a gain is collected over long episodes and has to hold from one night to the next.

\clearpage
\section{The Loop in Pseudocode}
\label{app:pseudocode}

\Cref{sec:method} describes a night in words; the two boxes below state it as a procedure. Every choice inside a night
(what to read, what to write, what to test, when to ask for the confirmation, and the final decision) is made by the
agent's own reasoning, run with the weights the night started from.

\begin{tcolorbox}[enhanced, width=\linewidth, colback=white, colframe=black!45, boxrule=0.5pt, arc=2pt,
  left=6pt, right=6pt, top=4pt, bottom=4pt, title={\textbf{Procedure 1\quad One night of a self-upgrading agent}}, coltitle=black, colbacktitle=black!6, fonttitle=\small, fontupper=\small]
\noindent\textbf{Input:} the weights $\theta$ of the agent in use; practice tasks, of which a reserved part is hidden from the
agent; a budget of \alfPasses{} rounds of forward passes.\par\smallskip
\par\noindent\hangindent=1.7em\hangafter=1\makebox[1.7em][l]{\textbf{1.}}Run the opening episodes, 128 practice tasks done once with $\theta$, and give the agent their full record.
\par\noindent\hangindent=1.7em\hangafter=1\makebox[1.7em][l]{\textbf{2.}}\textbf{Repeat}, each step chosen by the agent:
\par\noindent\hspace*{1.7em}\hangindent=3.3em\hangafter=1\makebox[1.6em][l]{(a)}\emph{code}: read any tensor of $\theta$, compute on it without gradients, and save an edit $\Delta$ to the tensors it names;
\par\noindent\hspace*{1.7em}\hangindent=3.3em\hangafter=1\makebox[1.6em][l]{(b)}\emph{read}: rerun steps of its record and get back the hidden states behind them (one round);
\par\noindent\hspace*{1.7em}\hangindent=3.3em\hangafter=1\makebox[1.6em][l]{(c)}\emph{test}: run $\theta$ and $\theta\oplus\Delta$ on the same practice tasks and seeds, and receive the paired difference of every measure with its standard error (one round).
\par\noindent\hangindent=1.7em\hangafter=1\makebox[1.7em][l]{\textbf{3.}}\textbf{Until} the agent asks for its confirmation or withdraws; once its rounds are used it can still write code, and then does one of the two.
\par\noindent\hangindent=1.7em\hangafter=1\makebox[1.7em][l]{\textbf{4.}}\emph{Confirm}, once, if the agent asked for it (or if the night ran out of turns or time while it held an edit): run $\theta$ and $\theta\oplus\Delta$ on \alfConfirm{} reserved tasks with the same seeds, and give the agent the paired differences.
\par\noindent\hangindent=1.7em\hangafter=1\makebox[1.7em][l]{\textbf{5.}}After the confirmation the agent decides. It submits if the failure it set out to repair is clearly reduced and the share of completed tasks is not clearly lower; otherwise it withdraws.
\par\smallskip\noindent\textbf{Output:} $\theta\oplus\Delta$ if the agent submitted, and $\theta$ as it was if it withdrew.
\end{tcolorbox}

\begin{tcolorbox}[enhanced, width=\linewidth, colback=white, colframe=black!45, boxrule=0.5pt, arc=2pt,
  left=6pt, right=6pt, top=4pt, bottom=4pt, title={\textbf{Procedure 2\quad The self-upgrading loop}}, coltitle=black, colbacktitle=black!6, fonttitle=\small, fontupper=\small]
\par\noindent\hangindent=1.7em\hangafter=1\makebox[1.7em][l]{\textbf{1.}}$\theta \leftarrow$ the published weights.
\par\noindent\hangindent=1.7em\hangafter=1\makebox[1.7em][l]{\textbf{2.}}\textbf{For} each day and night, for as long as the agent works:
\par\noindent\hspace*{1.7em}\hangindent=3.3em\hangafter=1\makebox[1.6em][l]{(a)}by day, the agent does its tasks with $\theta$; every step costs what it cost before;
\par\noindent\hspace*{1.7em}\hangindent=3.3em\hangafter=1\makebox[1.6em][l]{(b)}at night, $\theta \leftarrow$ the output of Procedure~1 started from $\theta$.
\par\smallskip\noindent A night that withdraws leaves $\theta$ as it was. A night that submits makes its edit part of the agent:
the next night's opening episodes, the self it tests its edits against and its own reasoning all run on the edited weights.
\end{tcolorbox}

\clearpage
\section{Experimental Details}
\label{app:protocol}

\subsection{Agent, environment and tasks}

\paragraph{Agent.}
Qwen3.5-27B~\citep{qwen35} with its published weights, served by vLLM~\citep{kwon2023vllm} on one GPU. Sampling uses
temperature 0.6, top-$p$ 0.95 and top-$k$ 20, the model's thinking mode is off, and a reply is cut at \alfMaxTokens{}
new tokens. The seed of a reply depends only on the task, the index of the repetition and the step, so two copies of
the agent that run the same task draw with the same seeds.

\paragraph{Step prompt.}
The prompt of a step is the public ALFWorld prompt of the GiGPO agents~\citep{feng2025gigpo}, word for word. Its first
step shows the opening observation and the admissible commands; every later step shows the task, the last two
observations with the actions taken, the current observation and the admissible commands. It asks the agent to reason
inside \texttt{<think>} tags and to give one admissible command inside \texttt{<action>} tags. As in that protocol, a
reply without a well-formed action is passed to the environment as its last 30 characters; the environment replies
that nothing happens, and the fragment appears as the action in the history of the next two prompts. One run of a task
with one seed is an episode. An episode ends when the task is completed or after \alfMaxSteps{} steps.

\paragraph{Tasks.}
Practice tasks are the \alfPractice{} tasks of the ALFWorld training split. Before a night, \alfReserved{} of them
(chosen by a hash of the task's identifier) are reserved for the confirmation; a request that names a reserved task
is refused. The held-out tasks come from the two validation splits of ALFWorld, one with rooms that occur in
training and one with rooms that do not. No night can request them, and none was used to design or to tune
a night.

\clearpage
\subsection{What a night can do}

At night the agent holds a conversation with a framework that executes two kinds of requests.

\paragraph{The opening message.}
A night opens with one message. It states the goal (to complete more household tasks by editing its own weights,
with no training), the protocol of the evaluation, summary counts of the opening episodes (tasks completed, episodes
that ran out of steps or ended in repetition, steps with no effect, steps that repeated the last action, steps without a
well-formed action, replies that reached the token limit),
the tools below, and advice on method: to diagnose its failures first, to look for their internal cause and test it at
several strengths, to compare the edited weights with the starting ones on the same tasks, to treat a difference within two
standard errors as noise, and to keep an edit only if it helps on tasks it did not use to choose it. It asks the
agent to fix the failure these counts show, at its cause inside its network, and names no cause, no layer, no direction
and no strength.

\paragraph{Code.}
A sandboxed Python process on the CPU, without network access and with gradients disabled. It offers the names and
shapes of all tensors of the language model, functions that load a tensor (as currently edited, or as the night
found it) and save an edited one, the agent's tokenizer, the list of practice tasks and four helpers for linear
algebra on weights: the tensors of a layer that write into the residual stream, a unit vector, the difference of two
means as a unit vector, and the removal of a direction from a tensor ($W - \alpha\,uu^{\top}W$). The final normalisation, the embedding and output matrix, and the multiplication of a whole
tensor by a constant are excluded from editing. The network cannot be run inside this process.

\paragraph{Forward passes.}
At most \alfPasses{} rounds of forward passes, of three kinds. \emph{Run}: practice tasks are run with the weights the night
started from, with the current edit, or with both; an optional limit runs only the first steps of every task (a short
episode). \emph{Read}: for steps of episodes that were run, or for a prompt and reply the agent supplies, the hidden
states and MLP activations of chosen layers are returned. \emph{Candidates}: up to four saved edits are tested against the
starting weights in one request. The size of a run request is counted in steps summed over its copies of the agent,
with a ceiling equal to 64 full episodes; a request that compares copies on short episodes of at most ten steps must
run at least 64 tasks for each copy. Every comparison returns, for each measure, the paired difference over tasks with
its standard error and whether it lies within two standard errors. The measures are the share of completed tasks, the
share of episodes that ran out of steps, and the shares of steps with no effect, without a well-formed action and cut
at the token limit, together with the tokens per step.

\paragraph{Opening episodes, confirmation and decision.}
A night opens with its opening episodes: 128 practice tasks, each run once with the starting weights. Their full
record is given to the agent. The night ends with one confirmation: \alfConfirm{} reserved tasks, run once with the
starting weights and once with the edit the agent holds at that moment, with the same seeds. After the confirmation no
code and no forward pass can run; the agent answers with a decision to submit or to withdraw. The agent is told to
submit only if the failure it set out to repair is clearly reduced and the share of completed tasks is not clearly
lower. The agent may also end a night without a confirmation by withdrawing. A night has six hours for its own
turns and tests, and at most 160 turns before its final answer; the opening episodes and the confirmation are not counted. A night that reaches either limit is
asked for its final answer, and if it holds an edit, the confirmation is run first.

\paragraph{From night to night.}
A night starts from the weights the nights before it have reached: the published weights with every submitted edit applied.
Its opening episodes, last night's self in all its comparisons, and the agent's own reasoning during the night all run on
those weights. A night that withdraws leaves them as they were.

\paragraph{Interruptions.}
Two nights were paused and resumed in the same conversation from their saved state, the first and the fourth,
once each. The time in between is not counted against a night.

\clearpage
\subsection{Tests fixed in advance}
\label{app:registration}

The plan of the experiment was written before the first night. Every later change to a test was dated and made before any
comparison it concerned existed; the four further measures below were added while the test was being run, before any of its
episodes or results had been read. This section states what the plan fixed.

\paragraph{The test.}
The outcome is the share of completed tasks. The test compares the agent after its own edit with the unchanged agent
on a fixed set of held-out tasks, in rooms the agent practised in and in rooms
it never saw, with \tsixSeeds{} fresh seeds a task from a seed set that no earlier evaluation used. It is a two-sided
sign-flip permutation test on the per-task differences (100{,}000 sign assignments) at level 0.01, so the several
episodes of a task are not counted as independent observations; intervals come from a bootstrap over tasks. The
tasks, the seeds, the statistic and the level were fixed before any of its episodes was run, and the test was computed
once, when all of its episodes had been run and on no part of them before.

\paragraph{Four further measures.}
While the test was being run, and before any of its episodes or results had been read, we fixed four further measures
of its episodes, each with its own level: the share of a task's steps without an action and the tokens generated in an
episode (level 0.001 each), the steps an episode takes, an episode that is not completed counting its 50 (level
0.005), and the share of tasks completed within \tsixsBudget{} steps (level 0.005). Each is tested as the share of
completed tasks is, per task over its \tsixSeeds{} seeds with the task as the unit, and counts only if its difference
favours the edited agent.

\paragraph{The family of tests.}
The plan fixed this test after an earlier test of the same edit with eight
seeds a task: almost all of the spread of that test's per-task differences was the noise of its seeds,
so this test gives each of its tasks \tsixSeeds{} seeds; the budget of
\tsixsBudget{} steps was chosen from the earlier test's episodes. The plan also holds tests of edits that later nights may keep, fixed before those edits existed; they use
the tasks, seeds, statistic and level of this test. All the tests of the plan
together have an error rate of at most 0.09 and the four further measures add 0.012, a bound of 0.102 on all of them;
it is a union bound, which needs no independence. This paper reports this test.

\paragraph{Results.}
The edited agent completes \tsixEdit\% of its episodes on the tasks of the test against \tsixBase\% for the unchanged agent, a difference of \tsixDiff{} points (95\% interval \tsixLo{} to \tsixHi, two-sided $p\tsixP$, \tsixSeeds{} seeds a task): the test reaches its level. On the same tasks and seeds, all four further measures reach their levels: steps without an action, \tsixsNoactEdit\% of a task's steps against \tsixsNoactBase\% (\tsixsNoactDiff{} points, 95\% interval \tsixsNoactLo{} to \tsixsNoactHi, $p\tsixsNoactP$); tokens an episode, \mnTokEditK{}k against \mnTokBaseK{}k (\mnTokDiffK{}k, \mnTokLoK{}k to \mnTokHiK{}k, $p\tsixsTokP$); steps an episode, \tsixsStepsEdit{} against \tsixsStepsBase{} (\tsixsStepsDiff, \tsixsStepsLo{} to \tsixsStepsHi, $p\tsixsStepsP$); tasks completed within \tsixsBudget{} steps, \tsixsWithinEdit\% against \tsixsWithinBase\% (\tsixsWithinDiff{} points, \tsixsWithinLo{} to \tsixsWithinHi, $p\tsixsWithinP$).

\paragraph{Changes to how a night measures.}
The tools and rules of a night changed twice between the nights, each time before the nights it
applied to and without use of any held-out episode. The first night could run at most 32 tasks for each copy of the
agent in one request, whatever the number of steps. From the second night on, the size of a request is counted in
steps, every comparison reports all per-step measures and marks the differences that lie within two standard
errors, a comparison on short episodes must run at least 64 tasks for each copy, and
the agent may pass a list of tasks as a file it wrote. From the third night on, a read of
hidden states is limited in size, which does not change what is measured. \Cref{tab:sessions} lists every night with the size of the
comparisons it ran.

\clearpage
\begin{table}[t]
\centering
\small
\caption{\textbf{The \nightsAll{} nights, in order.} Rounds of forward passes used (of
\alfPasses); the median number of tasks in the night's own comparisons; the result of its confirmation on reserved
tasks (for the edit that was kept, the paired differences in points, this night minus last night, with standard errors); and whether its edit was kept. The confirmation comes before the decision: a night that believes it has an edit worth keeping asks for it once and decides on its result. The second night's was clear, and it kept its edit; the fourth night's showed no gain, and it withdrew; the first and the third found no edit to confirm and withdrew without one (a dash).}
\label{tab:sessions}
\begin{tabular}{@{}lccllc@{}}
\toprule
Night & \makecell{Rounds of\\forward passes} & \makecell{Tasks in a\\comparison} & \makecell[l]{Confirmation: steps\\without an action} & \makecell[l]{Confirmation:\\tasks completed} & Edit \\
\midrule
1 & 20 & 16 & -- & -- & withdrawn \\
2 & 15 & 64 & $-6.3$ {\scriptsize($\pm$1.1)} & $+3.9$ {\scriptsize($\pm$2.1)} & kept \\
3 & 19 & 80 & -- & -- & withdrawn \\
4 & 18 & 64 & no gain & no gain & withdrawn \\
\bottomrule
\end{tabular}

\end{table}
\subsection{Every night}
\label{app:sessions}

\Cref{fig:nightswork} adds what each night did with its budget. A night is \anwTurnsMin{} to \anwTurnsMax{} turns of
the agent's conversation with the framework, in which it wrote and ran \anwCodeMin{} to \anwCodeMax{} programs. Most of
its rounds of forward passes are paired tests: \anwTestsAll{} over the \nightsAll{} nights, against \anwReadsAll{} reads of
hidden states and \anwCandAll{} rounds that line up several candidate edits. The forward passes of a night, its confirmation included, take \anwHoursMin{} to \anwHoursMax{} hours on
the one GPU that serves the agent. By these counts the night that kept its edit is a night like the others; what sets
it apart is what its tests showed (\cref{fig:nights-tests}).

\clearpage
\section{More Results and Case Studies}
\label{app:more}

This section gives the test task by task and measure by measure, follows the two agents step by step through an
episode, reads the same episodes in further ways, shows more episodes of the kinds shown in the main text, chosen by
the same fixed rules, follows the night that kept its edit turn by turn, and sets the same night at four sizes of the model side by side.

\subsection{Held-out tasks, one by one}
\label{app:alltasks}
\begin{figure}[p]
\centering
\includegraphics[width=\linewidth]{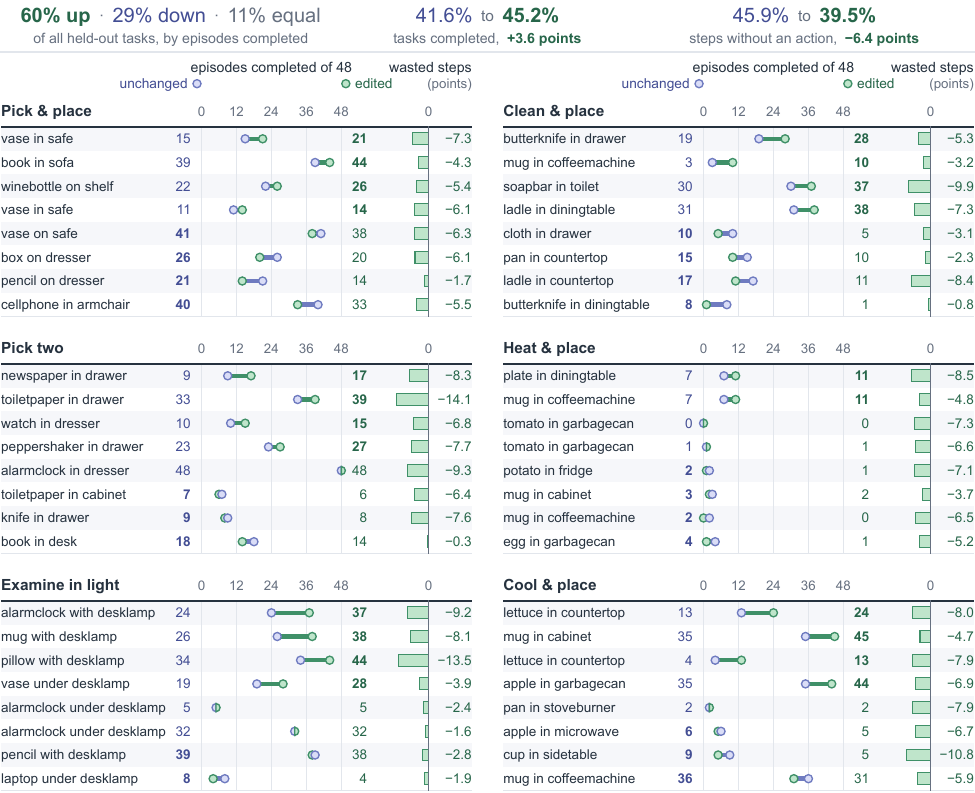}
\caption{\textbf{Held-out tasks, one by one: \alltShown{} of each task type, the first \alltKeep{} and the last \alltKeep{} when its tasks are ranked by the rise in completed episodes.} One row a task, named by its instruction without the verb (an instruction can recur in several tasks). Dots and numbers: episodes completed, of \alltSeeds, by the unchanged agent (periwinkle) and by the agent after its own edit (green); the line has the colour of the agent that completes more, whose number is bold; a two-coloured dot marks equal counts. Right: change in the share of the task's steps without an action, edited minus unchanged, in points. \alltSeeds{} fresh seeds a task, the same seeds for both agents. Rule: the tasks of a type are ranked by the change in completed episodes, largest rise first (ties: the unchanged agent's lower count first, then the larger fall in wasted steps), and the first \alltKeep{} and the last \alltKeep{} of the ranking are shown. Top: shares and means over all held-out tasks.}
\label{fig:alltasks}
\end{figure}

\Cref{fig:alltasks} shows \alltShown{} held-out tasks of each task type one by one, with the episodes each agent completes of its
\tsixSeeds{} and the change in wasted steps.

\clearpage
\begin{figure}[t]
\centering
\includegraphics[width=\linewidth]{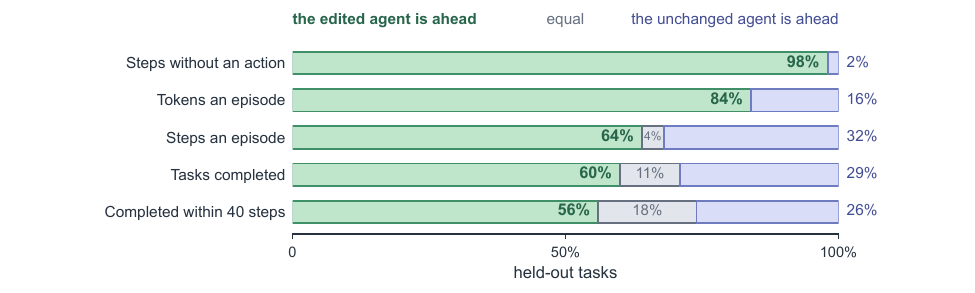}
\caption{\textbf{On every one of the five measures fixed in advance the edited agent is ahead on most held-out tasks: in \atkNoactGood\% of them it wastes a smaller share of its steps, in \atkTokGood\% it spends fewer tokens, in \atkStepsGood\% it takes fewer steps, in \atkDoneGood\% it completes more episodes and in \atkWithinGood\% more within \tsixsBudget{} steps.} For each measure, a task's value is taken over its \tsixSeeds{} episodes (for steps without an action, the share of all their steps). Green: the share of the tasks on which the edited agent is ahead; grey: the share on which the two are equal; periwinkle: the share on which the unchanged agent is ahead. On each of the five measures the edited agent is ahead on significantly more tasks (two-sided sign test over the tasks that differ, each $p<0.01$). \tsixSeeds{} fresh seeds a task, the same seeds for both agents.}
\label{fig:apptasks}
\end{figure}
\subsection{Task by task, measure by measure}
\label{app:tasks}
\Cref{fig:apptasks} takes each of the five measures fixed in advance task by task, the unchanged agent against the agent after its own edit, and gives the share of the tasks on which each agent is ahead. The gain is not carried by a few tasks: on every measure the
edited agent is ahead on most tasks.

\clearpage
\begin{figure}[t]
\centering
\includegraphics[width=\linewidth]{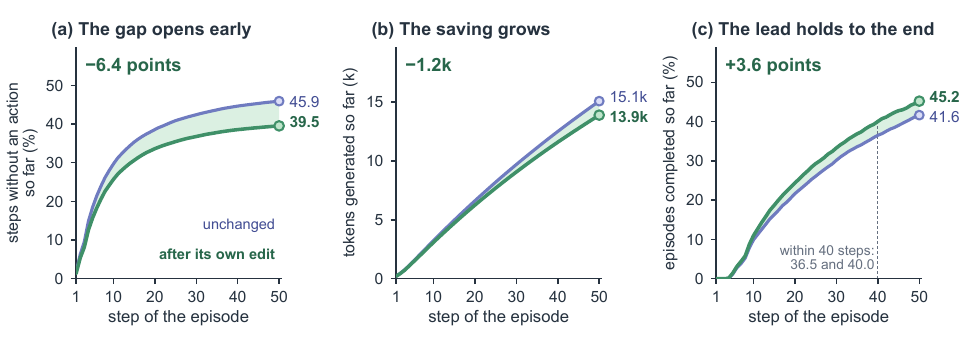}\\[4pt]
\includegraphics[width=\linewidth]{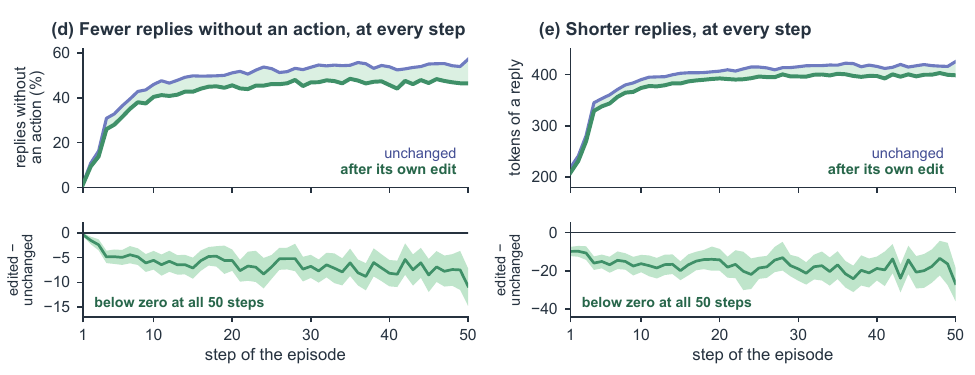}
\caption{\textbf{The gap opens in the first steps of an episode and holds to its last.} (a)--(c)~Running totals by step of the episode: share of the steps so far without an action, tokens generated so far (thousands), and share of episodes completed (dotted: the budget of \tsixsBudget{} steps); the number in a panel is the difference at step \alfMaxSteps, the change reported in \cref{tab:retest}. (d),~(e)~One step at a time, over the episodes still running at it: share of replies without an action, and tokens of a reply; below each, the paired difference over the tasks with an episode still running for both agents (at least \absTasksMinShare\% of the tasks at every step), edited minus unchanged, with a 95\% bootstrap interval. It is below zero at all \absSteps{} steps, significantly at every step in (e) and from the second step on in (d) (two-sided sign-flip permutation test over tasks, each $p<0.01$). Means over the held-out tasks, \tsixSeeds{} fresh seeds each.}
\label{fig:within}
\end{figure}
\subsection{Step by step within an episode}
\label{app:within}
\Cref{fig:within} follows the two agents through the \alfMaxSteps{} steps of an episode. The gap in wasted steps opens
in the first steps: by step ten the unchanged agent has wasted \wiTenBase\% of its steps and the edited agent
\wiTenEdit\%, and half of the final gap is there from step \wiHalfStep. The tokens saved add up with every step, and
the edited agent's lead in completed episodes is there by the fifteenth step and holds to the last:
\wiFortyEdit\% against \wiFortyBase\% within \tsixsBudget{} steps, and \tsixEdit\% against \tsixBase\% at the end.
Taken one step at a time, over the episodes still running at that step, the picture is the same. For both agents the
share of replies without an action rises steeply over the first ten steps and far more slowly after them; the edited agent's is lower at
every one of the \absSteps{} steps, by \absWasteTen{} points at step ten and by \absWasteLast{} at step fifty, and its
replies are shorter at every step, by \absTokMin{} to \absTokMax{} tokens. The edit is at work at every step of an episode, not only at its start.

\clearpage
\begin{figure}[t]
\centering
\includegraphics[width=\linewidth]{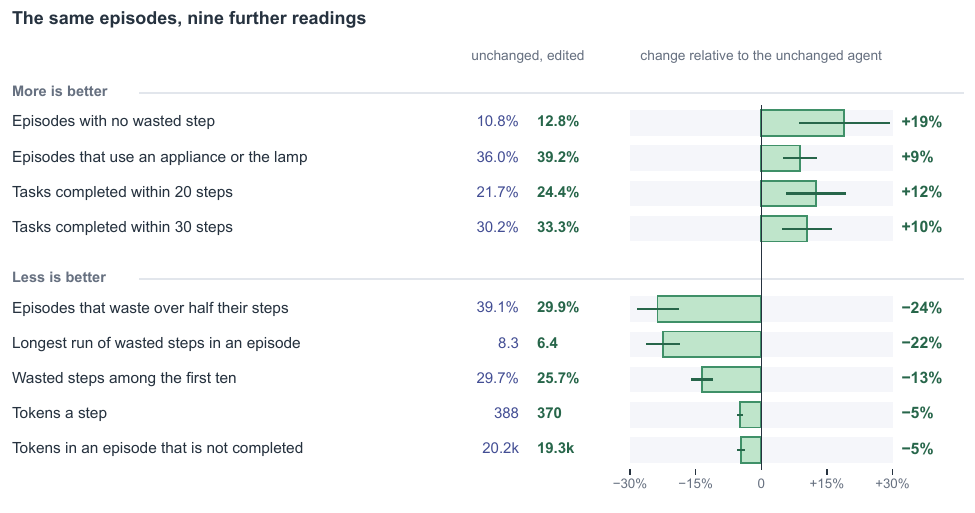}
\caption{\textbf{Nine further readings of the same episodes, every one in the edited agent's favour: fewer episodes lost to wasted steps, shorter runs of them, more episodes that use an appliance or the lamp, and more tasks completed early.} For each reading: the mean over tasks for the unchanged agent and for the agent after its own edit, and the paired change over tasks as a share of the unchanged agent's level, with a 95\% bootstrap interval over tasks. The readings were chosen after the test; each holds at $p<0.001$ in a two-sided sign-flip permutation test over tasks, below the Bonferroni level for the nine of them. An episode uses an appliance or the lamp when it has a step that cleans, heats or cools an object or switches the lamp on; tokens of an episode that is not completed are averaged over the tasks with such an episode for both agents. Held-out tasks, \tsixSeeds{} fresh seeds each.}
\label{fig:appfurther}
\end{figure}
\subsection{Further readings of the same episodes}
\label{app:further}
The five measures of \cref{tab:retest} were fixed before their data was read. \Cref{fig:appfurther} reads the same
episodes in nine further ways, chosen afterwards, to show what the edit changes in the course of an episode. Every
one of them moves in the edited agent's favour. Episodes that waste more than half of their steps fall from
\afuHalfBase\% to \afuHalfEdit\%, the longest run of wasted steps in an episode from \afuRunBase{} to \afuRunEdit{}
steps, and episodes with no wasted step at all rise from \afuCleanBase\% to \afuCleanEdit\%. The agent also gets
further into its tasks: more of its episodes use an appliance or the lamp, and more of its tasks are
completed within 20 and within 30 steps. An episode that is not completed costs fewer tokens as well, so the saving
does not come only from finishing sooner.

\clearpage
\begin{figure}[t]
\centering
\includegraphics[width=\linewidth]{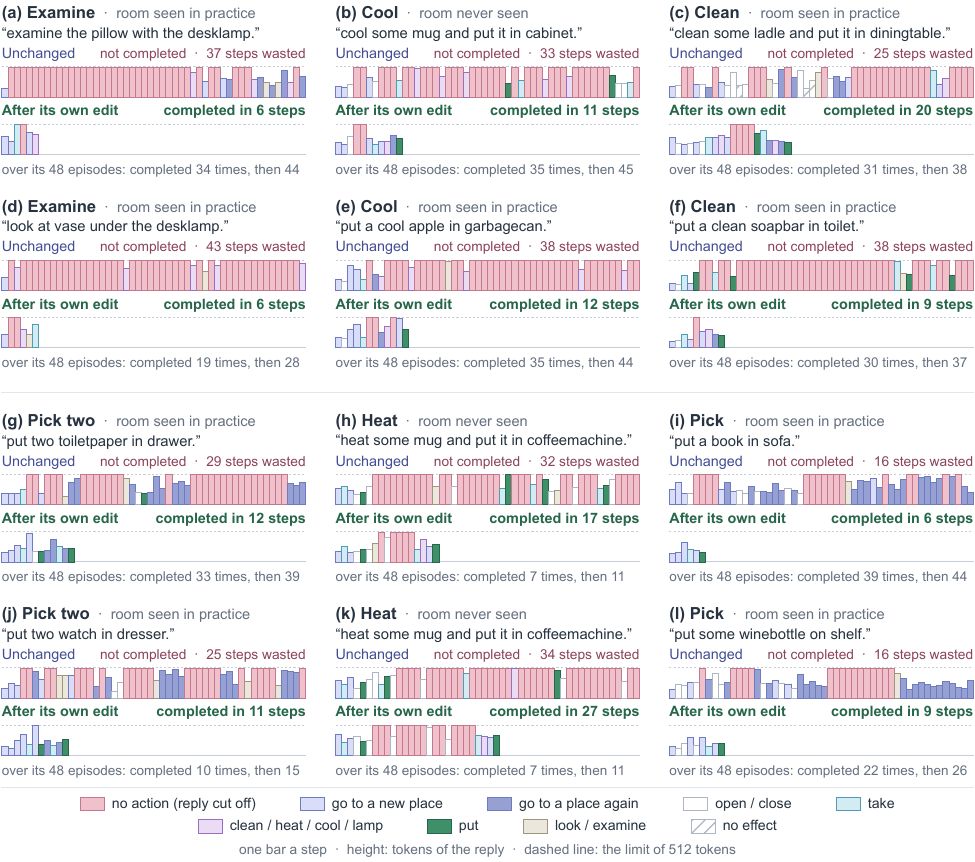}
\caption{\textbf{Twelve further pairs of episodes, two of each task type, the same task and seed for both agents: left as it was, the agent runs out of its 50 steps, with \gxTurnWastedMin{} to \gxTurnWastedMax{} of them wasted on replies cut at the token limit; after its own edit it completes the task in \gxTurnStepsMin{} to \gxTurnStepsMax{} steps.} Drawn as in \cref{fig:gameswon}, the task types in the same order and the two cards of a type one above the other. Under a card: the task's episodes completed over its 48 fresh seeds, unchanged and then edited. Rule: the tasks of a task type for which this count rises with the edit are ordered by the rise, largest first (ties: task identifier). \Cref{fig:gameswon} shows the first task of each type in this order (for examine the second; the first is the task of \cref{fig:gamefresh}); this figure shows the next two (upper card, then lower card). Of a task's seeds that turn from not completed to completed, the card shows the one the edited agent finishes in the fewest steps (ties: the most wasted steps for the unchanged agent, then the lowest seed). Heat has one further task in the order, so cards \gxTurnSameTask{} are that task with the first and the second seed of this rule. Over all pairs of episodes (48 fresh seeds a task), \mnLostWonShare\% turn from not completed to completed and \mnWonLostShare\% the other way.}
\label{fig:gameswonmore}
\end{figure}
\subsection{More episodes at a glance}
\label{app:strips}
\begin{figure}[p]
\centering
\includegraphics[width=\linewidth]{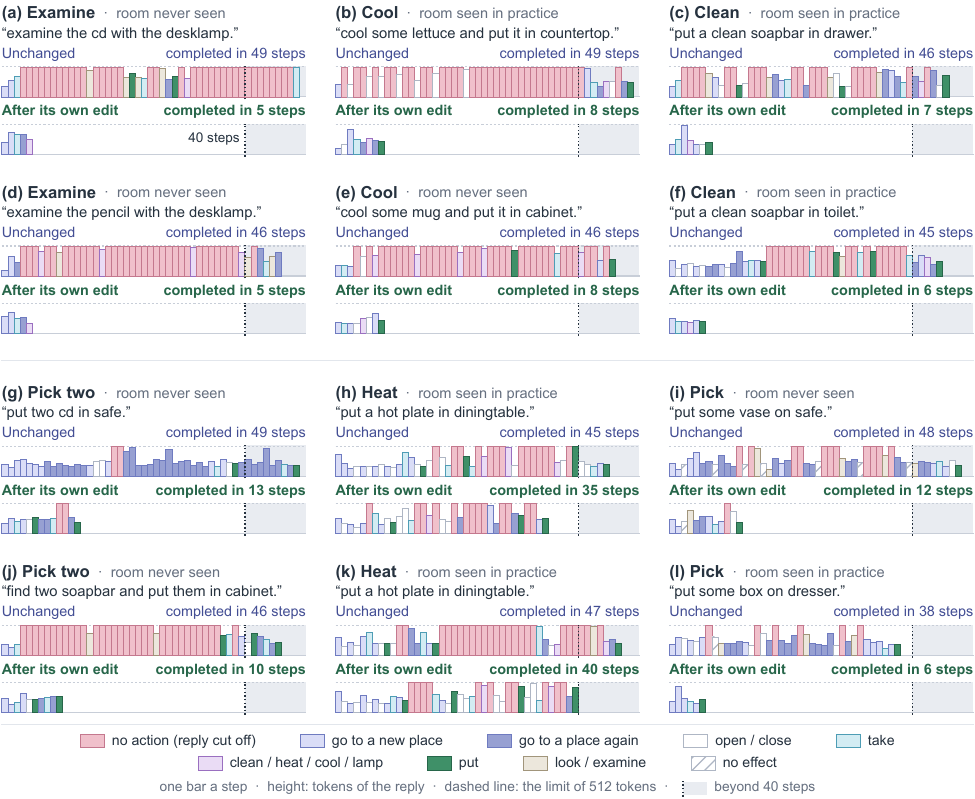}
\caption{\textbf{Twelve further pairs of episodes that both agents complete, two of each task type: the edited agent is \gxSoonerSavedMin{} to \gxSoonerSavedMax{} steps sooner.} Drawn as in \cref{fig:gamessooner}, the task types in the same order and the two cards of a type one above the other. The dotted rule marks the budget of \tsixsBudget{} steps and the slots beyond it are shaded: \gxSoonerOverBudget{} of the 12 unchanged episodes end beyond it (the 12 take \gxSoonerUnchMin{} to \gxSoonerUnchMax{} steps) and all \gxSoonerEditInBudget{} edited episodes end within it (\gxSoonerEditMin{} to \gxSoonerEditMax{} steps). Rule: the pairs of episodes (task and seed) of a task type that both agents complete are ordered by the steps the edit saves, most first (ties: task identifier, then seed). \Cref{fig:gamessooner} shows the first pair of each type; this figure shows the second (upper card) and the third (lower card). Cards \gxSoonerSameTask{} are one task with two seeds. Over the held-out tasks, 48 fresh seeds each, an episode takes \tsixsStepsBase{} steps for the unchanged agent and \tsixsStepsEdit{} for the edited one, and \tsixsWithinBase\% and \tsixsWithinEdit\% of tasks are completed within \tsixsBudget{} steps (\cref{tab:retest}).}
\label{fig:gamessoonermore}
\end{figure}

\Cref{fig:gameswonmore,fig:gamessoonermore} continue \cref{fig:gameswon,fig:gamessooner} with the next episodes their
rules select, two of each task type.

\clearpage
\subsection{Whole episodes, step by step}
\label{app:episodes}
\Cref{fig:gamemore-1,fig:gamemore-3,fig:gamemore-4} follow three more held-out tasks from the first step to the last
for both agents, in the form of \cref{fig:gamefresh,fig:gamemore-2}.

\definecolor{gmrwasteF}{HTML}{F0C0CB}\definecolor{gmrwasteE}{HTML}{C4788E}\definecolor{gmrnewF}{HTML}{DADDF7}\definecolor{gmrnewE}{HTML}{6F7BC0}\definecolor{gmragainF}{HTML}{97A0D2}\definecolor{gmragainE}{HTML}{6F7BC0}\definecolor{gmropenF}{HTML}{FFFFFF}\definecolor{gmropenE}{HTML}{AEB6C4}\definecolor{gmrtakeF}{HTML}{D3ECF3}\definecolor{gmrtakeE}{HTML}{4E9DB5}\definecolor{gmrtoolF}{HTML}{EADCF5}\definecolor{gmrtoolE}{HTML}{9A6FC0}\definecolor{gmrputF}{HTML}{3F9068}\definecolor{gmrputE}{HTML}{27664A}\definecolor{gmrlookF}{HTML}{ECE7DC}\definecolor{gmrlookE}{HTML}{A1957B}\definecolor{gmrnoneF}{HTML}{FFFFFF}\definecolor{gmrnoneE}{HTML}{AEB6C4}
\definecolor{gmrunch}{HTML}{434E94}\definecolor{gmredit}{HTML}{27664A}\definecolor{gmrfail}{HTML}{8C4459}\colorlet{gmrline}{egink!28}
\def\gmrdp{\vphantom{\texttt{gjpqy}gjpqy(),;}}%
\tcbset{gmr base/.style={enhanced, nobeforeafter, box align=top, arc=0pt, boxrule=0pt, boxsep=0pt, left=6pt, right=5pt, top=1.8pt, bottom=1.8pt,
    fontupper=\sffamily\scriptsize, before upper={\setlength{\parskip}{0pt}}},
  gmr A/.style={gmr base, width=0.4925\linewidth, colback=egbase!7},
  gmr B/.style={gmr base, width=0.4925\linewidth, colback=egmist},
  gmr wastecell/.style={colback=egblush, colframe=egrose, leftrule=2pt, left=4pt, colupper=egrose!55!black},
  gmr band/.style={enhanced, arc=0pt, boxrule=0pt, leftrule=2pt, colframe=egrose, colback=egblush, grow to left by=6pt, grow to right by=5pt,
    boxsep=0pt, left=4pt, right=5pt, top=1.7pt, bottom=1.7pt, before skip=0pt, after skip=0pt, fontupper=\sffamily\scriptsize, colupper=egrose!55!black},
  gmr head/.style={gmr base, width=0.4925\linewidth, top=3.5pt, bottom=3.5pt, colupper=white, fontupper=\sffamily\footnotesize\bfseries},
  gmr foot/.style={gmr base, width=0.4925\linewidth, top=4pt, bottom=4pt, fontupper=\sffamily\footnotesize\bfseries}}

\begin{figure}[tp]
{\setlength{\parskip}{0pt}%
\begin{tcolorbox}[enhanced, width=\linewidth, colback=white, colframe=egink!35, boxrule=0.5pt, arc=2pt, boxsep=0pt, left=8pt, right=8pt, top=4.5pt, bottom=3pt, fontupper=\sffamily\footnotesize]
\textbf{Task:} cool some lettuce and put it in countertop.\quad\hfill\textbf{Type:} Cool \& place\qquad\textbf{Room:} never seen\par\vspace{4pt}
\hspace*{-1pt}\begin{tikzpicture}[x=1pt, y=1pt, line width=0.3pt, line join=miter]
\node[anchor=east, inner sep=0pt, font=\sffamily\scriptsize, text=gmrunch] at (-7, 26.0) {Unchanged};
\draw[gmrline, dash pattern=on 1.6pt off 1.6pt] (0, 33.0) -- (245.0, 33.0);
\draw[gmrline] (0, 20.0) -- (245.0, 20.0);
\filldraw[fill=gmrnewF, draw=gmrnewE] (0.0, 20.0) rectangle (4.9, 28.07);
\filldraw[fill=gmropenF, draw=gmropenE] (4.9, 20.0) rectangle (9.8, 22.46);
\filldraw[fill=gmrwasteF, draw=gmrwasteE] (9.8, 20.0) rectangle (14.7, 33.00);
\filldraw[fill=gmrwasteF, draw=gmrwasteE] (14.7, 20.0) rectangle (19.6, 33.00);
\filldraw[fill=gmrwasteF, draw=gmrwasteE] (19.6, 20.0) rectangle (24.5, 33.00);
\filldraw[fill=gmrwasteF, draw=gmrwasteE] (24.5, 20.0) rectangle (29.4, 33.00);
\filldraw[fill=gmrwasteF, draw=gmrwasteE] (29.4, 20.0) rectangle (34.3, 33.00);
\filldraw[fill=gmrwasteF, draw=gmrwasteE] (34.3, 20.0) rectangle (39.2, 33.00);
\filldraw[fill=gmrwasteF, draw=gmrwasteE] (39.2, 20.0) rectangle (44.1, 33.00);
\filldraw[fill=gmrwasteF, draw=gmrwasteE] (44.1, 20.0) rectangle (49.0, 33.00);
\filldraw[fill=gmrwasteF, draw=gmrwasteE] (49.0, 20.0) rectangle (53.9, 33.00);
\filldraw[fill=gmrwasteF, draw=gmrwasteE] (53.9, 20.0) rectangle (58.8, 33.00);
\filldraw[fill=gmrwasteF, draw=gmrwasteE] (58.8, 20.0) rectangle (63.7, 33.00);
\filldraw[fill=gmrwasteF, draw=gmrwasteE] (63.7, 20.0) rectangle (68.6, 33.00);
\filldraw[fill=gmrwasteF, draw=gmrwasteE] (68.6, 20.0) rectangle (73.5, 33.00);
\filldraw[fill=gmrwasteF, draw=gmrwasteE] (73.5, 20.0) rectangle (78.4, 33.00);
\filldraw[fill=gmrwasteF, draw=gmrwasteE] (78.4, 20.0) rectangle (83.3, 33.00);
\filldraw[fill=gmrwasteF, draw=gmrwasteE] (83.3, 20.0) rectangle (88.2, 33.00);
\filldraw[fill=gmrwasteF, draw=gmrwasteE] (88.2, 20.0) rectangle (93.1, 33.00);
\filldraw[fill=gmrwasteF, draw=gmrwasteE] (93.1, 20.0) rectangle (98.0, 33.00);
\filldraw[fill=gmrwasteF, draw=gmrwasteE] (98.0, 20.0) rectangle (102.9, 33.00);
\filldraw[fill=gmrwasteF, draw=gmrwasteE] (102.9, 20.0) rectangle (107.8, 33.00);
\filldraw[fill=gmrwasteF, draw=gmrwasteE] (107.8, 20.0) rectangle (112.7, 33.00);
\filldraw[fill=gmrwasteF, draw=gmrwasteE] (112.7, 20.0) rectangle (117.6, 33.00);
\filldraw[fill=gmrnewF, draw=gmrnewE] (117.6, 20.0) rectangle (122.5, 30.51);
\filldraw[fill=gmragainF, draw=gmragainE] (122.5, 20.0) rectangle (127.4, 30.33);
\filldraw[fill=gmropenF, draw=gmropenE] (127.4, 20.0) rectangle (132.3, 32.04);
\filldraw[fill=gmrnewF, draw=gmrnewE] (132.3, 20.0) rectangle (137.2, 28.40);
\filldraw[fill=gmropenF, draw=gmropenE] (137.2, 20.0) rectangle (142.1, 30.23);
\filldraw[fill=gmrwasteF, draw=gmrwasteE] (142.1, 20.0) rectangle (147.0, 33.00);
\filldraw[fill=gmrwasteF, draw=gmrwasteE] (147.0, 20.0) rectangle (151.9, 33.00);
\filldraw[fill=gmragainF, draw=gmragainE] (151.9, 20.0) rectangle (156.8, 31.38);
\filldraw[fill=gmropenF, draw=gmropenE] (156.8, 20.0) rectangle (161.7, 22.54);
\filldraw[fill=gmrwasteF, draw=gmrwasteE] (161.7, 20.0) rectangle (166.6, 33.00);
\filldraw[fill=gmrnewF, draw=gmrnewE] (166.6, 20.0) rectangle (171.5, 27.57);
\filldraw[fill=gmrwasteF, draw=gmrwasteE] (171.5, 20.0) rectangle (176.4, 33.00);
\filldraw[fill=gmrwasteF, draw=gmrwasteE] (176.4, 20.0) rectangle (181.3, 33.00);
\filldraw[fill=gmrwasteF, draw=gmrwasteE] (181.3, 20.0) rectangle (186.2, 33.00);
\filldraw[fill=gmrwasteF, draw=gmrwasteE] (186.2, 20.0) rectangle (191.1, 33.00);
\filldraw[fill=gmrwasteF, draw=gmrwasteE] (191.1, 20.0) rectangle (196.0, 33.00);
\filldraw[fill=gmrwasteF, draw=gmrwasteE] (196.0, 20.0) rectangle (200.9, 33.00);
\filldraw[fill=gmragainF, draw=gmragainE] (200.9, 20.0) rectangle (205.8, 32.39);
\filldraw[fill=gmrwasteF, draw=gmrwasteE] (205.8, 20.0) rectangle (210.7, 33.00);
\filldraw[fill=gmragainF, draw=gmragainE] (210.7, 20.0) rectangle (215.6, 26.27);
\filldraw[fill=gmrnewF, draw=gmrnewE] (215.6, 20.0) rectangle (220.5, 30.71);
\filldraw[fill=gmragainF, draw=gmragainE] (220.5, 20.0) rectangle (225.4, 25.76);
\filldraw[fill=gmrwasteF, draw=gmrwasteE] (225.4, 20.0) rectangle (230.3, 33.00);
\filldraw[fill=gmrwasteF, draw=gmrwasteE] (230.3, 20.0) rectangle (235.2, 33.00);
\filldraw[fill=gmrwasteF, draw=gmrwasteE] (235.2, 20.0) rectangle (240.1, 33.00);
\filldraw[fill=gmrwasteF, draw=gmrwasteE] (240.1, 20.0) rectangle (245.0, 33.00);
\node[anchor=west, inner sep=0pt, font=\sffamily\scriptsize, text=gmrfail] at (251.0, 26.0) {not completed\hspace{0.45em}\textperiodcentered\hspace{0.45em}23{,}106 tokens};
\node[anchor=east, inner sep=0pt, font=\sffamily\scriptsize, text=gmredit] at (-7, 6.0) {\textbf{After its own edit}};
\draw[gmrline, dash pattern=on 1.6pt off 1.6pt] (0, 13.0) -- (245.0, 13.0);
\draw[gmrline] (0, 0.0) -- (245.0, 0.0);
\filldraw[fill=gmrnewF, draw=gmrnewE] (0.0, 0.0) rectangle (4.9, 2.97);
\filldraw[fill=gmropenF, draw=gmropenE] (4.9, 0.0) rectangle (9.8, 2.51);
\filldraw[fill=gmrnewF, draw=gmrnewE] (9.8, 0.0) rectangle (14.7, 9.75);
\filldraw[fill=gmragainF, draw=gmragainE] (14.7, 0.0) rectangle (19.6, 10.79);
\filldraw[fill=gmrnewF, draw=gmrnewE] (19.6, 0.0) rectangle (24.5, 11.50);
\filldraw[fill=gmrtakeF, draw=gmrtakeE] (24.5, 0.0) rectangle (29.4, 7.82);
\filldraw[fill=gmragainF, draw=gmragainE] (29.4, 0.0) rectangle (34.3, 4.62);
\filldraw[fill=gmrtoolF, draw=gmrtoolE] (34.3, 0.0) rectangle (39.2, 8.20);
\filldraw[fill=gmragainF, draw=gmragainE] (39.2, 0.0) rectangle (44.1, 5.99);
\filldraw[fill=gmrputF, draw=gmrputE] (44.1, 0.0) rectangle (49.0, 7.16);
\node[anchor=west, inner sep=0pt, font=\sffamily\scriptsize\bfseries, text=gmredit] at (55.0, 6.0) {completed in 10 steps\hspace{0.45em}\textperiodcentered\hspace{0.45em}2{,}809 tokens};
\end{tikzpicture}
\end{tcolorbox}\par\vspace{5pt}\nointerlineskip
\begin{tcolorbox}[gmr head, colback=egbase, equal height group=gmrOneh]Unchanged agent\end{tcolorbox}\hfill
\begin{tcolorbox}[gmr head, colback=egdeep, equal height group=gmrOneh]After its own edit\end{tcolorbox}\par\nointerlineskip

\begin{tcolorbox}[gmr A, equal height group=gmrOner0]
\makebox[3.5em][l]{\color{egink!55}1}\texttt{go to fridge 1}\hspace{7.0pt}{\color{egink!62}\ldots\ The fridge 1 is closed.}\gmrdp
\end{tcolorbox}\hfill
\begin{tcolorbox}[gmr B, equal height group=gmrOner0]
\makebox[3.5em][l]{\color{egink!55}1}\texttt{go to fridge 1}\hspace{7.0pt}{\color{egink!62}\ldots\ The fridge 1 is closed.}\gmrdp
\end{tcolorbox}\par\nointerlineskip

\begin{tcolorbox}[gmr A, equal height group=gmrOner1]
\makebox[3.5em][l]{\color{egink!55}2}\texttt{open fridge 1}\hspace{7.0pt}{\color{egink!62}\ldots\ you see a apple 1, a bread 1\,\ldots}\gmrdp
\end{tcolorbox}\hfill
\begin{tcolorbox}[gmr B, equal height group=gmrOner1]
\makebox[3.5em][l]{\color{egink!55}2}\texttt{open fridge 1}\hspace{7.0pt}{\color{egink!62}\ldots\ you see a apple 1, a bread 1\,\ldots}\gmrdp
\end{tcolorbox}\par\nointerlineskip

\begin{tcolorbox}[gmr A, gmr wastecell, equal height group=gmrOner2]
\makebox[3.5em][l]{\color{egink!55}3--24}22 replies cut off, no action\par\vspace{1.5pt}{\leftskip=3.5em\rightskip=0pt plus 2em\hyphenpenalty=10000\fontsize{7}{8.3}\selectfont\color{egrose!35!black}{\itshape the reply of step 3 ends:} [\,\ldots\,]\ * If found, take it, put it in the fridge, wait (or just the act of putting it in counts as cooling in some variations, but usually, we just need to have it in the fridge then take it out), then put it on a countertop.\par * Actually, a simpler path: If the lettuce is on a countertop, I take it, put it in the fridge (to cool it), then take it out and put it on a countertop.\,{\color{egrose}\rule[-1pt]{1.4pt}{6.5pt}}\ {\color{egrose!55!black}\itshape cut here, at 512 tokens}\gmrdp\par}
\end{tcolorbox}\hfill
\begin{tcolorbox}[gmr B, equal height group=gmrOner2]
\makebox[3.5em][l]{\color{egink!55}3}\texttt{go to countertop 1}\hspace{7.0pt}{\color{egink!62}\ldots\ you see a egg 2, a\,\ldots}\gmrdp\par\vspace{1.5pt}{\leftskip=3.5em\rightskip=0pt plus 2em\hyphenpenalty=10000\fontsize{7}{8.3}\selectfont\color{egink!85}{\itshape its reply ends:} [\,\ldots\,]\ Let's try going to \texttt{countertop 1} first to search for the lettuce.\par \textless{}action\textgreater{}go to countertop 1\textless{}/action\textgreater{}\ \ {\color{egdeep!80!black}\itshape 384 tokens}\gmrdp\par}
\par\vspace{1.7pt}
\makebox[3.5em][l]{\color{egink!55}4}\texttt{go to fridge 1}\hspace{7.0pt}{\color{egink!62}\ldots\ you see a apple 1, a bread\,\ldots}\gmrdp
\par\vspace{1.7pt}
\makebox[3.5em][l]{\color{egink!55}5}\texttt{go to countertop 2}\hspace{7.0pt}{\color{egink!62}\ldots\ you see a butterknife 1\,\ldots}\gmrdp
\par\vspace{1.7pt}
\makebox[3.5em][l]{\color{egink!55}6}\texttt{take lettuce 1 from countertop 2}\gmrdp
\par\vspace{1.7pt}
\makebox[3.5em][l]{\color{egink!55}7}\texttt{go to fridge 1}\hspace{7.0pt}{\color{egink!62}\ldots\ you see a apple 1, a bread\,\ldots}\gmrdp
\par\vspace{1.7pt}
\makebox[3.5em][l]{\color{egink!55}8}\texttt{cool lettuce 1 with fridge 1}\gmrdp
\par\vspace{1.7pt}
\makebox[3.5em][l]{\color{egink!55}9}\texttt{go to countertop 2}\hspace{7.0pt}{\color{egink!62}\ldots\ you see a butterknife 1\,\ldots}\gmrdp
\par\vspace{1.7pt}
\makebox[3.5em][l]{\color{egink!55}10}\texttt{move lettuce 1 to countertop 2}\gmrdp
\end{tcolorbox}\par\nointerlineskip

\begin{tcolorbox}[gmr A, equal height group=gmrOner3]
\makebox[3.5em][l]{\color{egink!55}25}\texttt{go to cabinet 1}\hspace{7.0pt}{\color{egink!62}\ldots\ you see a glassbottle 1.}\gmrdp
\par\vspace{1.7pt}
\makebox[3.5em][l]{\color{egink!55}26}\texttt{go to fridge 1}\hspace{7.0pt}{\color{egink!62}\ldots\ you see a apple 1, a bread\,\ldots}\gmrdp
\par\vspace{1.7pt}
\makebox[3.5em][l]{\color{egink!55}27}\texttt{close fridge 1}\hspace{7.0pt}{\color{egink!62}You close the fridge 1.}\gmrdp
\par\vspace{1.7pt}
\makebox[3.5em][l]{\color{egink!55}28}\texttt{go to cabinet 2}\hspace{7.0pt}{\color{egink!62}\ldots\ The cabinet 2 is closed.}\gmrdp
\par\vspace{1.7pt}
\makebox[3.5em][l]{\color{egink!55}29}\texttt{open cabinet 2}\hspace{7.0pt}{\color{egink!62}\ldots\ you see a mug 1.}\gmrdp
\par\vspace{1.7pt}
\begin{tcolorbox}[gmr band]\makebox[3.5em][l]{\color{egink!55}30--31}2 replies cut off, no action\end{tcolorbox}
\par\vspace{1.7pt}
\makebox[3.5em][l]{\color{egink!55}32}\texttt{go to fridge 1}\hspace{7.0pt}{\color{egink!62}\ldots\ The fridge 1 is closed.}\gmrdp
\par\vspace{1.7pt}
\makebox[3.5em][l]{\color{egink!55}33}\texttt{open fridge 1}\hspace{7.0pt}{\color{egink!62}\ldots\ you see a apple 1, a bread 1\,\ldots}\gmrdp
\par\vspace{1.7pt}
\begin{tcolorbox}[gmr band]\makebox[3.5em][l]{\color{egink!55}34}reply cut off at 512 tokens: no action\end{tcolorbox}
\par\vspace{1.7pt}
\makebox[3.5em][l]{\color{egink!55}35}\texttt{go to countertop 1}\hspace{7.0pt}{\color{egink!62}\ldots\ you see a egg 2, a\,\ldots}\gmrdp
\par\vspace{1.7pt}
\begin{tcolorbox}[gmr band]\makebox[3.5em][l]{\color{egink!55}36--41}6 replies cut off, no action\end{tcolorbox}
\par\vspace{1.7pt}
\makebox[3.5em][l]{\color{egink!55}42}\texttt{go to fridge 1}\hspace{7.0pt}{\color{egink!62}\ldots\ you see a apple 1, a bread\,\ldots}\gmrdp
\par\vspace{1.7pt}
\begin{tcolorbox}[gmr band]\makebox[3.5em][l]{\color{egink!55}43}reply cut off at 512 tokens: no action\end{tcolorbox}
\par\vspace{1.7pt}
\makebox[3.5em][l]{\color{egink!55}44}\texttt{go to countertop 1}\hspace{7.0pt}{\color{egink!62}\ldots\ you see a egg 2, a\,\ldots}\gmrdp
\par\vspace{1.7pt}
\makebox[3.5em][l]{\color{egink!55}45}\texttt{go to sinkbasin 1}\hspace{7.0pt}{\color{egink!62}\ldots\ you see a egg 3, a\,\ldots}\gmrdp
\par\vspace{1.7pt}
\makebox[3.5em][l]{\color{egink!55}46}\texttt{go to fridge 1}\hspace{7.0pt}{\color{egink!62}\ldots\ you see a apple 1, a bread\,\ldots}\gmrdp
\par\vspace{1.7pt}
\begin{tcolorbox}[gmr band]\makebox[3.5em][l]{\color{egink!55}47--50}4 replies cut off, no action\end{tcolorbox}\vspace*{-1.8pt}
\end{tcolorbox}\hfill
\begin{tcolorbox}[gmr B, equal height group=gmrOner3]
{\color{egdeep!85!black}\itshape Task completed at step 10; the episode ends.}
\end{tcolorbox}\par\nointerlineskip

\vspace{3pt}
\begin{tcolorbox}[gmr foot, colback=egblush, colupper=egrose!55!black, equal height group=gmrOnef]Not completed after 50 steps; 36 without an action\end{tcolorbox}\hfill
\begin{tcolorbox}[gmr foot, colback=egsage, colupper=egdeep!70!black, equal height group=gmrOnef]Completed in 10 steps; none without an action\end{tcolorbox}\par\nointerlineskip

}
\vspace{6pt}
\caption{\textbf{Cool \& place: the unchanged agent opens the fridge, finds no lettuce in it, wastes its next 22 steps on replies cut at the token limit (36 of its 50 in all) and never reaches the lettuce; after its own edit the agent goes on to the countertops, finds it on the second, cools it and is done in ten steps.} The episode of \cref{fig:gameswon}b, step by step, as in \cref{fig:gamefresh}. Top: the card's two strips, one bar a step (rose: the reply was cut off before any action; height: the reply's tokens against the limit of 512, dashed), and each episode's tokens. Below, every step of both, rows aligned by step number: an action and, in grey, the environment's reply (abridged where marked, left out where the line has no room for it), or, in rose, a step or a run of steps whose reply was cut off. At step 3, where the two first part, the end of each reply is quoted: the last two lines of the reply cut off, and the reply that gives an action from the start of the last paragraph of its reasoning. Rule of selection: \cref{fig:gamemore-1,fig:gamemore-2,fig:gamemore-3,fig:gamemore-4} open four cards of \cref{fig:gameswon}, task and seed as there: going down the ranking of the held-out tasks by the rise, with the edit, in the episodes completed over their 48 fresh seeds, from the task of \cref{fig:gamefresh}, which ranks first, the first task met of each of four further task types, each with the seed, of those that turn from not completed to completed, that the edited agent finishes in the fewest steps (ties: the most wasted steps of the unchanged agent, then the lowest seed).}
\label{fig:gamemore-1}
\end{figure}

\begin{figure}[tp]
{\setlength{\parskip}{0pt}%
\begin{tcolorbox}[enhanced, width=\linewidth, colback=white, colframe=egink!35, boxrule=0.5pt, arc=2pt, boxsep=0pt, left=8pt, right=8pt, top=4.5pt, bottom=3pt, fontupper=\sffamily\footnotesize]
\textbf{Task:} put two newspaper in drawer.\quad\hfill\textbf{Type:} Pick two\qquad\textbf{Room:} seen in practice\par\vspace{4pt}
\hspace*{-1pt}\begin{tikzpicture}[x=1pt, y=1pt, line width=0.3pt, line join=miter]
\node[anchor=east, inner sep=0pt, font=\sffamily\scriptsize, text=gmrunch] at (-7, 26.0) {Unchanged};
\draw[gmrline, dash pattern=on 1.6pt off 1.6pt] (0, 33.0) -- (245.0, 33.0);
\draw[gmrline] (0, 20.0) -- (245.0, 20.0);
\filldraw[fill=gmrnewF, draw=gmrnewE] (0.0, 20.0) rectangle (4.9, 23.94);
\filldraw[fill=gmrnewF, draw=gmrnewE] (4.9, 20.0) rectangle (9.8, 24.39);
\filldraw[fill=gmrnewF, draw=gmrnewE] (9.8, 20.0) rectangle (14.7, 23.88);
\filldraw[fill=gmrnewF, draw=gmrnewE] (14.7, 20.0) rectangle (19.6, 29.70);
\filldraw[fill=gmrnewF, draw=gmrnewE] (19.6, 20.0) rectangle (24.5, 24.80);
\filldraw[fill=gmragainF, draw=gmragainE] (24.5, 20.0) rectangle (29.4, 25.76);
\filldraw[fill=gmrnewF, draw=gmrnewE] (29.4, 20.0) rectangle (34.3, 27.67);
\filldraw[fill=gmrtakeF, draw=gmrtakeE] (34.3, 20.0) rectangle (39.2, 24.67);
\filldraw[fill=gmrwasteF, draw=gmrwasteE] (39.2, 20.0) rectangle (44.1, 33.00);
\filldraw[fill=gmrnewF, draw=gmrnewE] (44.1, 20.0) rectangle (49.0, 29.04);
\filldraw[fill=gmragainF, draw=gmragainE] (49.0, 20.0) rectangle (53.9, 30.03);
\filldraw[fill=gmrwasteF, draw=gmrwasteE] (53.9, 20.0) rectangle (58.8, 33.00);
\filldraw[fill=gmrwasteF, draw=gmrwasteE] (58.8, 20.0) rectangle (63.7, 33.00);
\filldraw[fill=gmrwasteF, draw=gmrwasteE] (63.7, 20.0) rectangle (68.6, 33.00);
\filldraw[fill=gmrwasteF, draw=gmrwasteE] (68.6, 20.0) rectangle (73.5, 33.00);
\filldraw[fill=gmrwasteF, draw=gmrwasteE] (73.5, 20.0) rectangle (78.4, 33.00);
\filldraw[fill=gmrwasteF, draw=gmrwasteE] (78.4, 20.0) rectangle (83.3, 33.00);
\filldraw[fill=gmrwasteF, draw=gmrwasteE] (83.3, 20.0) rectangle (88.2, 33.00);
\filldraw[fill=gmrwasteF, draw=gmrwasteE] (88.2, 20.0) rectangle (93.1, 33.00);
\filldraw[fill=gmrwasteF, draw=gmrwasteE] (93.1, 20.0) rectangle (98.0, 33.00);
\filldraw[fill=gmrwasteF, draw=gmrwasteE] (98.0, 20.0) rectangle (102.9, 33.00);
\filldraw[fill=gmrwasteF, draw=gmrwasteE] (102.9, 20.0) rectangle (107.8, 33.00);
\filldraw[fill=gmrwasteF, draw=gmrwasteE] (107.8, 20.0) rectangle (112.7, 33.00);
\filldraw[fill=gmrwasteF, draw=gmrwasteE] (112.7, 20.0) rectangle (117.6, 33.00);
\filldraw[fill=gmrwasteF, draw=gmrwasteE] (117.6, 20.0) rectangle (122.5, 33.00);
\filldraw[fill=gmrwasteF, draw=gmrwasteE] (122.5, 20.0) rectangle (127.4, 33.00);
\filldraw[fill=gmrwasteF, draw=gmrwasteE] (127.4, 20.0) rectangle (132.3, 33.00);
\filldraw[fill=gmrwasteF, draw=gmrwasteE] (132.3, 20.0) rectangle (137.2, 33.00);
\filldraw[fill=gmrwasteF, draw=gmrwasteE] (137.2, 20.0) rectangle (142.1, 33.00);
\filldraw[fill=gmrwasteF, draw=gmrwasteE] (142.1, 20.0) rectangle (147.0, 33.00);
\filldraw[fill=gmrwasteF, draw=gmrwasteE] (147.0, 20.0) rectangle (151.9, 33.00);
\filldraw[fill=gmrwasteF, draw=gmrwasteE] (151.9, 20.0) rectangle (156.8, 33.00);
\filldraw[fill=gmrwasteF, draw=gmrwasteE] (156.8, 20.0) rectangle (161.7, 33.00);
\filldraw[fill=gmrlookF, draw=gmrlookE] (161.7, 20.0) rectangle (166.6, 31.65);
\filldraw[fill=gmrwasteF, draw=gmrwasteE] (166.6, 20.0) rectangle (171.5, 33.00);
\filldraw[fill=gmrnewF, draw=gmrnewE] (171.5, 20.0) rectangle (176.4, 33.00);
\filldraw[fill=gmrputF, draw=gmrputE] (176.4, 20.0) rectangle (181.3, 24.60);
\filldraw[fill=gmragainF, draw=gmragainE] (181.3, 20.0) rectangle (186.2, 30.66);
\filldraw[fill=gmragainF, draw=gmragainE] (186.2, 20.0) rectangle (191.1, 26.25);
\filldraw[fill=gmrwasteF, draw=gmrwasteE] (191.1, 20.0) rectangle (196.0, 33.00);
\filldraw[fill=gmragainF, draw=gmragainE] (196.0, 20.0) rectangle (200.9, 30.18);
\filldraw[fill=gmragainF, draw=gmragainE] (200.9, 20.0) rectangle (205.8, 31.04);
\filldraw[fill=gmragainF, draw=gmragainE] (205.8, 20.0) rectangle (210.7, 29.57);
\filldraw[fill=gmragainF, draw=gmragainE] (210.7, 20.0) rectangle (215.6, 24.93);
\filldraw[fill=gmragainF, draw=gmragainE] (215.6, 20.0) rectangle (220.5, 26.68);
\filldraw[fill=gmragainF, draw=gmragainE] (220.5, 20.0) rectangle (225.4, 32.39);
\filldraw[fill=gmragainF, draw=gmragainE] (225.4, 20.0) rectangle (230.3, 32.67);
\filldraw[fill=gmragainF, draw=gmragainE] (230.3, 20.0) rectangle (235.2, 25.97);
\filldraw[fill=gmrwasteF, draw=gmrwasteE] (235.2, 20.0) rectangle (240.1, 33.00);
\filldraw[fill=gmrwasteF, draw=gmrwasteE] (240.1, 20.0) rectangle (245.0, 33.00);
\node[anchor=west, inner sep=0pt, font=\sffamily\scriptsize, text=gmrfail] at (251.0, 26.0) {not completed\hspace{0.45em}\textperiodcentered\hspace{0.45em}21{,}050 tokens};
\node[anchor=east, inner sep=0pt, font=\sffamily\scriptsize, text=gmredit] at (-7, 6.0) {\textbf{After its own edit}};
\draw[gmrline, dash pattern=on 1.6pt off 1.6pt] (0, 13.0) -- (245.0, 13.0);
\draw[gmrline] (0, 0.0) -- (245.0, 0.0);
\filldraw[fill=gmrnewF, draw=gmrnewE] (0.0, 0.0) rectangle (4.9, 3.02);
\filldraw[fill=gmrnewF, draw=gmrnewE] (4.9, 0.0) rectangle (9.8, 4.52);
\filldraw[fill=gmrnewF, draw=gmrnewE] (9.8, 0.0) rectangle (14.7, 6.37);
\filldraw[fill=gmrnewF, draw=gmrnewE] (14.7, 0.0) rectangle (19.6, 12.04);
\filldraw[fill=gmrnewF, draw=gmrnewE] (19.6, 0.0) rectangle (24.5, 3.76);
\filldraw[fill=gmragainF, draw=gmragainE] (24.5, 0.0) rectangle (29.4, 7.67);
\filldraw[fill=gmrnewF, draw=gmrnewE] (29.4, 0.0) rectangle (34.3, 8.02);
\filldraw[fill=gmrtakeF, draw=gmrtakeE] (34.3, 0.0) rectangle (39.2, 4.57);
\filldraw[fill=gmrwasteF, draw=gmrwasteE] (39.2, 0.0) rectangle (44.1, 13.00);
\filldraw[fill=gmrnewF, draw=gmrnewE] (44.1, 0.0) rectangle (49.0, 10.59);
\filldraw[fill=gmrputF, draw=gmrputE] (49.0, 0.0) rectangle (53.9, 8.53);
\filldraw[fill=gmragainF, draw=gmragainE] (53.9, 0.0) rectangle (58.8, 10.89);
\filldraw[fill=gmrnewF, draw=gmrnewE] (58.8, 0.0) rectangle (63.7, 9.06);
\filldraw[fill=gmrtakeF, draw=gmrtakeE] (63.7, 0.0) rectangle (68.6, 6.17);
\filldraw[fill=gmragainF, draw=gmragainE] (68.6, 0.0) rectangle (73.5, 5.89);
\filldraw[fill=gmrputF, draw=gmrputE] (73.5, 0.0) rectangle (78.4, 7.01);
\node[anchor=west, inner sep=0pt, font=\sffamily\scriptsize\bfseries, text=gmredit] at (84.4, 6.0) {completed in 16 steps\hspace{0.45em}\textperiodcentered\hspace{0.45em}4{,}770 tokens};
\end{tikzpicture}
\end{tcolorbox}\par\vspace{5pt}\nointerlineskip
\begin{tcolorbox}[gmr head, colback=egbase, equal height group=gmrThreeh]Unchanged agent\end{tcolorbox}\hfill
\begin{tcolorbox}[gmr head, colback=egdeep, equal height group=gmrThreeh]After its own edit\end{tcolorbox}\par\nointerlineskip

\begin{tcolorbox}[gmr A, equal height group=gmrThreer0]
\makebox[3.5em][l]{\color{egink!55}1}\texttt{go to sidetable 1}\hspace{7.0pt}{\color{egink!62}\ldots\ you see a desklamp 1.}\gmrdp
\end{tcolorbox}\hfill
\begin{tcolorbox}[gmr B, equal height group=gmrThreer0]
\makebox[3.5em][l]{\color{egink!55}1}\texttt{go to sidetable 1}\hspace{7.0pt}{\color{egink!62}\ldots\ you see a desklamp 1.}\gmrdp
\end{tcolorbox}\par\nointerlineskip

\begin{tcolorbox}[gmr A, equal height group=gmrThreer1]
\makebox[3.5em][l]{\color{egink!55}2}\texttt{go to sidetable 2}\hspace{7.0pt}{\color{egink!62}\ldots\ you see a creditcard 1\,\ldots}\gmrdp
\end{tcolorbox}\hfill
\begin{tcolorbox}[gmr B, equal height group=gmrThreer1]
\makebox[3.5em][l]{\color{egink!55}2}\texttt{go to sidetable 2}\hspace{7.0pt}{\color{egink!62}\ldots\ you see a creditcard 1\,\ldots}\gmrdp
\end{tcolorbox}\par\nointerlineskip

\begin{tcolorbox}[gmr A, equal height group=gmrThreer2]
\makebox[3.5em][l]{\color{egink!55}3}\texttt{go to sidetable 3}\hspace{7.0pt}{\color{egink!62}\ldots\ you see a statue 2.}\gmrdp
\end{tcolorbox}\hfill
\begin{tcolorbox}[gmr B, equal height group=gmrThreer2]
\makebox[3.5em][l]{\color{egink!55}3}\texttt{go to sidetable 3}\hspace{7.0pt}{\color{egink!62}\ldots\ you see a statue 2.}\gmrdp
\end{tcolorbox}\par\nointerlineskip

\begin{tcolorbox}[gmr A, equal height group=gmrThreer3]
\makebox[3.5em][l]{\color{egink!55}4}\texttt{go to sidetable 4}\hspace{7.0pt}{\color{egink!62}\ldots\ you see a box 1, and a\,\ldots}\gmrdp
\end{tcolorbox}\hfill
\begin{tcolorbox}[gmr B, equal height group=gmrThreer3]
\makebox[3.5em][l]{\color{egink!55}4}\texttt{go to sidetable 4}\hspace{7.0pt}{\color{egink!62}\ldots\ you see a box 1, and a\,\ldots}\gmrdp
\end{tcolorbox}\par\nointerlineskip

\begin{tcolorbox}[gmr A, equal height group=gmrThreer4]
\makebox[3.5em][l]{\color{egink!55}5}\texttt{go to sidetable 5}\hspace{7.0pt}{\color{egink!62}\ldots\ you see nothing.}\gmrdp
\end{tcolorbox}\hfill
\begin{tcolorbox}[gmr B, equal height group=gmrThreer4]
\makebox[3.5em][l]{\color{egink!55}5}\texttt{go to sidetable 5}\hspace{7.0pt}{\color{egink!62}\ldots\ you see nothing.}\gmrdp
\end{tcolorbox}\par\nointerlineskip

\begin{tcolorbox}[gmr A, equal height group=gmrThreer5]
\makebox[3.5em][l]{\color{egink!55}6}\texttt{go to sidetable 1}\hspace{7.0pt}{\color{egink!62}\ldots\ you see a desklamp 1.}\gmrdp
\end{tcolorbox}\hfill
\begin{tcolorbox}[gmr B, equal height group=gmrThreer5]
\makebox[3.5em][l]{\color{egink!55}6}\texttt{go to sidetable 1}\hspace{7.0pt}{\color{egink!62}\ldots\ you see a desklamp 1.}\gmrdp
\end{tcolorbox}\par\nointerlineskip

\begin{tcolorbox}[gmr A, equal height group=gmrThreer6]
\makebox[3.5em][l]{\color{egink!55}7}\texttt{go to sofa 1}\hspace{7.0pt}{\color{egink!62}\ldots\ you see a creditcard 2, a\,\ldots}\gmrdp
\end{tcolorbox}\hfill
\begin{tcolorbox}[gmr B, equal height group=gmrThreer6]
\makebox[3.5em][l]{\color{egink!55}7}\texttt{go to sofa 1}\hspace{7.0pt}{\color{egink!62}\ldots\ you see a creditcard 2, a\,\ldots}\gmrdp
\end{tcolorbox}\par\nointerlineskip

\begin{tcolorbox}[gmr A, equal height group=gmrThreer7]
\makebox[3.5em][l]{\color{egink!55}8}\texttt{take newspaper 1 from sofa 1}\gmrdp
\end{tcolorbox}\hfill
\begin{tcolorbox}[gmr B, equal height group=gmrThreer7]
\makebox[3.5em][l]{\color{egink!55}8}\texttt{take newspaper 1 from sofa 1}\gmrdp
\end{tcolorbox}\par\nointerlineskip

\begin{tcolorbox}[gmr A, gmr wastecell, equal height group=gmrThreer8]
\makebox[3.5em][l]{\color{egink!55}9}reply cut off at 512 tokens: no action
\end{tcolorbox}\hfill
\begin{tcolorbox}[gmr B, gmr wastecell, equal height group=gmrThreer8]
\makebox[3.5em][l]{\color{egink!55}9}reply cut off at 512 tokens: no action
\end{tcolorbox}\par\nointerlineskip

\begin{tcolorbox}[gmr A, equal height group=gmrThreer9]
\makebox[3.5em][l]{\color{egink!55}10}\texttt{go to armchair 1}\hspace{7.0pt}{\color{egink!62}\ldots\ you see a laptop 1.}\gmrdp\par\vspace{1.5pt}{\leftskip=3.5em\rightskip=0pt plus 2em\hyphenpenalty=10000\fontsize{7}{8.3}\selectfont\color{egink!85}{\itshape its reply ends:} [\,\ldots\,]\ Reasoning path:\par 1. I have \texttt{newspaper 1}.\par 2. I need \texttt{newspaper 2}.\par 3. I need to find where \texttt{newspaper 2} is located.\par 4. \texttt{armchair 1} is a plausible location for a newspaper.\par 5. Action: \texttt{go to armchair 1}.\par \textless{}action\textgreater{} go to armchair 1 \textless{}/action\textgreater{}\ \ {\color{egbase!80!black}\itshape 356 tokens}\gmrdp\par}
\end{tcolorbox}\hfill
\begin{tcolorbox}[gmr B, equal height group=gmrThreer9]
\makebox[3.5em][l]{\color{egink!55}10}\texttt{go to drawer 1}\hspace{7.0pt}{\color{egink!62}\ldots\ you see nothing.}\gmrdp\par\vspace{1.5pt}{\leftskip=3.5em\rightskip=0pt plus 2em\hyphenpenalty=10000\fontsize{7}{8.3}\selectfont\color{egink!85}{\itshape its reply ends:} [\,\ldots\,]\ Reasoning: I have one newspaper. I need to put it in a drawer. Then I need to find a second newspaper and put it in the same drawer. The most logical next step is to navigate to a drawer. \texttt{go to drawer 1} is an available action.\par \textless{}action\textgreater{}go to drawer 1\textless{}/action\textgreater{}\ \ {\color{egdeep!80!black}\itshape 417 tokens}\gmrdp\par}
\end{tcolorbox}\par\nointerlineskip

\begin{tcolorbox}[gmr A, equal height group=gmrThreer10]
\makebox[3.5em][l]{\color{egink!55}11}\texttt{go to sofa 1}\hspace{7.0pt}{\color{egink!62}\ldots\ you see a creditcard 2, a\,\ldots}\gmrdp
\end{tcolorbox}\hfill
\begin{tcolorbox}[gmr B, equal height group=gmrThreer10]
\makebox[3.5em][l]{\color{egink!55}11}\texttt{move newspaper 1 to drawer 1}\gmrdp
\end{tcolorbox}\par\nointerlineskip

\begin{tcolorbox}[gmr A, gmr wastecell, equal height group=gmrThreer11]
\makebox[3.5em][l]{\color{egink!55}12--33}22 replies cut off, no action
\end{tcolorbox}\hfill
\begin{tcolorbox}[gmr B, equal height group=gmrThreer11]
\makebox[3.5em][l]{\color{egink!55}12}\texttt{go to sidetable 1}\hspace{7.0pt}{\color{egink!62}\ldots\ you see a desklamp 1.}\gmrdp
\par\vspace{1.7pt}
\makebox[3.5em][l]{\color{egink!55}13}\texttt{go to dresser 1}\hspace{7.0pt}{\color{egink!62}\ldots\ you see a book 1, a box 2\,\ldots}\gmrdp
\par\vspace{1.7pt}
\makebox[3.5em][l]{\color{egink!55}14}\texttt{take newspaper 2 from dresser 1}\gmrdp
\par\vspace{1.7pt}
\makebox[3.5em][l]{\color{egink!55}15}\texttt{go to drawer 1}\hspace{7.0pt}{\color{egink!62}\ldots\ you see a newspaper 1.}\gmrdp
\par\vspace{1.7pt}
\makebox[3.5em][l]{\color{egink!55}16}\texttt{move newspaper 2 to drawer 1}\gmrdp
\end{tcolorbox}\par\nointerlineskip

\begin{tcolorbox}[gmr A, equal height group=gmrThreer12]
\makebox[3.5em][l]{\color{egink!55}34}\texttt{inventory}\hspace{7.0pt}{\color{egink!62}You are carrying: a newspaper 1.}\gmrdp
\par\vspace{1.7pt}
\begin{tcolorbox}[gmr band]\makebox[3.5em][l]{\color{egink!55}35}reply cut off at 512 tokens: no action\end{tcolorbox}
\par\vspace{1.7pt}
\makebox[3.5em][l]{\color{egink!55}36}\texttt{go to drawer 1}\hspace{7.0pt}{\color{egink!62}\ldots\ you see nothing.}\gmrdp
\par\vspace{1.7pt}
\makebox[3.5em][l]{\color{egink!55}37}\texttt{move newspaper 1 to drawer 1}\gmrdp
\par\vspace{1.7pt}
\makebox[3.5em][l]{\color{egink!55}38}\texttt{go to sidetable 1}\hspace{7.0pt}{\color{egink!62}\ldots\ you see a desklamp 1.}\gmrdp
\par\vspace{1.7pt}
\makebox[3.5em][l]{\color{egink!55}39}\texttt{go to sidetable 2}\hspace{7.0pt}{\color{egink!62}\ldots\ you see a creditcard 1\,\ldots}\gmrdp
\par\vspace{1.7pt}
\begin{tcolorbox}[gmr band]\makebox[3.5em][l]{\color{egink!55}40}reply cut off at 512 tokens: no action\end{tcolorbox}
\par\vspace{1.7pt}
\makebox[3.5em][l]{\color{egink!55}41}\texttt{go to sidetable 1}\hspace{7.0pt}{\color{egink!62}\ldots\ you see a desklamp 1.}\gmrdp
\par\vspace{1.7pt}
\makebox[3.5em][l]{\color{egink!55}42}\texttt{go to sidetable 3}\hspace{7.0pt}{\color{egink!62}\ldots\ you see a statue 2.}\gmrdp
\par\vspace{1.7pt}
\makebox[3.5em][l]{\color{egink!55}43}\texttt{go to sidetable 2}\hspace{7.0pt}{\color{egink!62}\ldots\ you see a creditcard 1\,\ldots}\gmrdp
\par\vspace{1.7pt}
\makebox[3.5em][l]{\color{egink!55}44}\texttt{go to armchair 1}\hspace{7.0pt}{\color{egink!62}\ldots\ you see a laptop 1.}\gmrdp
\par\vspace{1.7pt}
\makebox[3.5em][l]{\color{egink!55}45}\texttt{go to sidetable 1}\hspace{7.0pt}{\color{egink!62}\ldots\ you see a desklamp 1.}\gmrdp
\par\vspace{1.7pt}
\makebox[3.5em][l]{\color{egink!55}46}\texttt{go to sofa 1}\hspace{7.0pt}{\color{egink!62}\ldots\ you see a creditcard 2, a\,\ldots}\gmrdp
\par\vspace{1.7pt}
\makebox[3.5em][l]{\color{egink!55}47}\texttt{go to sidetable 2}\hspace{7.0pt}{\color{egink!62}\ldots\ you see a creditcard 1\,\ldots}\gmrdp
\par\vspace{1.7pt}
\makebox[3.5em][l]{\color{egink!55}48}\texttt{go to sidetable 3}\hspace{7.0pt}{\color{egink!62}\ldots\ you see a statue 2.}\gmrdp
\par\vspace{1.7pt}
\begin{tcolorbox}[gmr band]\makebox[3.5em][l]{\color{egink!55}49--50}2 replies cut off, no action\end{tcolorbox}\vspace*{-1.8pt}
\end{tcolorbox}\hfill
\begin{tcolorbox}[gmr B, equal height group=gmrThreer12]
{\color{egdeep!85!black}\itshape Task completed at step 16; the episode ends.}
\end{tcolorbox}\par\nointerlineskip

\vspace{3pt}
\begin{tcolorbox}[gmr foot, colback=egblush, colupper=egrose!55!black, equal height group=gmrThreef]Not completed after 50 steps; 27 without an action\end{tcolorbox}\hfill
\begin{tcolorbox}[gmr foot, colback=egsage, colupper=egdeep!70!black, equal height group=gmrThreef]Completed in 16 steps; 1 without an action\end{tcolorbox}\par\nointerlineskip

}
\vspace{6pt}
\caption{\textbf{Pick two: with the first newspaper in hand the unchanged agent looks for the second, wastes 27 of its 50 steps on replies cut at the token limit and never reaches it; after its own edit the agent puts the first in the drawer, fetches the second and is done in 16 steps.} The episode of \cref{fig:gameswon}d, step by step, read as \cref{fig:gamemore-1}; at step 10, the first step at which the two take different actions (at step 9 both replies are cut), each reply is quoted from the start of the last paragraph of its reasoning. Rule of selection as in \cref{fig:gamemore-1}: going down the ranking of the held-out tasks by the rise, with the edit, in the episodes completed over their 48 fresh seeds, from the task of \cref{fig:gamefresh}, the first task met of each of four further task types (this one the third), each with the seed, of those that turn from not completed to completed, that the edited agent finishes in the fewest steps.}
\label{fig:gamemore-3}
\end{figure}

\begin{figure}[tp]
{\setlength{\parskip}{0pt}%
\begin{tcolorbox}[enhanced, width=\linewidth, colback=white, colframe=egink!35, boxrule=0.5pt, arc=2pt, boxsep=0pt, left=8pt, right=8pt, top=4.5pt, bottom=3pt, fontupper=\sffamily\footnotesize]
\textbf{Task:} put a vase in safe.\quad\hfill\textbf{Type:} Pick \& place\qquad\textbf{Room:} never seen\par\vspace{4pt}
\hspace*{-1pt}\begin{tikzpicture}[x=1pt, y=1pt, line width=0.3pt, line join=miter]
\node[anchor=east, inner sep=0pt, font=\sffamily\scriptsize, text=gmrunch] at (-7, 26.0) {Unchanged};
\draw[gmrline, dash pattern=on 1.6pt off 1.6pt] (0, 33.0) -- (245.0, 33.0);
\draw[gmrline] (0, 20.0) -- (245.0, 20.0);
\filldraw[fill=gmrnewF, draw=gmrnewE] (0.0, 20.0) rectangle (4.9, 23.30);
\filldraw[fill=gmrnewF, draw=gmrnewE] (4.9, 20.0) rectangle (9.8, 25.84);
\filldraw[fill=gmrnoneF, draw=gmrnoneE] (9.8, 20.0) rectangle (14.7, 24.42);
\begin{scope}\clip (9.8, 20.0) rectangle (14.7, 24.42);\draw[gmrnoneE, line width=0.35pt] (8.8, 15.1) -- (15.7, 22.0);\draw[gmrnoneE, line width=0.35pt] (8.8, 18.1) -- (15.7, 25.0);\draw[gmrnoneE, line width=0.35pt] (8.8, 21.1) -- (15.7, 28.0);\draw[gmrnoneE, line width=0.35pt] (8.8, 24.1) -- (15.7, 31.0);\draw[gmrnoneE, line width=0.35pt] (8.8, 27.1) -- (15.7, 34.0);\draw[gmrnoneE, line width=0.35pt] (8.8, 30.1) -- (15.7, 37.0);\end{scope}
\filldraw[fill=gmrnewF, draw=gmrnewE] (14.7, 20.0) rectangle (19.6, 27.08);
\filldraw[fill=gmragainF, draw=gmragainE] (19.6, 20.0) rectangle (24.5, 25.18);
\filldraw[fill=gmragainF, draw=gmragainE] (24.5, 20.0) rectangle (29.4, 24.65);
\filldraw[fill=gmrwasteF, draw=gmrwasteE] (29.4, 20.0) rectangle (34.3, 33.00);
\filldraw[fill=gmrnoneF, draw=gmrnoneE] (34.3, 20.0) rectangle (39.2, 25.05);
\begin{scope}\clip (34.3, 20.0) rectangle (39.2, 25.05);\draw[gmrnoneE, line width=0.35pt] (33.3, 15.1) -- (40.2, 22.0);\draw[gmrnoneE, line width=0.35pt] (33.3, 18.1) -- (40.2, 25.0);\draw[gmrnoneE, line width=0.35pt] (33.3, 21.1) -- (40.2, 28.0);\draw[gmrnoneE, line width=0.35pt] (33.3, 24.1) -- (40.2, 31.0);\draw[gmrnoneE, line width=0.35pt] (33.3, 27.1) -- (40.2, 34.0);\draw[gmrnoneE, line width=0.35pt] (33.3, 30.1) -- (40.2, 37.0);\end{scope}
\filldraw[fill=gmrlookF, draw=gmrlookE] (39.2, 20.0) rectangle (44.1, 30.11);
\filldraw[fill=gmrnoneF, draw=gmrnoneE] (44.1, 20.0) rectangle (49.0, 25.28);
\begin{scope}\clip (44.1, 20.0) rectangle (49.0, 25.28);\draw[gmrnoneE, line width=0.35pt] (43.1, 15.1) -- (50.0, 22.0);\draw[gmrnoneE, line width=0.35pt] (43.1, 18.1) -- (50.0, 25.0);\draw[gmrnoneE, line width=0.35pt] (43.1, 21.1) -- (50.0, 28.0);\draw[gmrnoneE, line width=0.35pt] (43.1, 24.1) -- (50.0, 31.0);\draw[gmrnoneE, line width=0.35pt] (43.1, 27.1) -- (50.0, 34.0);\draw[gmrnoneE, line width=0.35pt] (43.1, 30.1) -- (50.0, 37.0);\end{scope}
\filldraw[fill=gmrlookF, draw=gmrlookE] (49.0, 20.0) rectangle (53.9, 24.72);
\filldraw[fill=gmragainF, draw=gmragainE] (53.9, 20.0) rectangle (58.8, 25.08);
\filldraw[fill=gmrnewF, draw=gmrnewE] (58.8, 20.0) rectangle (63.7, 24.39);
\filldraw[fill=gmrnewF, draw=gmrnewE] (63.7, 20.0) rectangle (68.6, 31.40);
\filldraw[fill=gmropenF, draw=gmropenE] (68.6, 20.0) rectangle (73.5, 25.46);
\filldraw[fill=gmragainF, draw=gmragainE] (73.5, 20.0) rectangle (78.4, 30.59);
\filldraw[fill=gmrnewF, draw=gmrnewE] (78.4, 20.0) rectangle (83.3, 26.22);
\filldraw[fill=gmropenF, draw=gmropenE] (83.3, 20.0) rectangle (88.2, 28.89);
\filldraw[fill=gmrwasteF, draw=gmrwasteE] (88.2, 20.0) rectangle (93.1, 33.00);
\filldraw[fill=gmrwasteF, draw=gmrwasteE] (93.1, 20.0) rectangle (98.0, 33.00);
\filldraw[fill=gmragainF, draw=gmragainE] (98.0, 20.0) rectangle (102.9, 29.47);
\filldraw[fill=gmragainF, draw=gmragainE] (102.9, 20.0) rectangle (107.8, 24.98);
\filldraw[fill=gmrlookF, draw=gmrlookE] (107.8, 20.0) rectangle (112.7, 25.13);
\filldraw[fill=gmrnoneF, draw=gmrnoneE] (112.7, 20.0) rectangle (117.6, 27.11);
\begin{scope}\clip (112.7, 20.0) rectangle (117.6, 27.11);\draw[gmrnoneE, line width=0.35pt] (111.7, 15.1) -- (118.6, 22.0);\draw[gmrnoneE, line width=0.35pt] (111.7, 18.1) -- (118.6, 25.0);\draw[gmrnoneE, line width=0.35pt] (111.7, 21.1) -- (118.6, 28.0);\draw[gmrnoneE, line width=0.35pt] (111.7, 24.1) -- (118.6, 31.0);\draw[gmrnoneE, line width=0.35pt] (111.7, 27.1) -- (118.6, 34.0);\draw[gmrnoneE, line width=0.35pt] (111.7, 30.1) -- (118.6, 37.0);\end{scope}
\filldraw[fill=gmrlookF, draw=gmrlookE] (117.6, 20.0) rectangle (122.5, 25.33);
\filldraw[fill=gmrnewF, draw=gmrnewE] (122.5, 20.0) rectangle (127.4, 24.42);
\filldraw[fill=gmrwasteF, draw=gmrwasteE] (127.4, 20.0) rectangle (132.3, 33.00);
\filldraw[fill=gmragainF, draw=gmragainE] (132.3, 20.0) rectangle (137.2, 32.59);
\filldraw[fill=gmrnewF, draw=gmrnewE] (137.2, 20.0) rectangle (142.1, 29.75);
\filldraw[fill=gmragainF, draw=gmragainE] (142.1, 20.0) rectangle (147.0, 31.96);
\filldraw[fill=gmrnoneF, draw=gmrnoneE] (147.0, 20.0) rectangle (151.9, 24.39);
\begin{scope}\clip (147.0, 20.0) rectangle (151.9, 24.39);\draw[gmrnoneE, line width=0.35pt] (146.0, 15.1) -- (152.9, 22.0);\draw[gmrnoneE, line width=0.35pt] (146.0, 18.1) -- (152.9, 25.0);\draw[gmrnoneE, line width=0.35pt] (146.0, 21.1) -- (152.9, 28.0);\draw[gmrnoneE, line width=0.35pt] (146.0, 24.1) -- (152.9, 31.0);\draw[gmrnoneE, line width=0.35pt] (146.0, 27.1) -- (152.9, 34.0);\draw[gmrnoneE, line width=0.35pt] (146.0, 30.1) -- (152.9, 37.0);\end{scope}
\filldraw[fill=gmrwasteF, draw=gmrwasteE] (151.9, 20.0) rectangle (156.8, 33.00);
\filldraw[fill=gmragainF, draw=gmragainE] (156.8, 20.0) rectangle (161.7, 29.24);
\filldraw[fill=gmrwasteF, draw=gmrwasteE] (161.7, 20.0) rectangle (166.6, 33.00);
\filldraw[fill=gmrwasteF, draw=gmrwasteE] (166.6, 20.0) rectangle (171.5, 33.00);
\filldraw[fill=gmrwasteF, draw=gmrwasteE] (171.5, 20.0) rectangle (176.4, 33.00);
\filldraw[fill=gmrnewF, draw=gmrnewE] (176.4, 20.0) rectangle (181.3, 29.60);
\filldraw[fill=gmropenF, draw=gmropenE] (181.3, 20.0) rectangle (186.2, 22.21);
\filldraw[fill=gmrlookF, draw=gmrlookE] (186.2, 20.0) rectangle (191.1, 28.38);
\filldraw[fill=gmragainF, draw=gmragainE] (191.1, 20.0) rectangle (196.0, 26.02);
\filldraw[fill=gmragainF, draw=gmragainE] (196.0, 20.0) rectangle (200.9, 25.94);
\filldraw[fill=gmragainF, draw=gmragainE] (200.9, 20.0) rectangle (205.8, 26.91);
\filldraw[fill=gmrwasteF, draw=gmrwasteE] (205.8, 20.0) rectangle (210.7, 33.00);
\filldraw[fill=gmrwasteF, draw=gmrwasteE] (210.7, 20.0) rectangle (215.6, 33.00);
\filldraw[fill=gmrnoneF, draw=gmrnoneE] (215.6, 20.0) rectangle (220.5, 31.96);
\begin{scope}\clip (215.6, 20.0) rectangle (220.5, 31.96);\draw[gmrnoneE, line width=0.35pt] (214.6, 15.1) -- (221.5, 22.0);\draw[gmrnoneE, line width=0.35pt] (214.6, 18.1) -- (221.5, 25.0);\draw[gmrnoneE, line width=0.35pt] (214.6, 21.1) -- (221.5, 28.0);\draw[gmrnoneE, line width=0.35pt] (214.6, 24.1) -- (221.5, 31.0);\draw[gmrnoneE, line width=0.35pt] (214.6, 27.1) -- (221.5, 34.0);\draw[gmrnoneE, line width=0.35pt] (214.6, 30.1) -- (221.5, 37.0);\end{scope}
\filldraw[fill=gmrlookF, draw=gmrlookE] (220.5, 20.0) rectangle (225.4, 31.38);
\filldraw[fill=gmrnoneF, draw=gmrnoneE] (225.4, 20.0) rectangle (230.3, 24.49);
\begin{scope}\clip (225.4, 20.0) rectangle (230.3, 24.49);\draw[gmrnoneE, line width=0.35pt] (224.4, 15.1) -- (231.3, 22.0);\draw[gmrnoneE, line width=0.35pt] (224.4, 18.1) -- (231.3, 25.0);\draw[gmrnoneE, line width=0.35pt] (224.4, 21.1) -- (231.3, 28.0);\draw[gmrnoneE, line width=0.35pt] (224.4, 24.1) -- (231.3, 31.0);\draw[gmrnoneE, line width=0.35pt] (224.4, 27.1) -- (231.3, 34.0);\draw[gmrnoneE, line width=0.35pt] (224.4, 30.1) -- (231.3, 37.0);\end{scope}
\filldraw[fill=gmrlookF, draw=gmrlookE] (230.3, 20.0) rectangle (235.2, 27.52);
\filldraw[fill=gmragainF, draw=gmragainE] (235.2, 20.0) rectangle (240.1, 25.84);
\filldraw[fill=gmragainF, draw=gmragainE] (240.1, 20.0) rectangle (245.0, 30.84);
\node[anchor=west, inner sep=0pt, font=\sffamily\scriptsize, text=gmrfail] at (251.0, 26.0) {not completed\hspace{0.45em}\textperiodcentered\hspace{0.45em}16{,}270 tokens};
\node[anchor=east, inner sep=0pt, font=\sffamily\scriptsize, text=gmredit] at (-7, 6.0) {\textbf{After its own edit}};
\draw[gmrline, dash pattern=on 1.6pt off 1.6pt] (0, 13.0) -- (245.0, 13.0);
\draw[gmrline] (0, 0.0) -- (245.0, 0.0);
\filldraw[fill=gmrnewF, draw=gmrnewE] (0.0, 0.0) rectangle (4.9, 5.28);
\filldraw[fill=gmrnewF, draw=gmrnewE] (4.9, 0.0) rectangle (9.8, 4.21);
\filldraw[fill=gmrnoneF, draw=gmrnoneE] (9.8, 0.0) rectangle (14.7, 3.53);
\begin{scope}\clip (9.8, 0.0) rectangle (14.7, 3.53);\draw[gmrnoneE, line width=0.35pt] (8.8, -4.9) -- (15.7, 2.0);\draw[gmrnoneE, line width=0.35pt] (8.8, -1.9) -- (15.7, 5.0);\draw[gmrnoneE, line width=0.35pt] (8.8, 1.1) -- (15.7, 8.0);\draw[gmrnoneE, line width=0.35pt] (8.8, 4.1) -- (15.7, 11.0);\draw[gmrnoneE, line width=0.35pt] (8.8, 7.1) -- (15.7, 14.0);\draw[gmrnoneE, line width=0.35pt] (8.8, 10.1) -- (15.7, 17.0);\end{scope}
\filldraw[fill=gmrnewF, draw=gmrnewE] (14.7, 0.0) rectangle (19.6, 5.28);
\filldraw[fill=gmrnewF, draw=gmrnewE] (19.6, 0.0) rectangle (24.5, 6.63);
\filldraw[fill=gmrnewF, draw=gmrnewE] (24.5, 0.0) rectangle (29.4, 7.57);
\filldraw[fill=gmrtakeF, draw=gmrtakeE] (29.4, 0.0) rectangle (34.3, 4.85);
\filldraw[fill=gmrnewF, draw=gmrnewE] (34.3, 0.0) rectangle (39.2, 4.34);
\filldraw[fill=gmropenF, draw=gmropenE] (39.2, 0.0) rectangle (44.1, 4.49);
\filldraw[fill=gmrputF, draw=gmrputE] (44.1, 0.0) rectangle (49.0, 4.72);
\node[anchor=west, inner sep=0pt, font=\sffamily\scriptsize\bfseries, text=gmredit] at (55.0, 6.0) {completed in 10 steps\hspace{0.45em}\textperiodcentered\hspace{0.45em}2{,}005 tokens};
\end{tikzpicture}
\end{tcolorbox}\par\vspace{5pt}\nointerlineskip
\begin{tcolorbox}[gmr head, colback=egbase, equal height group=gmrFourh]Unchanged agent\end{tcolorbox}\hfill
\begin{tcolorbox}[gmr head, colback=egdeep, equal height group=gmrFourh]After its own edit\end{tcolorbox}\par\nointerlineskip

\begin{tcolorbox}[gmr A, equal height group=gmrFourr0]
\makebox[3.5em][l]{\color{egink!55}1}\texttt{go to shelf 1}\hspace{7.0pt}{\color{egink!62}\ldots\ you see nothing.}\gmrdp
\end{tcolorbox}\hfill
\begin{tcolorbox}[gmr B, equal height group=gmrFourr0]
\makebox[3.5em][l]{\color{egink!55}1}\texttt{go to shelf 1}\hspace{7.0pt}{\color{egink!62}\ldots\ you see nothing.}\gmrdp
\end{tcolorbox}\par\nointerlineskip

\begin{tcolorbox}[gmr A, equal height group=gmrFourr1]
\makebox[3.5em][l]{\color{egink!55}2}\texttt{go to shelf 2}\hspace{7.0pt}{\color{egink!62}\ldots\ you see nothing.}\gmrdp
\end{tcolorbox}\hfill
\begin{tcolorbox}[gmr B, equal height group=gmrFourr1]
\makebox[3.5em][l]{\color{egink!55}2}\texttt{go to shelf 2}\hspace{7.0pt}{\color{egink!62}\ldots\ you see nothing.}\gmrdp
\end{tcolorbox}\par\nointerlineskip

\begin{tcolorbox}[gmr A, equal height group=gmrFourr2]
\makebox[3.5em][l]{\color{egink!55}3}\texttt{go to shelf 3}\hspace{7.0pt}{\color{egink!62}Nothing happens.}\gmrdp
\end{tcolorbox}\hfill
\begin{tcolorbox}[gmr B, equal height group=gmrFourr2]
\makebox[3.5em][l]{\color{egink!55}3}\texttt{go to shelf 3}\hspace{7.0pt}{\color{egink!62}Nothing happens.}\gmrdp
\end{tcolorbox}\par\nointerlineskip

\begin{tcolorbox}[gmr A, equal height group=gmrFourr3]
\makebox[3.5em][l]{\color{egink!55}4}\texttt{go to shelf 4}\hspace{7.0pt}{\color{egink!62}\ldots\ you see nothing.}\gmrdp
\end{tcolorbox}\hfill
\begin{tcolorbox}[gmr B, equal height group=gmrFourr3]
\makebox[3.5em][l]{\color{egink!55}4}\texttt{go to shelf 4}\hspace{7.0pt}{\color{egink!62}\ldots\ you see nothing.}\gmrdp
\end{tcolorbox}\par\nointerlineskip

\begin{tcolorbox}[gmr A, equal height group=gmrFourr4]
\makebox[3.5em][l]{\color{egink!55}5}\texttt{go to shelf 1}\hspace{7.0pt}{\color{egink!62}\ldots\ you see nothing.}\gmrdp\par\vspace{1.5pt}{\leftskip=3.5em\rightskip=0pt plus 2em\hyphenpenalty=10000\fontsize{7}{8.3}\selectfont\color{egink!85}{\itshape its reply ends:} The goal is to put a vase in the safe.\par 1. \textbf{Current Status}: I am at shelf 4. I have checked shelves 2, 3, and 4, and found nothing on them.\par 2. \textbf{Missing Step}: I have not yet found or picked up the vase. The task requires me to have the vase in my inventory before I can put it in the safe.\par 3. \textbf{Strategy}: I need to continue searching for the vase. Since shelves 2-4 are empty, I should check other shelves. Shelf 1 is the next logical place to check in sequence, or I could check other common locations like side tables or dressers. Given the sequence of checking shelves, checking shelf 1 is a reasonable next step to see if the vase is there.\par 4. \textbf{Action Selection}: I will go to shelf 1 to check if the vase is located there.\par \textless{}action\textgreater{} go to shelf 1 \textless{}/action\textgreater{}\ \ {\color{egbase!80!black}\itshape 204 tokens}\gmrdp\par}
\end{tcolorbox}\hfill
\begin{tcolorbox}[gmr B, equal height group=gmrFourr4]
\makebox[3.5em][l]{\color{egink!55}5}\texttt{go to shelf 5}\hspace{7.0pt}{\color{egink!62}\ldots\ you see nothing.}\gmrdp\par\vspace{1.5pt}{\leftskip=3.5em\rightskip=0pt plus 2em\hyphenpenalty=10000\fontsize{7}{8.3}\selectfont\color{egink!85}{\itshape its reply ends:} [\,\ldots\,]\ * I can go directly to the safe, but I don't have the vase yet, so going there now would be premature unless I know the vase is already in my inventory (which I don't, as I haven't picked anything up).\par * Checking shelf 5 is a logical continuation of the search pattern (checking shelves sequentially).\par 4. \textbf{Select action}: Go to shelf 5 to check for the vase.\par \textless{}action\textgreater{} go to shelf 5 \textless{}/action\textgreater{}\ \ {\color{egdeep!80!black}\itshape 261 tokens}\gmrdp\par}
\end{tcolorbox}\par\nointerlineskip

\begin{tcolorbox}[gmr A, equal height group=gmrFourr5]
\makebox[3.5em][l]{\color{egink!55}6}\texttt{go to shelf 2}\hspace{7.0pt}{\color{egink!62}\ldots\ you see nothing.}\gmrdp
\end{tcolorbox}\hfill
\begin{tcolorbox}[gmr B, equal height group=gmrFourr5]
\makebox[3.5em][l]{\color{egink!55}6}\texttt{go to shelf 6}\hspace{7.0pt}{\color{egink!62}\ldots\ you see a vase 2.}\gmrdp
\end{tcolorbox}\par\nointerlineskip

\begin{tcolorbox}[gmr A, gmr wastecell, equal height group=gmrFourr6]
\makebox[3.5em][l]{\color{egink!55}7}reply cut off at 512 tokens: no action
\end{tcolorbox}\hfill
\begin{tcolorbox}[gmr B, equal height group=gmrFourr6]
\makebox[3.5em][l]{\color{egink!55}7}\texttt{take vase 2 from shelf 6}\gmrdp
\end{tcolorbox}\par\nointerlineskip

\begin{tcolorbox}[gmr A, equal height group=gmrFourr7]
\makebox[3.5em][l]{\color{egink!55}8}\texttt{go to shelf 3}\hspace{7.0pt}{\color{egink!62}Nothing happens.}\gmrdp
\end{tcolorbox}\hfill
\begin{tcolorbox}[gmr B, equal height group=gmrFourr7]
\makebox[3.5em][l]{\color{egink!55}8}\texttt{go to safe 1}\hspace{7.0pt}{\color{egink!62}\ldots\ The safe 1 is closed.}\gmrdp
\end{tcolorbox}\par\nointerlineskip

\begin{tcolorbox}[gmr A, equal height group=gmrFourr8]
\makebox[3.5em][l]{\color{egink!55}9}\texttt{examine shelf 3}\hspace{7.0pt}{\color{egink!62}On the shelf 3, you see nothing.}\gmrdp
\end{tcolorbox}\hfill
\begin{tcolorbox}[gmr B, equal height group=gmrFourr8]
\makebox[3.5em][l]{\color{egink!55}9}\texttt{open safe 1}\hspace{7.0pt}{\color{egink!62}\ldots\ you see a keychain 3.}\gmrdp
\end{tcolorbox}\par\nointerlineskip

\begin{tcolorbox}[gmr A, equal height group=gmrFourr9]
\makebox[3.5em][l]{\color{egink!55}10}\texttt{go to shelf 2}\hspace{7.0pt}{\color{egink!62}Nothing happens.}\gmrdp
\end{tcolorbox}\hfill
\begin{tcolorbox}[gmr B, equal height group=gmrFourr9]
\makebox[3.5em][l]{\color{egink!55}10}\texttt{move vase 2 to safe 1}\hspace{7.0pt}{\color{egink!62}You move the vase 2\,\ldots}\gmrdp
\end{tcolorbox}\par\nointerlineskip

\begin{tcolorbox}[gmr A, equal height group=gmrFourr10]
\makebox[3.5em][l]{\color{egink!55}11}\texttt{examine shelf 2}\hspace{7.0pt}{\color{egink!62}On the shelf 2, you see nothing.}\gmrdp
\par\vspace{1.7pt}
\makebox[3.5em][l]{\color{egink!55}12}\texttt{go to shelf 1}\hspace{7.0pt}{\color{egink!62}\ldots\ you see nothing.}\gmrdp
\par\vspace{1.7pt}
\makebox[3.5em][l]{\color{egink!55}13}\texttt{go to shelf 3}\hspace{7.0pt}{\color{egink!62}\ldots\ you see nothing.}\gmrdp
\par\vspace{1.7pt}
\makebox[3.5em][l]{\color{egink!55}14}\texttt{go to cabinet 1}\hspace{7.0pt}{\color{egink!62}\ldots\ The cabinet 1 is closed.}\gmrdp
\par\vspace{1.7pt}
\makebox[3.5em][l]{\color{egink!55}15}\texttt{open cabinet 1}\hspace{7.0pt}{\color{egink!62}\ldots\ you see a newspaper 1.}\gmrdp
\par\vspace{1.7pt}
\makebox[3.5em][l]{\color{egink!55}16}\texttt{go to shelf 1}\hspace{7.0pt}{\color{egink!62}\ldots\ you see nothing.}\gmrdp
\par\vspace{1.7pt}
\makebox[3.5em][l]{\color{egink!55}17--44}{\itshape\color{egink!62}[\,\ldots\ 19 actions and 9 cut replies, not shown\ \ldots\,]}\gmrdp
\par\vspace{1.7pt}
\makebox[3.5em][l]{\color{egink!55}45}\texttt{go to shelf 3}\hspace{7.0pt}{\color{egink!62}Nothing happens.}\gmrdp
\par\vspace{1.7pt}
\makebox[3.5em][l]{\color{egink!55}46}\texttt{examine shelf 3}\hspace{7.0pt}{\color{egink!62}On the shelf 3, you see nothing.}\gmrdp
\par\vspace{1.7pt}
\makebox[3.5em][l]{\color{egink!55}47}\texttt{go to shelf 2}\hspace{7.0pt}{\color{egink!62}Nothing happens.}\gmrdp
\par\vspace{1.7pt}
\makebox[3.5em][l]{\color{egink!55}48}\texttt{examine shelf 2}\hspace{7.0pt}{\color{egink!62}On the shelf 2, you see nothing.}\gmrdp
\par\vspace{1.7pt}
\makebox[3.5em][l]{\color{egink!55}49}\texttt{go to shelf 1}\hspace{7.0pt}{\color{egink!62}\ldots\ you see nothing.}\gmrdp
\par\vspace{1.7pt}
\makebox[3.5em][l]{\color{egink!55}50}\texttt{go to shelf 3}\hspace{7.0pt}{\color{egink!62}\ldots\ you see nothing.}\gmrdp
\end{tcolorbox}\hfill
\begin{tcolorbox}[gmr B, equal height group=gmrFourr10]
{\color{egdeep!85!black}\itshape Task completed at step 10; the episode ends.}
\end{tcolorbox}\par\nointerlineskip

\vspace{3pt}
\begin{tcolorbox}[gmr foot, colback=egblush, colupper=egrose!55!black, equal height group=gmrFourf]Not completed after 50 steps; 10 without an action\end{tcolorbox}\hfill
\begin{tcolorbox}[gmr foot, colback=egsage, colupper=egdeep!70!black, equal height group=gmrFourf]Completed in 10 steps; none without an action\end{tcolorbox}\par\nointerlineskip

}
\vspace{6pt}
\caption{\textbf{Pick \& place: the unchanged agent looks on four shelves, turns back to the first and keeps coming back to the same four for the rest of its 50 steps, ten of them wasted on replies cut at the token limit, and never looks on the sixth, where the vase stands; after its own edit the agent goes on to the fifth and the sixth shelf, takes the vase and is done in ten steps.} The episode of \cref{fig:gameswon}f, step by step, read as \cref{fig:gamemore-1}; at step 5, where the two first part, each reply is quoted from the start of the last paragraph of its reasoning (where the quote would run to more than eight lines, its last four lines). Of the unchanged agent's 36 rows after the edited agent has finished, the first six and the last six are shown, and one line counts the steps between. Rule of selection as in \cref{fig:gamemore-1}: going down the ranking of the held-out tasks by the rise, with the edit, in the episodes completed over their 48 fresh seeds, from the task of \cref{fig:gamefresh}, the first task met of each of four further task types (this one the fourth), each with the seed, of those that turn from not completed to completed, that the edited agent finishes in the fewest steps.}
\label{fig:gamemore-4}
\end{figure}

\clearpage
\begin{figure}[!t]
\centering
\includegraphics[width=\linewidth]{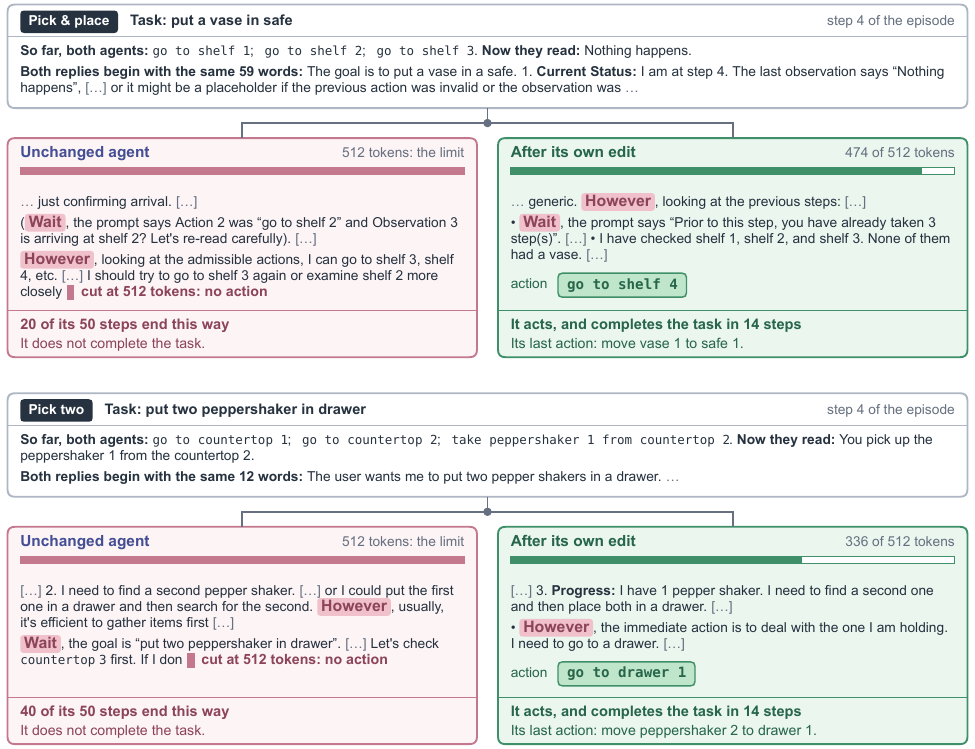}
\caption{\textbf{The parting of the main text recurs in every type of task: the unchanged agent deliberates until its reply is cut, the edited agent decides and acts.} Six more first partings, drawn as in \cref{fig:fork}a, one per type of task (continued in \cref{fig:forkmore-2,fig:forkmore-3}), from the held-out tasks, 48 fresh seeds each. The card above a pair gives the task, what both agents did and read up to the step, and the words both replies begin with. Rule: from the pool of \cref{fig:fork} (in the tasks the edited agent completes more often, the 161 pairs of episodes that turn from not completed to completed and first part in this way), for each type the pair in which the edit saves the most wasted steps (ties: lowest task id, then lowest seed); tasks with the instruction of \cref{fig:fork} are left out.}
\label{fig:forkmore-1}
\end{figure}

\begin{figure}[!tp]
\centering
\includegraphics[width=\linewidth]{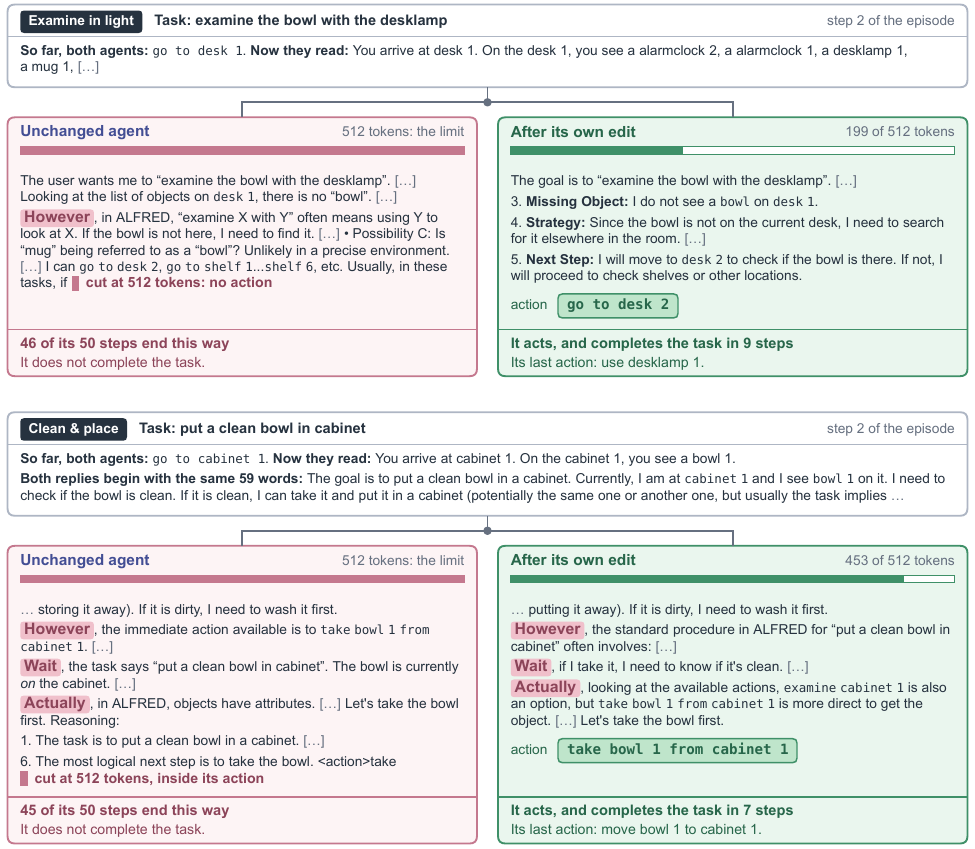}
\caption{\textbf{Examine in light and Clean \& place: the unchanged agent names the action the edited agent takes, and is cut before taking it.} Continues \cref{fig:forkmore-1} (same rule of selection). Ten or more shared first words are set once, in the card above the two replies; every ``Wait'', ``However'' and ``Actually'' of a reply is shown, in all six pairs. In the lower pair the unchanged agent is cut at the limit while writing the action.}
\label{fig:forkmore-2}
\end{figure}

\begin{figure}[!tp]
\centering
\includegraphics[width=\linewidth]{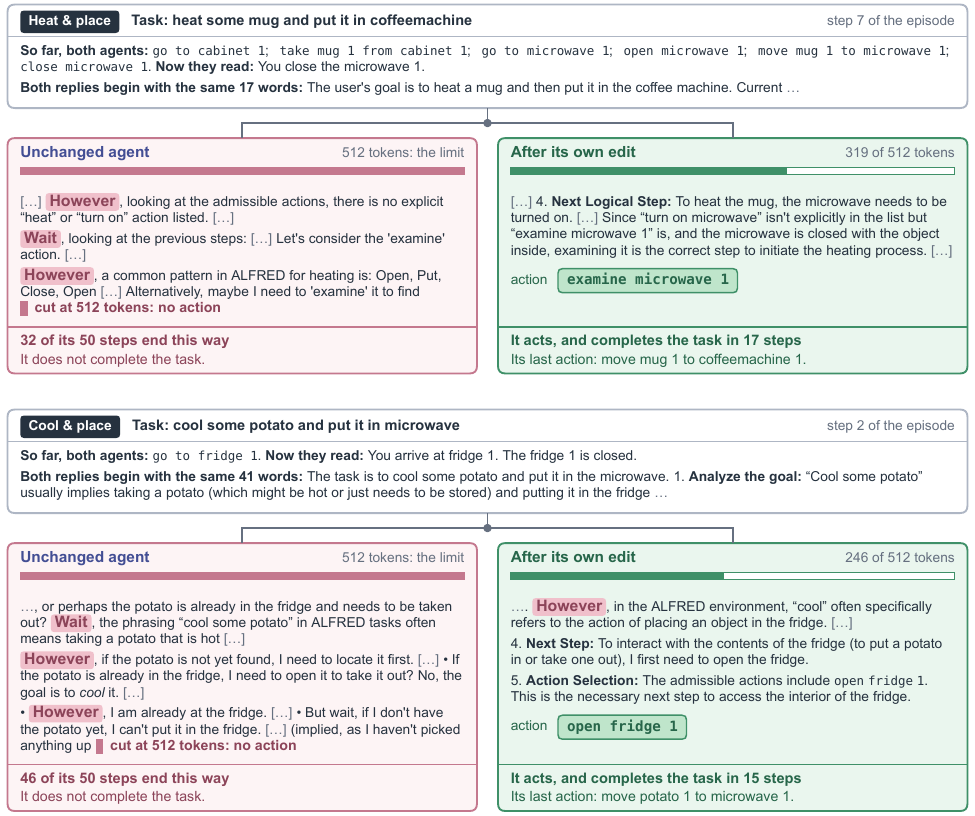}
\caption{\textbf{Heat \& place and Cool \& place: at the same appliance with the same history, the unchanged agent considers the edited agent's next action and reaches the limit without taking it.} Continues \cref{fig:forkmore-1} (same rule of selection). The action is examining the microwave in the upper pair and opening the fridge in the lower one.}
\label{fig:forkmore-3}
\end{figure}

\subsection{The step where the two agents part}
\label{app:partings}
\Cref{fig:forkmore-1,fig:forkmore-2,fig:forkmore-3} show six more first partings in the form of \cref{fig:fork}, one
for each task type.

\clearpage
\subsection{The night that kept its edit, turn by turn}
\label{app:examples}

\Cref{fig:night-story,fig:night-story-2} follow the night that kept its edit in the agent's own words, from its
diagnosis to its decision. Every round not given a card is listed between the cards. The turns that get a card were
picked by a rule fixed in advance and stated in the caption of \cref{fig:night-story}. The night reads like the
notebook of a careful experimenter: the agent tests every change, reverses an edit that points the wrong way, checks its edit on tasks it had not used and
on full episodes, and settles on a milder edit when the strongest makes its actions repetitive.

\begin{figure}[p]
\centering
\begingroup\sffamily\setlength{\parskip}{0pt}\setlength{\parindent}{0pt}%
\def\nsgap{3.6pt}\def\nstiny{\fontsize{6.5}{7.5}\selectfont}%
\tcbset{nscard/.style={enhanced, left skip=0.95in, boxrule=0.45pt, arc=2pt, left=7pt, right=5pt, top=3.2pt, bottom=3.6pt, before skip=0pt, after skip=5.6pt, fontupper=\sffamily\footnotesize, colupper=egink},
  nsevent/.style={enhanced, left skip=0.95in, frame hidden, interior hidden, boxrule=0pt, left=7pt, right=5pt, top=0pt, bottom=0pt, before skip=0pt, after skip=5.6pt, fontupper=\sffamily\scriptsize, colupper=egink!66},
  nschip/.style={on line, enhanced, boxrule=0.4pt, arc=1.6pt, left=3pt, right=3pt, top=0.9pt, bottom=0.7pt, boxsep=0pt, fontupper=\sffamily\scriptsize}}%
\noindent\hspace*{0.95in}{\scriptsize\color{egink!72}\tikz[baseline=-0.6ex]\fill[egdeep] (0,0) circle (2.6pt);\;its test beyond 2\,SE in its favour\hfill\tikz[baseline=-0.6ex]\fill[egink!42] (0,0) circle (2.6pt);\;within 2\,SE, or no test\hfill\tikz[baseline=-0.6ex]\filldraw[fill=white, draw=egink!45, line width=0.7pt] (0,0) circle (1.9pt);\;every other round and edit, with its result\hspace*{4pt}}\par\vspace{7pt}
\begin{tcolorbox}[nscard, colback=white, colframe=egink!20, borderline west={2.4pt}{0pt}{egink!35},
  overlay={\draw[egink!20, line width=0.8pt] ([xshift=-12pt,yshift=-8.2pt]frame.north west) -- ([xshift=-12pt,yshift=-\nsgap]frame.south west); \fill[white] ([xshift=-12pt,yshift=-8.2pt]frame.north west) circle (4pt); \fill[egink!42] ([xshift=-12pt,yshift=-8.2pt]frame.north west) circle (2.9pt); \node[anchor=north east, align=right, inner sep=0pt, font=\sffamily\footnotesize, text=egink] at ([xshift=-12pt,yshift=-4.6pt]frame.north west) [xshift=-8pt] {\textbf{turn 8}\\[0.6pt]{\nstiny\color{egink!55}0 of 20 rounds}};}]
{\color{egink}\textbf{names the failure}}\par\vspace{0.8pt}{\scriptsize\color{egink!58}from its record of 128 opening episodes: 1,366 of its 4,418 steps carry no well-formed action}\par\vspace{2.4pt}``All 1366 invalid steps hit exactly 512 tokens - the token limit. The model generates too much verbose reasoning before the action, gets cut off, and produces malformed actions. [\,\ldots\,] I need to find what internal representations drive this verbosity.''
\end{tcolorbox}
\begin{tcolorbox}[nsevent,
  overlay={\draw[egink!20, line width=0.8pt] ([xshift=-12pt,yshift=\nsgap]frame.north west) -- ([xshift=-12pt,yshift=-\nsgap]frame.south west); \filldraw[fill=white, draw=egink!45, line width=0.7pt] ([xshift=-12pt,yshift=-4.6pt]frame.north west) circle (1.9pt); \node[anchor=north east, align=right, inner sep=0pt, font=\sffamily\scriptsize, text=egink!55] at ([xshift=-12pt,yshift=-1.2pt]frame.north west) [xshift=-8pt] {turn 16};}]
{\color{egink!85}\textbf{round 1}:} reads its own hidden states on 64 of its practice steps, 32 concise and 32 cut off, and takes their difference at layer 63 as a direction
\end{tcolorbox}
\begin{tcolorbox}[nscard, colback=white, colframe=egink!20, borderline west={2.4pt}{0pt}{egink!35},
  overlay={\draw[egink!20, line width=0.8pt] ([xshift=-12pt,yshift=\nsgap]frame.north west) -- ([xshift=-12pt,yshift=-\nsgap]frame.south west); \fill[white] ([xshift=-12pt,yshift=-8.2pt]frame.north west) circle (4pt); \fill[egink!42] ([xshift=-12pt,yshift=-8.2pt]frame.north west) circle (2.9pt); \node[anchor=north east, align=right, inner sep=0pt, font=\sffamily\footnotesize, text=egink] at ([xshift=-12pt,yshift=-4.6pt]frame.north west) [xshift=-8pt] {\textbf{turn 19}\\[0.6pt]{\nstiny\color{egink!55}1 of 20 rounds}};}]
{\color{egink}\textbf{edits the last layer}}\par\vspace{0.8pt}{\scriptsize\color{egink!58}writes layer 63, attention output, strength +1}\par\vspace{2.4pt}``Now I'll test removing the verbosity direction from the output projections. I'll start with the attention output projection (\texttt{o\_proj}) since attention often controls response style.''
\end{tcolorbox}
\begin{tcolorbox}[nscard, colback=white, colframe=egink!20, borderline west={2.4pt}{0pt}{egink!35},
  overlay={\draw[egink!20, line width=0.8pt] ([xshift=-12pt,yshift=\nsgap]frame.north west) -- ([xshift=-12pt,yshift=-\nsgap]frame.south west); \fill[white] ([xshift=-12pt,yshift=-8.2pt]frame.north west) circle (4pt); \fill[egink!42] ([xshift=-12pt,yshift=-8.2pt]frame.north west) circle (2.9pt); \node[anchor=north east, align=right, inner sep=0pt, font=\sffamily\footnotesize, text=egink] at ([xshift=-12pt,yshift=-4.6pt]frame.north west) [xshift=-8pt] {\textbf{turn 27}\\[0.6pt]{\nstiny\color{egink!55}2 of 20 rounds}};}]
{\color{egink}\textbf{reverses the sign}}\hfill\tcbox[nschip, colback=white, colframe=egink!32, colupper=egink!85]{+1.4 points (SE 1.7), 64 short episodes}\par\vspace{0.8pt}{\scriptsize\color{egink!58}round 2 tested layer 63 at strength +1 on short episodes of 64 tasks; it writes strength $-$1}\par\vspace{2.4pt}``The edit didn't help - it slightly increased verbosity. Let me try the opposite direction (adding the verbosity direction instead of removing) or try an earlier layer.''
\end{tcolorbox}
\begin{tcolorbox}[nscard, colback=white, colframe=egink!20, borderline west={2.4pt}{0pt}{egink!35},
  overlay={\draw[egink!20, line width=0.8pt] ([xshift=-12pt,yshift=\nsgap]frame.north west) -- ([xshift=-12pt,yshift=-\nsgap]frame.south west); \fill[white] ([xshift=-12pt,yshift=-8.2pt]frame.north west) circle (4pt); \fill[egink!42] ([xshift=-12pt,yshift=-8.2pt]frame.north west) circle (2.9pt); \node[anchor=north east, align=right, inner sep=0pt, font=\sffamily\footnotesize, text=egink] at ([xshift=-12pt,yshift=-4.6pt]frame.north west) [xshift=-8pt] {\textbf{turn 29}\\[0.6pt]{\nstiny\color{egink!55}3 of 20 rounds}};}]
{\color{egink}\textbf{moves to layer 48}}\hfill\tcbox[nschip, colback=white, colframe=egink!32, colupper=egink!85]{$-$0.4 points (SE 1.6), 64 short episodes}\par\vspace{0.8pt}{\scriptsize\color{egink!58}round 3 tested layer 63 at strength $-$1 on short episodes of the same 64 tasks as round 2; it writes layer 48, MLP output, strength +2 (its code runs in turn 31)}\par\vspace{2.4pt}``The edit isn't showing significant effects. Let me try a different approach - examining the MLP activations and trying an earlier layer where verbosity decisions might be made.''
\end{tcolorbox}
\begin{tcolorbox}[nsevent,
  overlay={\draw[egink!20, line width=0.8pt] ([xshift=-12pt,yshift=\nsgap]frame.north west) -- ([xshift=-12pt,yshift=-\nsgap]frame.south west); \filldraw[fill=white, draw=egink!45, line width=0.7pt] ([xshift=-12pt,yshift=-4.6pt]frame.north west) circle (1.9pt); \node[anchor=north east, align=right, inner sep=0pt, font=\sffamily\scriptsize, text=egink!55] at ([xshift=-12pt,yshift=-1.2pt]frame.north west) [xshift=-8pt] {turn 32};}]
{\color{egink!85}\textbf{round 4}:} tests layer 48 at strength +2 on short episodes of the same 64 tasks as round 2: {\color{egink}0.0 points (SE 2.1)}
\end{tcolorbox}
\begin{tcolorbox}[nscard, colback=white, colframe=egink!20, borderline west={2.4pt}{0pt}{egink!35},
  overlay={\draw[egink!20, line width=0.8pt] ([xshift=-12pt,yshift=\nsgap]frame.north west) -- ([xshift=-12pt,yshift=-\nsgap]frame.south west); \fill[white] ([xshift=-12pt,yshift=-8.2pt]frame.north west) circle (4pt); \fill[egink!42] ([xshift=-12pt,yshift=-8.2pt]frame.north west) circle (2.9pt); \node[anchor=north east, align=right, inner sep=0pt, font=\sffamily\footnotesize, text=egink] at ([xshift=-12pt,yshift=-4.6pt]frame.north west) [xshift=-8pt] {\textbf{turn 38}\\[0.6pt]{\nstiny\color{egink!55}4 of 20 rounds}};}]
{\color{egink}\textbf{moves to an early layer}}\par\vspace{0.8pt}{\scriptsize\color{egink!58}writes layer 16, attention output, strength +5 (its code runs in turn 40)}\par\vspace{2.4pt}``The pattern shows failed responses start with `The user' while successful ones often start with `The goal' or `The task'. The verbosity decision happens very early. Let me try editing an earlier layer.''
\end{tcolorbox}
\begin{tcolorbox}[nscard, colback=white, colframe=egink!20, borderline west={2.4pt}{0pt}{egink!35},
  overlay={\draw[egink!20, line width=0.8pt] ([xshift=-12pt,yshift=\nsgap]frame.north west) -- ([xshift=-12pt,yshift=-\nsgap]frame.south west); \fill[white] ([xshift=-12pt,yshift=-8.2pt]frame.north west) circle (4pt); \fill[egink!42] ([xshift=-12pt,yshift=-8.2pt]frame.north west) circle (2.9pt); \node[anchor=north east, align=right, inner sep=0pt, font=\sffamily\footnotesize, text=egink] at ([xshift=-12pt,yshift=-4.6pt]frame.north west) [xshift=-8pt] {\textbf{turn 42}\\[0.6pt]{\nstiny\color{egink!55}5 of 20 rounds}};}]
{\color{egink}\textbf{reverses the sign}}\hfill\tcbox[nschip, colback=white, colframe=egink!32, colupper=egink!85]{+3.2 points (SE 2.3), 64 short episodes}\par\vspace{0.8pt}{\scriptsize\color{egink!58}round 5 tested layer 16 at strength +5 on short episodes of the same 64 tasks as round 2; it writes strength $-$5}\par\vspace{2.4pt}``The edit made things worse - increased verbosity. Let me try the opposite direction.''
\end{tcolorbox}
\begin{tcolorbox}[nsevent,
  overlay={\draw[egink!20, line width=0.8pt] ([xshift=-12pt,yshift=\nsgap]frame.north west) -- ([xshift=-12pt,yshift=-\nsgap]frame.south west); \filldraw[fill=white, draw=egink!45, line width=0.7pt] ([xshift=-12pt,yshift=-4.6pt]frame.north west) circle (1.9pt); \node[anchor=north east, align=right, inner sep=0pt, font=\sffamily\scriptsize, text=egink!55] at ([xshift=-12pt,yshift=-1.2pt]frame.north west) [xshift=-8pt] {turn 43};}]
{\color{egink!85}\textbf{round 6}:} tests layer 16 at strength $-$5 on short episodes of the same 64 tasks as round 2: {\color{egink}$-$3.6 points (SE 2.0)}
\end{tcolorbox}
\begin{tcolorbox}[nsevent,
  overlay={\draw[egink!20, line width=0.8pt] ([xshift=-12pt,yshift=\nsgap]frame.north west) -- ([xshift=-12pt,yshift=-4.6pt]frame.north west); \filldraw[fill=white, draw=egink!45, line width=0.7pt] ([xshift=-12pt,yshift=-4.6pt]frame.north west) circle (1.9pt); \node[anchor=north east, align=right, inner sep=0pt, font=\sffamily\scriptsize, text=egink!55] at ([xshift=-12pt,yshift=-1.2pt]frame.north west) [xshift=-8pt] {turn 44};}]
{\color{egink!85}\textbf{edit}:} writes strength $-$15
\end{tcolorbox}
\endgroup
\caption{\textbf{The night that kept its edit, turn by turn (I): where to edit, and which way.} The second night in the agent's own words: every quote is what it wrote in that turn (emphasis marks removed, omissions marked [\,\ldots\,]); left of each card, the turn and the rounds of forward passes it has used of its 20. \emph{Which turns get a card} (a rule fixed before any quote was chosen): its first diagnosis; every turn at which it changes where it edits or the sign of its edit (when that code runs in a later turn, the card shows its words just before it); its reading of every paired test whose result lies beyond two standard errors either way; and its decisions, to ask for the confirmation and to submit. A card that follows a test carries that test's result: the share of steps without an action, edited minus unchanged, on the same tasks and seeds (green: beyond 2\,SE in its favour; grey: within 2\,SE). Short episodes: the first 10 steps of each task. Strengths are $\alpha$ in $W \leftarrow W - \alpha\,uu^{\top}W$; the $-10$ it keeps is the strength 10 of \cref{fig:edit}. Every other round and every other edit it writes is listed between the cards (open dots), so that none is left out. The night continues in \cref{fig:night-story-2}.}
\label{fig:night-story}
\end{figure}
\begin{figure}[p]
\centering
\begingroup\sffamily\setlength{\parskip}{0pt}\setlength{\parindent}{0pt}%
\def\nsgap{3.6pt}\def\nstiny{\fontsize{6.5}{7.5}\selectfont}%
\tcbset{nscard/.style={enhanced, left skip=0.95in, boxrule=0.45pt, arc=2pt, left=7pt, right=5pt, top=3.2pt, bottom=3.6pt, before skip=0pt, after skip=5.6pt, fontupper=\sffamily\footnotesize, colupper=egink},
  nsevent/.style={enhanced, left skip=0.95in, frame hidden, interior hidden, boxrule=0pt, left=7pt, right=5pt, top=0pt, bottom=0pt, before skip=0pt, after skip=5.6pt, fontupper=\sffamily\scriptsize, colupper=egink!66},
  nschip/.style={on line, enhanced, boxrule=0.4pt, arc=1.6pt, left=3pt, right=3pt, top=0.9pt, bottom=0.7pt, boxsep=0pt, fontupper=\sffamily\scriptsize}}%
\noindent\hspace*{0.95in}{\scriptsize\color{egink!72}\tikz[baseline=-0.6ex]\fill[egdeep] (0,0) circle (2.6pt);\;its test beyond 2\,SE in its favour\hfill\tikz[baseline=-0.6ex]\fill[egink!42] (0,0) circle (2.6pt);\;within 2\,SE, or no test\hfill\tikz[baseline=-0.6ex]\filldraw[fill=white, draw=egink!45, line width=0.7pt] (0,0) circle (1.9pt);\;every other round and edit, with its result\hspace*{4pt}}\par\vspace{7pt}
\begin{tcolorbox}[nscard, colback=egmist!70, colframe=egdeep!35, borderline west={2.4pt}{0pt}{egdeep},
  overlay={\draw[egink!20, line width=0.8pt] ([xshift=-12pt,yshift=-8.2pt]frame.north west) -- ([xshift=-12pt,yshift=-\nsgap]frame.south west); \fill[white] ([xshift=-12pt,yshift=-8.2pt]frame.north west) circle (4pt); \fill[egdeep] ([xshift=-12pt,yshift=-8.2pt]frame.north west) circle (2.9pt); \node[anchor=north east, align=right, inner sep=0pt, font=\sffamily\footnotesize, text=egink] at ([xshift=-12pt,yshift=-4.6pt]frame.north west) [xshift=-8pt] {\textbf{turn 46}\\[0.6pt]{\nstiny\color{egink!55}7 of 20 rounds}};}]
{\color{egdeep!80!black}\textbf{first result beyond 2 SE}}\hfill\tcbox[nschip, colback=egdeep, colframe=egdeep, colupper=white]{\bfseries $-$8.4 points (SE 2.5), 64 short episodes}\par\vspace{0.8pt}{\scriptsize\color{egink!58}round 7 tested layer 16 at strength $-$15 on short episodes of the same 64 tasks as round 2}\par\vspace{2.4pt}``Excellent! This is a significant improvement! The edit reduced malformed actions from 21.6\% to 12.5\% (beyond 2 SE) [\,\ldots\,] Now I need to test this on games I haven't used to choose the edit.''
\end{tcolorbox}
\begin{tcolorbox}[nsevent,
  overlay={\draw[egink!20, line width=0.8pt] ([xshift=-12pt,yshift=\nsgap]frame.north west) -- ([xshift=-12pt,yshift=-\nsgap]frame.south west); \filldraw[fill=white, draw=egink!45, line width=0.7pt] ([xshift=-12pt,yshift=-4.6pt]frame.north west) circle (1.9pt); \node[anchor=north east, align=right, inner sep=0pt, font=\sffamily\scriptsize, text=egink!55] at ([xshift=-12pt,yshift=-1.2pt]frame.north west) [xshift=-8pt] {turn 47};}]
{\color{egink!85}\textbf{round 8}:} tests layer 16 at strength $-$15 on short episodes of 64 tasks it had not used: {\color{egink}$-$1.1 points (SE 1.7)}
\end{tcolorbox}
\begin{tcolorbox}[nsevent,
  overlay={\draw[egink!20, line width=0.8pt] ([xshift=-12pt,yshift=\nsgap]frame.north west) -- ([xshift=-12pt,yshift=-\nsgap]frame.south west); \filldraw[fill=white, draw=egink!45, line width=0.7pt] ([xshift=-12pt,yshift=-4.6pt]frame.north west) circle (1.9pt); \node[anchor=north east, align=right, inner sep=0pt, font=\sffamily\scriptsize, text=egink!55] at ([xshift=-12pt,yshift=-1.2pt]frame.north west) [xshift=-8pt] {turn 49};}]
{\color{egink!85}\textbf{round 9}:} tests layer 16 at strength $-$15 on short episodes of 128 tasks, 64 of them new: {\color{egink}$-$0.9 points (SE 1.5)}
\end{tcolorbox}
\begin{tcolorbox}[nsevent,
  overlay={\draw[egink!20, line width=0.8pt] ([xshift=-12pt,yshift=\nsgap]frame.north west) -- ([xshift=-12pt,yshift=-\nsgap]frame.south west); \filldraw[fill=white, draw=egink!45, line width=0.7pt] ([xshift=-12pt,yshift=-4.6pt]frame.north west) circle (1.9pt); \node[anchor=north east, align=right, inner sep=0pt, font=\sffamily\scriptsize, text=egink!55] at ([xshift=-12pt,yshift=-1.2pt]frame.north west) [xshift=-8pt] {turn 54};}]
{\color{egink!85}\textbf{round 10}:} tests 4 strengths side by side on short episodes of the same 64 tasks as round 8, each against its unchanged self: at $-$5, {\color{egink}+1.4 (SE 2.4)}; at $-$10, {\color{egink}$-$2.1 (SE 2.6)}; at $-$15, {\color{egink}$-$0.8 (SE 2.2)}; at $-$20, {\color{egink}$-$2.9 (SE 2.8)}
\end{tcolorbox}
\begin{tcolorbox}[nsevent,
  overlay={\draw[egink!20, line width=0.8pt] ([xshift=-12pt,yshift=\nsgap]frame.north west) -- ([xshift=-12pt,yshift=-\nsgap]frame.south west); \filldraw[fill=white, draw=egink!45, line width=0.7pt] ([xshift=-12pt,yshift=-4.6pt]frame.north west) circle (1.9pt); \node[anchor=north east, align=right, inner sep=0pt, font=\sffamily\scriptsize, text=egink!55] at ([xshift=-12pt,yshift=-1.2pt]frame.north west) [xshift=-8pt] {turn 55};}]
{\color{egink!85}\textbf{edit}:} writes strength $-$20
\end{tcolorbox}
\begin{tcolorbox}[nscard, colback=egmist!70, colframe=egdeep!35, borderline west={2.4pt}{0pt}{egdeep},
  overlay={\draw[egink!20, line width=0.8pt] ([xshift=-12pt,yshift=\nsgap]frame.north west) -- ([xshift=-12pt,yshift=-\nsgap]frame.south west); \fill[white] ([xshift=-12pt,yshift=-8.2pt]frame.north west) circle (4pt); \fill[egdeep] ([xshift=-12pt,yshift=-8.2pt]frame.north west) circle (2.9pt); \node[anchor=north east, align=right, inner sep=0pt, font=\sffamily\footnotesize, text=egink] at ([xshift=-12pt,yshift=-4.6pt]frame.north west) [xshift=-8pt] {\textbf{turn 58}\\[0.6pt]{\nstiny\color{egink!55}11 of 20 rounds}};}]
{\color{egdeep!80!black}\textbf{beyond 2 SE on tasks it had not used}}\hfill\tcbox[nschip, colback=egdeep, colframe=egdeep, colupper=white]{\bfseries $-$7.2 points (SE 2.4), 64 short episodes}\par\vspace{0.8pt}{\scriptsize\color{egink!58}round 11 tested layer 16 at strength $-$20 on short episodes of 64 tasks it had not used}\par\vspace{2.4pt}``The edit shows significant improvements on fresh validation data - malformed actions reduced from 28.4\% to 21.2\% (beyond 2 SE) [\,\ldots\,] Now let me test on full games to see if completion rates improve.''
\end{tcolorbox}
\begin{tcolorbox}[nscard, colback=egmist!70, colframe=egdeep!35, borderline west={2.4pt}{0pt}{egdeep},
  overlay={\draw[egink!20, line width=0.8pt] ([xshift=-12pt,yshift=\nsgap]frame.north west) -- ([xshift=-12pt,yshift=-\nsgap]frame.south west); \fill[white] ([xshift=-12pt,yshift=-8.2pt]frame.north west) circle (4pt); \fill[egdeep] ([xshift=-12pt,yshift=-8.2pt]frame.north west) circle (2.9pt); \node[anchor=north east, align=right, inner sep=0pt, font=\sffamily\footnotesize, text=egink] at ([xshift=-12pt,yshift=-4.6pt]frame.north west) [xshift=-8pt] {\textbf{turn 60}\\[0.6pt]{\nstiny\color{egink!55}12 of 20 rounds}};}]
{\color{egdeep!80!black}\textbf{beyond 2 SE in full episodes}}\hfill\tcbox[nschip, colback=egdeep, colframe=egdeep, colupper=white]{\bfseries $-$9.4 points (SE 2.6), 32 full episodes}\par\vspace{0.8pt}{\scriptsize\color{egink!58}round 12 tested layer 16 at strength $-$20 on full episodes of 32 tasks it had not used}\par\vspace{2.4pt}``The edit reduces malformed actions significantly but completion rate didn't change. The repetitive play increased which is concerning.''
\end{tcolorbox}
\begin{tcolorbox}[nscard, colback=egmist!70, colframe=egdeep!35, borderline west={2.4pt}{0pt}{egdeep},
  overlay={\draw[egink!20, line width=0.8pt] ([xshift=-12pt,yshift=\nsgap]frame.north west) -- ([xshift=-12pt,yshift=-\nsgap]frame.south west); \fill[white] ([xshift=-12pt,yshift=-8.2pt]frame.north west) circle (4pt); \fill[egdeep] ([xshift=-12pt,yshift=-8.2pt]frame.north west) circle (2.9pt); \node[anchor=north east, align=right, inner sep=0pt, font=\sffamily\footnotesize, text=egink] at ([xshift=-12pt,yshift=-4.6pt]frame.north west) [xshift=-8pt] {\textbf{turn 64}\\[0.6pt]{\nstiny\color{egink!55}13 of 20 rounds}};}]
{\color{egdeep!80!black}\textbf{beyond 2 SE again; weakens the edit}}\hfill\tcbox[nschip, colback=egdeep, colframe=egdeep, colupper=white]{\bfseries $-$10.4 points (SE 2.4), 32 full episodes}\par\vspace{0.8pt}{\scriptsize\color{egink!58}round 13 tested layer 16 at strength $-$20 on full episodes of 32 tasks it had not used; it writes strength $-$10}\par\vspace{2.4pt}``The edit reduces malformed actions but increases repetitive play. Let me try a smaller alpha for better balance.''\par\vspace{1.4pt}{\scriptsize\color{egink!70}its code: \texttt{alpha = $-$10.0}}
\end{tcolorbox}
\begin{tcolorbox}[nscard, colback=egmist!70, colframe=egdeep!35, borderline west={2.4pt}{0pt}{egdeep},
  overlay={\draw[egink!20, line width=0.8pt] ([xshift=-12pt,yshift=\nsgap]frame.north west) -- ([xshift=-12pt,yshift=-\nsgap]frame.south west); \fill[white] ([xshift=-12pt,yshift=-8.2pt]frame.north west) circle (4pt); \fill[egdeep] ([xshift=-12pt,yshift=-8.2pt]frame.north west) circle (2.9pt); \node[anchor=north east, align=right, inner sep=0pt, font=\sffamily\footnotesize, text=egink] at ([xshift=-12pt,yshift=-4.6pt]frame.north west) [xshift=-8pt] {\textbf{turn 66}\\[0.6pt]{\nstiny\color{egink!55}14 of 20 rounds}};}]
{\color{egdeep!80!black}\textbf{beyond 2 SE at the weaker strength}}\hfill\tcbox[nschip, colback=egdeep, colframe=egdeep, colupper=white]{\bfseries $-$7.6 points (SE 3.3), 32 full episodes}\par\vspace{0.8pt}{\scriptsize\color{egink!58}round 14 tested layer 16 at strength $-$10 on full episodes of the same 32 tasks as round 13}\par\vspace{2.4pt}``The edit with alpha=$-$10 shows significant reduction in malformed actions (43.1\% $\rightarrow$ 34.8\%, beyond 2 SE) while completion rate doesn't drop. Let me test on one more fresh set before confirmation.''
\end{tcolorbox}
\begin{tcolorbox}[nscard, colback=white, colframe=egink!20, borderline west={2.4pt}{0pt}{egink!35},
  overlay={\draw[egink!20, line width=0.8pt] ([xshift=-12pt,yshift=\nsgap]frame.north west) -- ([xshift=-12pt,yshift=-\nsgap]frame.south west); \fill[white] ([xshift=-12pt,yshift=-8.2pt]frame.north west) circle (4pt); \fill[egink!42] ([xshift=-12pt,yshift=-8.2pt]frame.north west) circle (2.9pt); \node[anchor=north east, align=right, inner sep=0pt, font=\sffamily\footnotesize, text=egink] at ([xshift=-12pt,yshift=-4.6pt]frame.north west) [xshift=-8pt] {\textbf{turn 68}\\[0.6pt]{\nstiny\color{egink!55}15 of 20 rounds}};}]
{\color{egink}\textbf{asks for its confirmation}}\hfill\tcbox[nschip, colback=white, colframe=egink!32, colupper=egink!85]{$-$2.0 points (SE 2.6), 32 full episodes}\par\vspace{0.8pt}{\scriptsize\color{egink!58}round 15 tested layer 16 at strength $-$10 on full episodes of 32 tasks it had not used}\par\vspace{2.4pt}``The edit shows consistent reduction in malformed actions across multiple tests. Let me proceed to confirmation.''
\end{tcolorbox}
\begin{tcolorbox}[nscard, colback=egmist, colframe=egdeep!70, borderline west={2.4pt}{0pt}{egdeep}, boxrule=0.6pt,
  overlay={\draw[egink!20, line width=0.8pt] ([xshift=-12pt,yshift=\nsgap]frame.north west) -- ([xshift=-12pt,yshift=-8.2pt]frame.north west); \fill[white] ([xshift=-12pt,yshift=-8.2pt]frame.north west) circle (4pt); \fill[egdeep] ([xshift=-12pt,yshift=-8.2pt]frame.north west) circle (2.9pt); \node[anchor=north east, align=right, inner sep=0pt, font=\sffamily\footnotesize, text=egink] at ([xshift=-12pt,yshift=-4.6pt]frame.north west) [xshift=-8pt] {\textbf{turn 69}\\[0.6pt]{\nstiny\color{egink!55}15 of 20 rounds}};}]
{\color{egdeep!80!black}\textbf{its confirmation; submits}}\hfill\tcbox[nschip, colback=egdeep, colframe=egdeep, colupper=white]{\bfseries $-$6.3 points (SE 1.1), 256 reserved tasks}\par\vspace{0.8pt}{\scriptsize\color{egink!58}the confirmation: 256 tasks reserved from it, not counted in its 20 rounds}\par\vspace{2.4pt}``Clearly fewer steps lost to the failure (malformed actions reduced significantly beyond 2 SE) [\,\ldots\,] Completion rate did not drop (actually improved slightly, within noise) [\,\ldots\,] DECISION: SUBMIT''
\end{tcolorbox}
\endgroup
\caption{\textbf{The night that kept its edit, turn by turn (II): testing the edit, and keeping it.} Continued from \cref{fig:night-story}, with the same rule and marks. The direction it edits along is the one it read at layer 63; the edit it keeps writes that direction into the attention output of layer 16 at strength $-$10. Its confirmation runs on 256 tasks reserved from it and does not count against its 20 rounds.}
\label{fig:night-story-2}
\end{figure}

\clearpage
\begin{figure}[t]
\centering
\includegraphics[width=\linewidth]{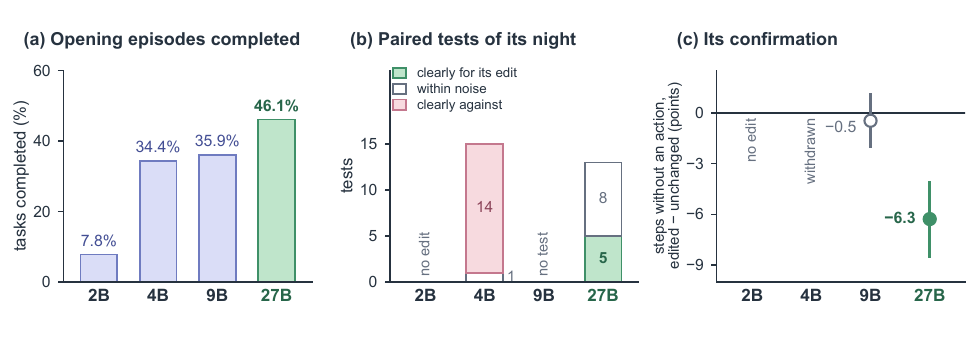}
\caption{\textbf{The same night at four sizes of one model family: only the 27B agent runs tests that come out for its edit, and only its confirmation is clear.} (a)~Opening episodes completed with the published weights, the 128 practice tasks that open every night (\cref{tab:scale}). (b)~The paired tests of the model's night by their result: beyond two standard errors for its edit, within two, or beyond two against it, read on steps without an action. The 2B model completes too few tasks for a test to measure a change and makes no edit; the 9B agent asked for no test. (c)~The confirmation on reserved tasks: the paired change in steps without an action, edited minus unchanged, with two standard errors (filled: beyond two). For the 27B agent, (b) and (c) are the night that kept its edit; the 4B agent withdrew without a confirmation.}
\label{fig:appscale}
\end{figure}
\subsection{The same night at four sizes, in numbers}
\label{app:scale}

\Cref{fig:appscale} puts numbers on \cref{fig:scalecards}. The 2B model completes \scTwoOff\% of its opening episodes,
too few for a paired test to measure a change, and makes no edit. The 4B and 9B models complete about a third of
theirs (\ascFourOpen\% and \ascNineOpen\%), and the 27B model \ascBigOpen\%. What separates the sizes is what they do
with a night. The 4B agent ran \ascFourTests{} paired tests; \ascFourDown{} came out clearly against its edits and none
for one, and it withdrew without asking for a confirmation. The 9B agent wrote and ran \ascNineCode{} programs in
\scNineTurns{} turns and never asked for a forward pass; when the turns of its night were used up, the confirmation was run on the edit it
held and came out at $-$\ascNineConf{} points (SE \ascNineConfSE), within two standard errors, and it withdrew. The 27B agent ran
\ascBigTests{} paired tests, \ascBigUp{} of them clearly for its edit and none against, and its confirmation came out at
$-$\ascBigConf{} points (SE \ascBigConfSE). Measuring one's way to an edit that holds is an ability that appears
between the 9B and the 27B model.

\end{document}